\documentclass[10pt]{IEEEtran}
\usepackage{amsmath,amsthm,amssymb}
\usepackage{graphicx,epstopdf,epsfig}
\usepackage{adjustbox}
\usepackage[dvipsnames]{xcolor}
\usepackage[caption=false,font=footnotesize]{subfig}
\usepackage[linesnumbered,ruled]{algorithm2e}
\usepackage{float}
\usepackage{url}

\def\I{\mathbb{I}}
\def\A{\mathbf{A}}
\def\c{\boldsymbol{c}}
\def\m{\boldsymbol{m}}

\newtheorem{theorem}{Theorem}[section]
\newtheorem{proposition}[theorem]{Proposition}

\begin{document}

\title{Fast Adaptive Bilateral Filtering}

\author{Ruturaj~G.~Gavaskar and~Kunal~N.~Chaudhury,~\IEEEmembership{Senior~Member,~IEEE}

\thanks{Address: Department of Electrical Engineering, Indian Institute of Science, Bangalore 560012, India. K.~N.~Chaudhury was supported by an EMR Grant SERB/F/6047/2016-2017 from the Department of Science and Technology, Government of India. Correspondence: kunal@iisc.ac.in.}}

\markboth{Submitted to IEEE Transactions on Image Processing}{}
\maketitle

\begin{abstract}
In the classical bilateral filter, a fixed Gaussian range kernel is used along with a spatial kernel for edge-preserving smoothing.
We consider a generalization of this filter, the so-called adaptive bilateral filter, where the center and width of the Gaussian range kernel is allowed to change from pixel to pixel. Though this variant was originally proposed for sharpening and noise removal, it can also be used for other applications such as artifact removal and texture filtering. Similar to the bilateral filter, the brute-force implementation of its adaptive counterpart requires intense  computations. While several fast algorithms have been proposed in the literature for bilateral filtering, most of them work only with a fixed range kernel. In this paper, we propose a fast algorithm for adaptive bilateral filtering, whose complexity does not scale with the spatial filter width. This is based on the observation that the concerned filtering can be performed purely in range space using an appropriately defined local histogram. We show that by replacing the histogram with a polynomial and the finite range-space sum with an integral, we can approximate the filter using analytic functions. In particular, an efficient algorithm is derived using the following innovations: the polynomial is fitted by matching its moments to those of the target histogram (this is done using fast convolutions), and the analytic functions are recursively computed using integration-by-parts. Our algorithm can accelerate the brute-force implementation by at least $20 \times$, without perceptible distortions in the visual quality. We demonstrate  the effectiveness of our algorithm for sharpening, JPEG deblocking, and texture filtering.
\end{abstract}


\section{Introduction}
\label{sec:Int}

The bilateral filter \cite{Aurich1995,Smith1997,Tomasi1998} is widely used in computer vision and image processing for edge-preserving smoothing \cite{Paris2009book}. Unlike linear convolutional filters, the bilateral filter uses a range kernel along with a spatial kernel, where both kernels are usually Gaussian \cite{Tomasi1998}. The input to the range kernel is the intensity difference between the pixel of interest and its neighbor. If the difference is large (e.g., the pixels are from different sides of an edge), then the weight assigned to the neighboring pixel is small, and it is essentially excluded from the aggregation. This mechanism, which avoids the mixing of pixels with large intensity differences, ensures the preservation of sharp edges. However, this also makes the filter non-linear and computation intensive.

An adaptive variant of the bilateral filter was introduced in \cite{Zhang2008}, where the center and width of the Gaussian range kernel is allowed to change from pixel to pixel.
It was used for image sharpness enhancement along with noise removal.
Adaptation of the range kernel was necessary since the standard bilateral filter cannot perform sharpening.
The amount of sharpening and noise removal at a particular pixel is controlled by adapting the center and width of the kernel.
A variable-width range kernel was also used in \cite{Zhang2009,Mangiat2011} for removing compression and registration artifacts.

\begin{figure*}[t!]
    \centering
    \subfloat[Input ($600 \times 450$).]{\includegraphics[height=0.20\linewidth,keepaspectratio]{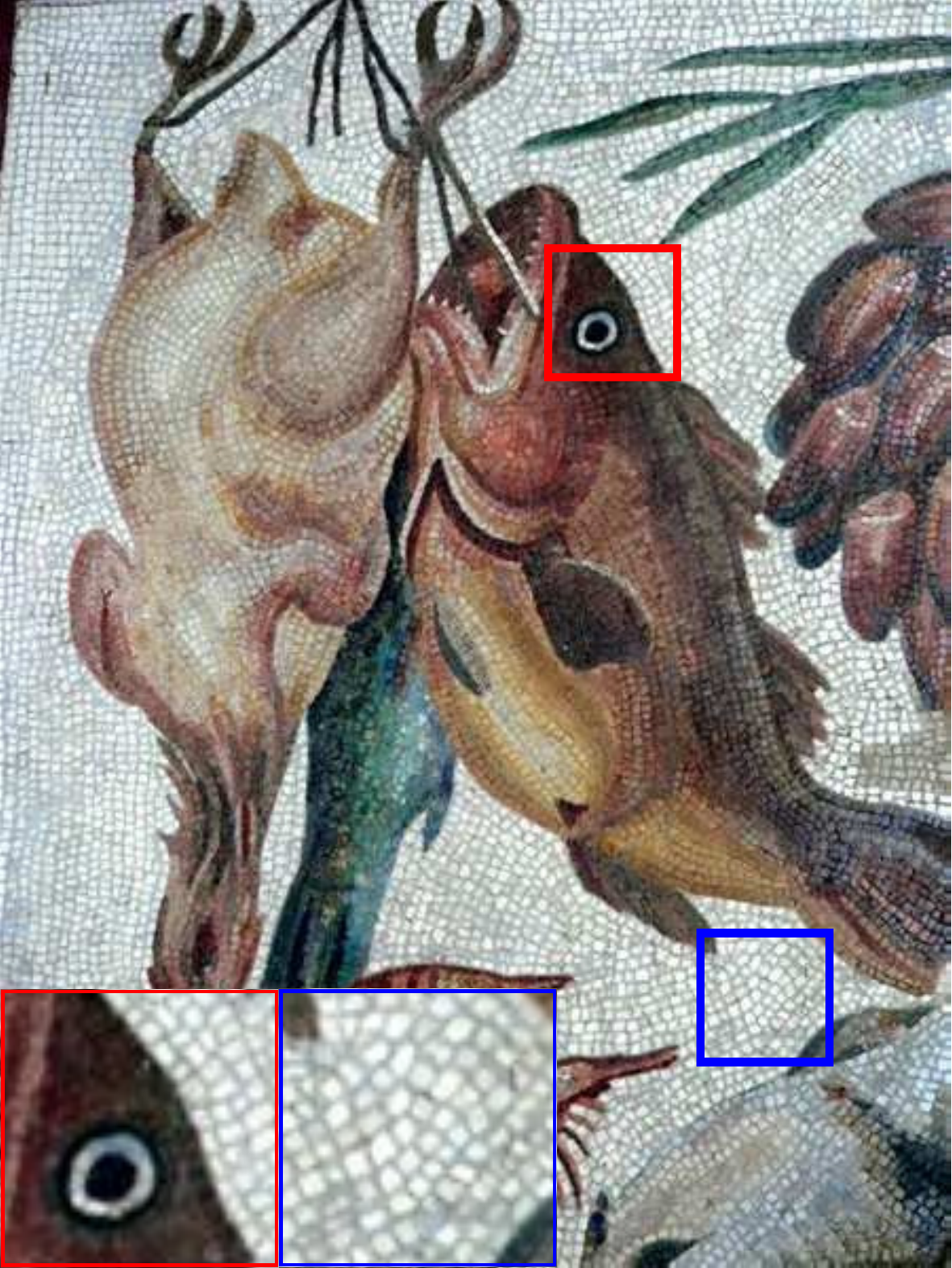}}
    \hspace{0.1mm}
    \subfloat[Bilateral, $\sigma=30$.]{\includegraphics[height=0.20\linewidth,keepaspectratio]{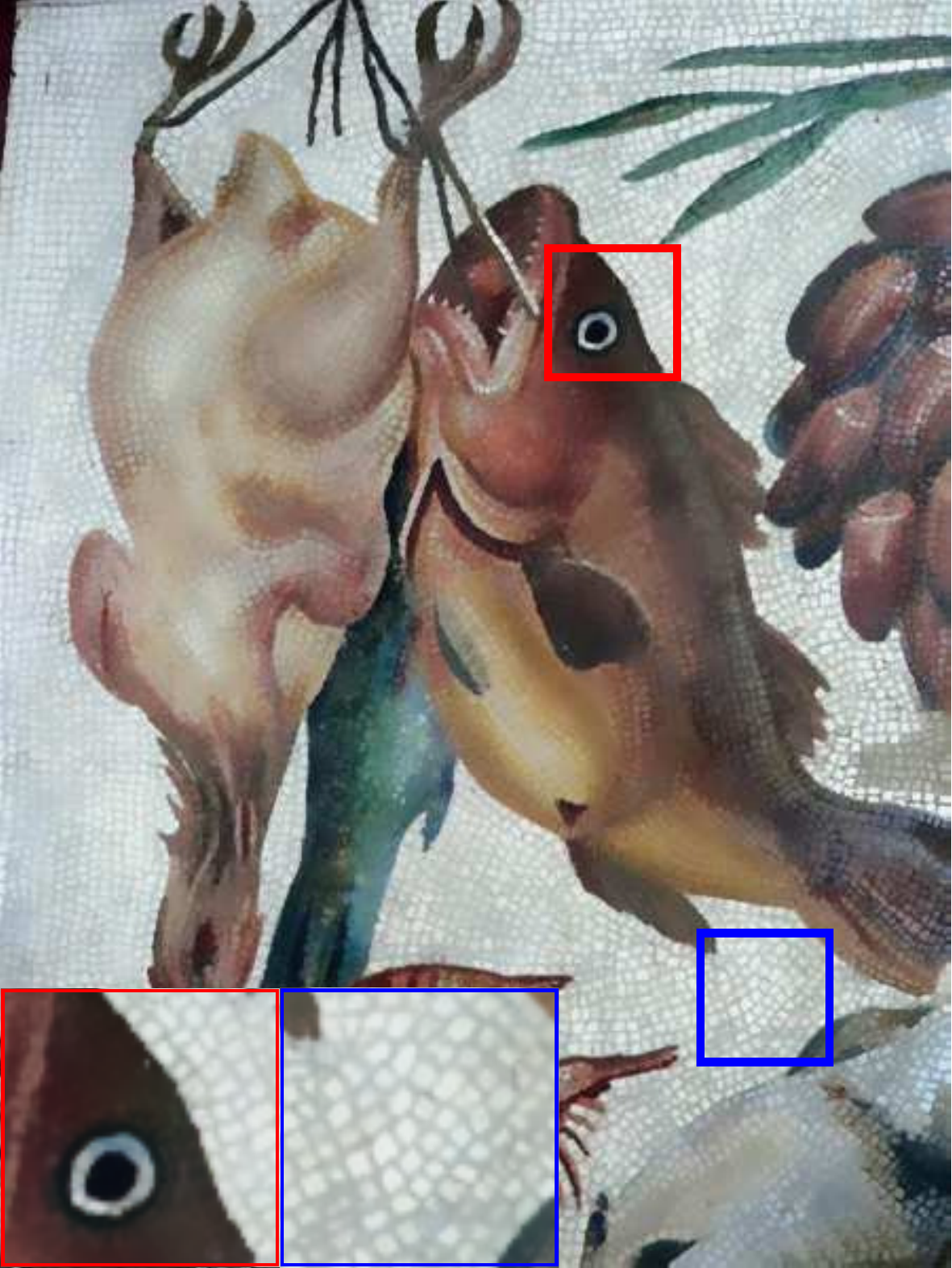}}
    \hspace{0.1mm}
    \subfloat[Bilateral, $\sigma=120$.]{\includegraphics[height=0.20\linewidth,keepaspectratio]{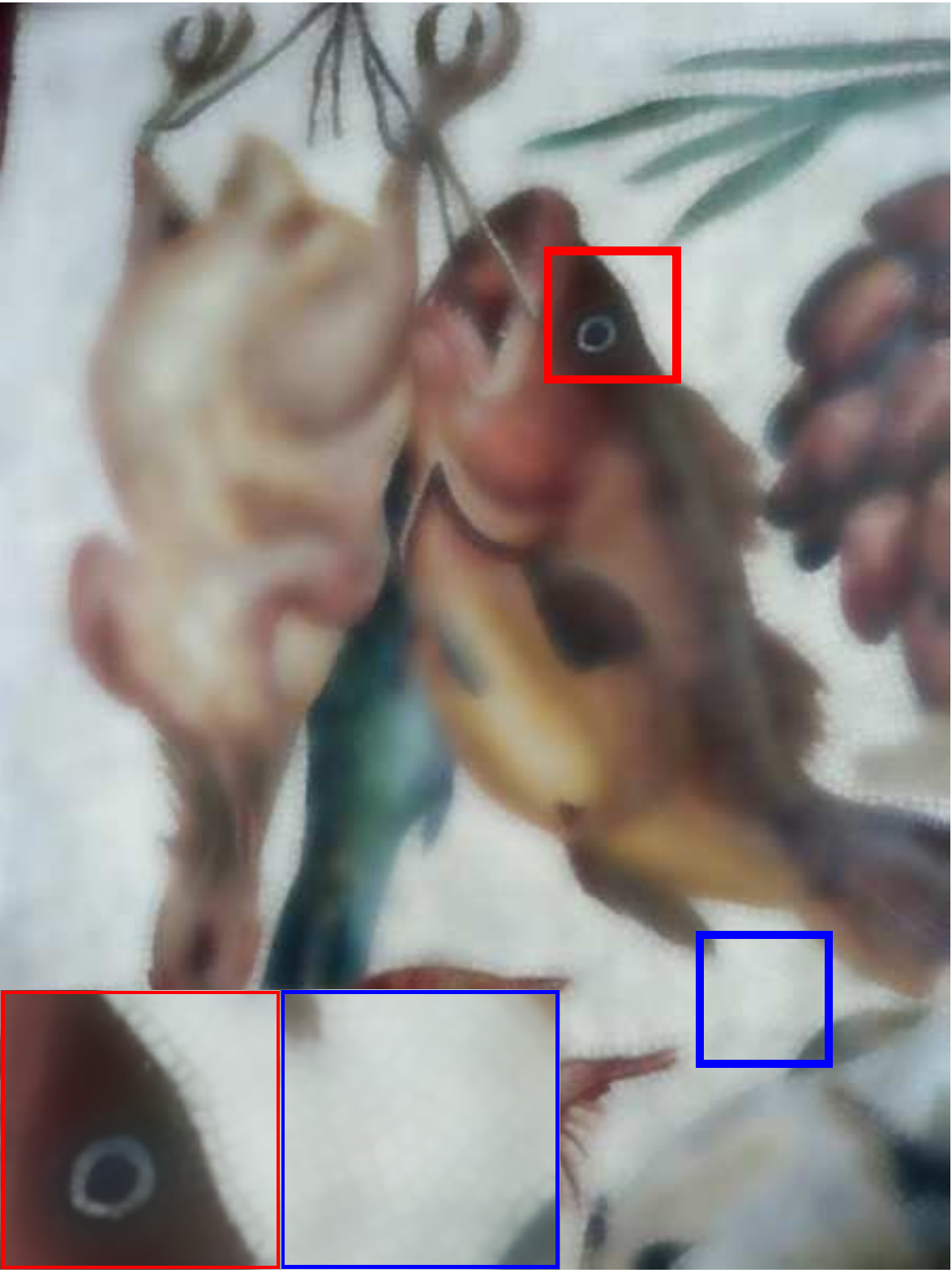}}
    \hspace{0.1mm}
    \subfloat[Adaptive bilateral.]{\includegraphics[height=0.20\linewidth,keepaspectratio]{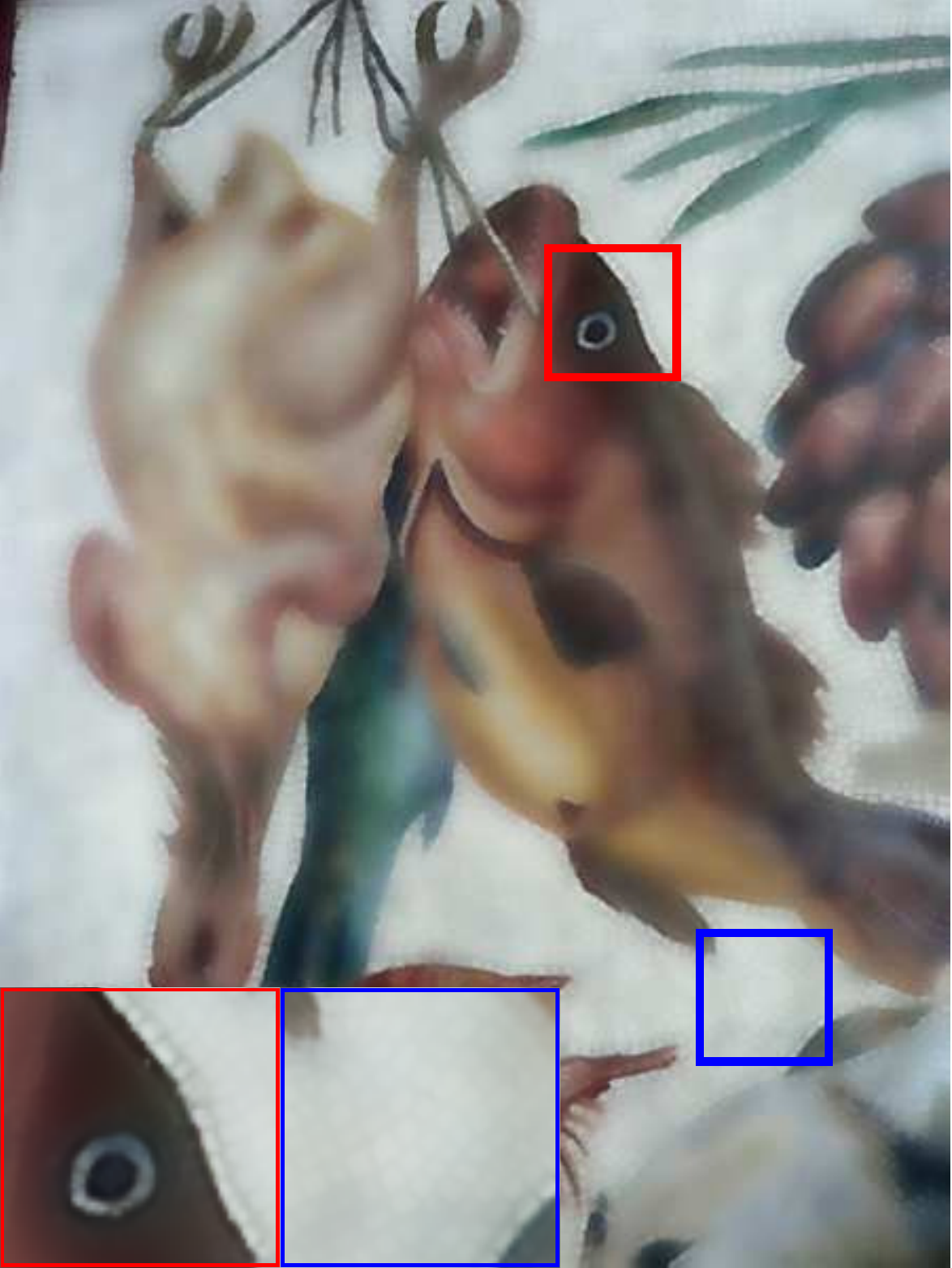}}
    \hspace{0.1mm}
    \subfloat[$\sigma(i)$.]{\includegraphics[height=0.20\linewidth,keepaspectratio]{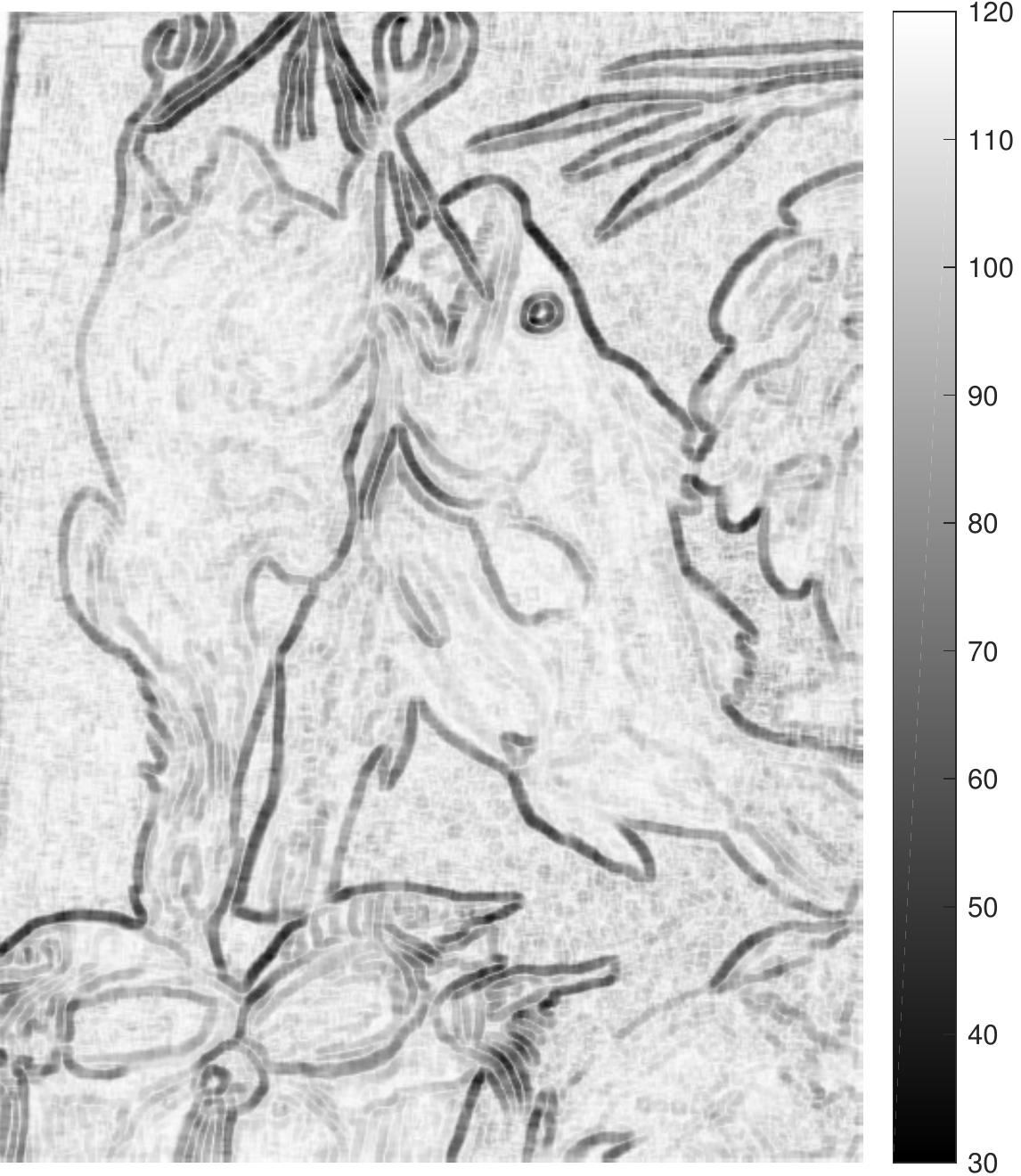}}
    \caption{Texture smoothing using the classical bilateral filter and its adaptive variant. If we use the former, then the range kernel width $\sigma$ needs to be sufficiently small to preserve strong edges; however, as shown in (b), we are unable to smooth out coarse textures as a result. On the other hand, as shown in (c), if we use a large $\sigma$ to smooth out textures, then the edges get blurred. This is evident from the zoomed patches shown in the insets.  As shown in (d), the adaptive bilateral filter achieves both objectives, and this is done by locally adapting the kernel width; the distribution of $\sigma$ is shown in (e). The exact rule for setting $\sigma$ is discussed in Section \ref{sec:app}. We can see some residual textures in (d). As we will see in Section \ref{sec:app}, this can be removed completely by iterating the filter. As a final step, we use the adaptive bilateral filter to sharpen the edges which are inevitably smoothed to some extent. Image courtesy of \cite{Cho2014}.}
    \label{fig:Demo}
\end{figure*}

\subsection{Adaptive bilateral filter} 

In this paper, we will use the filter definition in \cite{Zhang2008}, which is as follows.
Let $f : \I \rightarrow \mathbb{R}$ be the input image, where $\I \subset \mathbb{Z}^2$ is the image domain. 
The output image $g : \I \rightarrow \mathbb{R}$ is given by
\begin{equation}
\label{eq:BF_adaptive1}
g(i) = \eta(i)^{-1} \sum_{j \in \Omega} \omega(j) \phi_i \big( f(i-j)-\theta(i) \big) f(i-j),
\end{equation}
where
\begin{equation}
\label{eq:BF_adaptive2}
\eta(i) = \sum_{j \in \Omega} \omega(j) \phi_i \big( f(i-j)-\theta(i) \big).
\end{equation}
Here $\Omega$ is a window centered at the origin, and $\phi_i : \mathbb{R} \rightarrow \mathbb{R}$ is the local Gaussian range kernel:
\begin{equation}
\label{eq:r_kernel}
\phi_i(t) = \exp \left(-\frac{t^2}{2 \sigma(i)^2} \right).
\end{equation}
Importantly, the center $\theta(i)$ in \eqref{eq:BF_adaptive1} and the width $\sigma(i)$ in \eqref{eq:r_kernel} are spatially varying functions.
The spatial kernel $\omega : \Omega \rightarrow \mathbb{R}$ in \eqref{eq:BF_adaptive1} is Gaussian:
\begin{equation}
\label{eq:sp_kernel}
\omega(j) = \exp \left( -\frac{\lVert j \rVert^2}{2 \rho^2} \right).
\end{equation}
The window is typically set to be $\Omega=[-3\rho,3\rho]^2$. Following \cite{Zhang2008}, we will refer to \eqref{eq:BF_adaptive1} as the \textit{adaptive bilateral filter}.

In the classical bilateral filter, the width of the range kernel is the same at each pixel \cite{Tomasi1998}.
On the other hand, the center $\theta(i)$ is simply the intensity of the pixel of interest $f(i)$.
The center is set to be $\theta(i) = f(i) + \zeta(i)$ in \cite{Zhang2008}, where $\zeta(i)$ is an offset image.
Apart from the applications in \cite{Zhang2008,Zhang2009,Mangiat2011}, the adaptive bilateral filter is naturally more versatile than its non-adaptive counterpart. We demonstrate this using a novel texture filtering application in Section \ref{sec:app}. 
The objective in texture filtering is to remove coarse textures (along with fine details) from the  image, while preserving the underlying structure \cite{Cho2014}. This is illustrated using an example in Figure \ref{fig:Demo}.
As evident from the example, by adapting the width at each pixel (which controls the aggregation), we are simultaneously able to preserve sharp edges and  smooth coarse textures.
This is somewhat difficult to achieve using a fixed kernel (see discussion in Section \ref{sec:app}).

A well-known limitation of the bilateral filter is that it is computation intensive \cite{Paris2009book}.
Needless to say, this also applies for the adaptive bilateral filter.
Considerable work has been done to accelerate the bilateral filter.
However the task of speeding up its adaptive counterpart has received scant attention, if any.
In this paper, we develop a fast algorithm for approximating \eqref{eq:BF_adaptive1} that works with any spatial kernel.
When the spatial kernel is box or Gaussian, the per-pixel complexity of our algorithm is independent of the window size (constant-time algorithm).
In practice, our algorithm can accelerate the brute-force computation by at least $20\times$, without appreciably compromising the filtering quality.

Apart from the bilateral filter, other edge-preserving filters have also been proposed in the literature.
These include the content adaptive bilateral filter \cite{Li2013}, and the guided image filters \cite{He2013,Li2015}.
The former \cite{Li2013} is a variant of the bilateral filter in which the widths of both the spatial and range kernels are allowed to change from pixel to pixel. However, to the best of our knowledge, there is no known fast (e.g., constant-time) algorithm for implementing this filter.
As the name suggests, in the guided filters \cite{He2013,Li2015}, edge-preserving is performed using the edges present in the so-called ``guide'' image.
Similar to our algorithm, both these filters have constant per-pixel complexity w.r.t. the filter size.
These filters have been shown to be useful for a large number of applications \cite{Li2013,He2013,Li2015}. In particular, it is in principle possible to use these filters for the deblocking and texture filtering applications considered  in Section \ref{sec:app}, though its not obvious how one could use then for image sharpening.

We now review existing fast algorithms for classical bilateral filtering, and explain why most of them cannot be used for the adaptive variant.
In particular, we discuss the fast algorithm in  \cite{Mozerov2015} that motivated the present work.

\subsection{Fast bilateral filter}

It follows from \eqref{eq:BF_adaptive1} and \eqref{eq:BF_adaptive2} that the brute-force computation of the classical bilateral filter requires $O(\rho^2)$ operations per pixel, where we recall that $\rho$ is the spatial kernel width.
This makes the real-time implementation challenging when $\rho$ is large, e.g., in applications that require greater aggregation. 
In fact, based on scale-space considerations, one would expect $\rho$ to scale with the image resolution.
To address this, several fast algorithms have been proposed in the literature whose run-time is independent of $\rho$.
The speedup achieved by most of these so-called $O(1)$ algorithms relies on the fact that box and Gaussian convolutions can be computed at fixed cost for any arbitrary $\rho$ \cite{Deriche1993,Young1995}. These algorithms can be broadly classified into four classes as discussed below.

The first class includes the algorithms in \cite{Durand2002} and \cite{Yang2009} that are based on quantization and interpolation.
Here, the intensity range of the input image is quantized into a small set of levels, and exact bilateral filtering is performed for each level using convolutions. The bilateral filtering at an intermediate intensity is approximated by interpolating the exact bilateral filterings. The convolutions are performed on images that are defined for a fixed range kernel.
Consequently, these algorithms cannot be used for adaptive bilateral filtering.

The algorithms in the next class use the so-called shiftable approximation of the range kernel to approximate the bilateral filtering using  convolutions \cite{Porikli2008,Chaudhury2011,Chaudhury2013,Sugimoto2015,Chaudhury2016,Ghosh2016,Nair2017,Papari2017,Ghosh2018}.
In particular, Taylor approximation is used in \cite{Porikli2008}, trigonometric (Fourier) approximation is used in  \cite{Chaudhury2011,Chaudhury2013,Sugimoto2015,Nair2017}, and a Gaussian-weighted polynomial approximation is used in \cite{Chaudhury2016}. The shiftable approximation in \cite{Papari2017} is derived using eigendecomposition.
These algorithms are fundamentally based on the decomposition of a fixed range kernel. As a result, they cannot be used to speed up \eqref{eq:BF_adaptive1}.

A third set of algorithms use high-dimensional convolutions for fast bilateral filtering \cite{Paris2006,Dai2016}.
They are based on the observation that by concatenating the spatial and range dimensions, the bilateral filter can be expressed as a convolution in this higher dimensional space. However, if the range kernel changes at every pixel, the high-dimensional filtering cannot be expressed as a convolution. Therefore, these methods also fail for our case of interest.

The fourth class comprises of histogram-based algorithms.
Here the bilateral filtering is expressed as an operator acting on the local histogram.
It was observed in \cite{Boomgaard2002,Koenderink1999,vanDeWeijer2001} that the bilateral filter is in fact related to other non-linear filters that operate on local intensity histograms.
For box (uniform) spatial kernels, an $O(1)$ algorithm based on integral histograms was first proposed in \cite{Porikli2008}.
Later, it was shown in \cite{Zhang2012} how this can be extended to Gaussian spatial kernels using a combination of box kernels.
A similar approach was advocated in \cite{Pan2014} for arbitrary spatial kernels.
More recently, a novel approach based on imposing a prior on the local histogram was proposed by Mozerov and van de Weijer \cite{Mozerov2015}.
We will focus on this approach and show how it can be extended for adaptive bilateral filtering. First, we review its salient features.

\subsection{Speedup using histogram approximation}

The spatial kernel is included in the specification of the histogram in \cite{Mozerov2015}, along with the frequency of local intensity values.
As a result, the idea of using local histograms can be generalized to arbitrary spatial kernels, beyond just box kernels \cite{Porikli2008}.
By approximating the histogram using an uniform distribution, the authors were able to express the filtering in terms of the standard error function.
This resulted in a fast algorithm which was eventually used  for bilateral filtering of color images.
An interesting aspect of this method is that, unlike other histogram based methods, it does not involve any form of kernel approximation;  the approximation is purely of the local histogram.
The approximation process involves just one convolution with the spatial kernel, which does not involve the range kernel parameters.
The parameters only appear in the standard error function mentioned earlier, which is evaluated at each pixel.
Therefore, the formal structure of the algorithm remains unaffected if we adapt the parameters.
In particular, we will show how a fast and accurate algorithm for adaptive bilateral filtering can be developed by refining the approximation in \cite{Mozerov2015}.

\subsection{Contributions}

Our primary contribution is to develop the core idea in \cite{Mozerov2015} to formulate a fast algorithm for adaptive bilateral filtering.
We propose a computational framework for improving the histogram approximation in \cite{Mozerov2015}, while ensuring that the resulting algorithm is significantly faster than the brute-force computation. Moreover, we also demonstrate the utility of the proposed algorithm for applications such as image sharpening, JPEG deblocking, and texture filtering.

On the technical front, we demonstrate that a uniform prior on the histogram (as used in \cite{Mozerov2015}) results in a poor approximation.
Indeed, one would expect the intensity distribution around an edge, or in a textured region, to be far from uniform.
This prompted us to model the local histogram using polynomials.
A computational requirement arising from this choice is to approximate the local histogram without computing it explicitly.
We propose to do this by matching the moments of the polynomial to those of the histogram.
The advantage with this proposal is that the moments of the histogram can be computed using convolutions.
The moment matching problem reduces to inverting a linear system $\mathbf{A}\boldsymbol{x}=\boldsymbol{b}$, where both $\mathbf{A}$ and $\boldsymbol{b}$ vary from pixel to pixel.
This is of course computationally prohibitive.
However, we show that it suffices to consider a single $\mathbf{A}$ by appropriately transforming the local range data.
In fact, this particular $\mathbf{A}$ turns out to be a well-studied matrix, whose inverse has a closed-form expression.
Similar to \cite{Mozerov2015}, we are able to approximate \eqref{eq:BF_adaptive1} and \eqref{eq:BF_adaptive2} using a series of definite integrals. In fact, the integrals are related to each other, and we devise a fast recursion using integration-by-parts.
An important distinction from \cite{Mozerov2015} is that we are able to efficiently compute the limits of integration using an existing fast algorithm. 
The limits are set in a somewhat heuristic fashion in \cite{Mozerov2015}. As we will show later, setting the limits correctly can help improve the approximation accuracy. 

The theoretical complexity of the proposed algorithm is $O(1)$ with respect to $\rho$.
In fact, the speedup over the brute-force implementation is quite significant in practice. 
For the applications in Section  \ref{sec:app}, we can obtain at least $20\times$ acceleration (this can go up to $60\times$), while ensuring an acceptable approximation accuracy (PSNR $\geq 40$ dB, \cite{Porikli2008,Mozerov2015}). 
To the best of our knowledge, this is the first $O(1)$ algorithm for adaptive bilateral filtering reported in the literature.

\subsection{Organization}

The remaining paper is organized as follows.
The proposed approximation of the adaptive bilateral filter and the resulting fast algorithm are developed in Section \ref{sec:method}.
In Section \ref{sec:comp}, we compare our method with \cite{Mozerov2015} and the brute-force implementation in terms of timing and approximation accuracy.
In Section \ref{sec:app}, we apply our algorithm for image sharpening, compression artifact removal, and texture filtering.
We conclude the paper with a discussion in Section \ref{sec:con}.

\section{Analytic Approximation}
\label{sec:method}

\subsection{Main idea}

It was observed in \cite{Mozerov2015} that the classical bilateral filter can be expressed as a weighted average of local pixel intensities. We adopt the same idea for the adaptive bilateral filter. More specifically, for $i \in \I$, let
\begin{equation*}
\Lambda_i = \big\{f(i-j) : j \in \Omega \big\}.
\end{equation*}
For $t \in \Lambda_i$, define the (weighted) histogram 
\begin{equation}
\label{hist}
h_i(t) = \sum_{j \in \Omega} \omega(j) \delta \big(f(i-j)-t \big),
\end{equation}
where $\delta$ is the Kronecker delta, namely, $\delta(0) = 1$, and $\delta(t) = 0$ for $t \neq 0$. Note that we can write \eqref{eq:BF_adaptive1} and \eqref{eq:BF_adaptive2} in terms of \eqref{hist}:
\begin{equation}
\label{BF1hist}
g(i) = \eta(i)^{-1} \sum_{t \in \Lambda_i} t h_i(t) \phi_i \big( t-\theta(i) \big),
\end{equation}
and
\begin{equation}
\label{BF2hist}
\eta(i) = \sum_{t \in \Lambda_i} h_i(t) \phi_i \big( t-\theta(i) \big).
\end{equation}
Consider the bounds on $\Lambda_i$,
\begin{equation}
\label{bounds}
\alpha_i = \min \ \{t : t \in \Lambda_i\} \quad \text{and} \quad \beta_i  = \max \ \{t : t \in \Lambda_i\}.
\end{equation}
Note that if $\alpha_i = \beta_i$, that is, if the neighborhood of $i$ contains exactly one intensity value, then $g(i) = f(i)$, and hence no processing is required. For the rest of the discussion, we will assume that $\alpha_i \neq \beta_i$, unless stated otherwise. Our proposal is to approximate $h_i$ (which has a finite domain) with a function $p_i$ defined on the interval $[\alpha_i,\beta_i]$, and to replace the finite sums in \eqref{BF1hist} and \eqref{BF2hist} with integrals. In other words, we consider the following approximations of \eqref{BF1hist} and \eqref{BF2hist}: 
\begin{equation}
\label{BFapp}
\hat{g}(i)=\hat{\eta}(i)^{-1}  \int_{\alpha_i}^{\beta_i} t p_i(t) \phi_i \big( t - \theta(i) \big) \ dt,
\end{equation}
and
\begin{equation*}
\hat{\eta}(i) = \int_{\alpha_i}^{\beta_i} p_i(t) \phi_i \big(t - \theta(i) \big) \ dt.
\end{equation*}
At first sight, it appears that these integrals are more difficult to compute than the finite sums in \eqref{BF1hist} and \eqref{BF2hist}. However, notice that if $p_i$ is a polynomial, say,
\begin{equation}
\label{poly}
p_i(t) = c_0 + c_1 t + \cdots + c_N t^N,
\end{equation}
then we can write \eqref{BFapp} as
\begin{equation}
\label{app1}
\hat{g}(i)=\hat{\eta}(i)^{-1} \sum_{n=0}^N c_n   \int_{\alpha_i}^{\beta_i} t^{n+1} \phi_i \big( t - \theta(i) \big) \ dt,
\end{equation}
where
\begin{equation}
\label{app2}
\hat{\eta}(i)= \sum_{n=0}^N c_n   \int_{\alpha_i}^{\beta_i} t^{n} \phi_i \big(t - \theta(i)\big) \ dt.
\end{equation}
The integrals in \eqref{app1} and \eqref{app2} are of the form:
\begin{equation}
\label{int1}
\int_a^b t^n \exp \big( -\lambda_i (t-s_i)^2 \big) dt.
\end{equation}
When $n=0$, we can compute \eqref{int1} using just two evaluations of the standard error function \cite{Press2007}.
When $n \geq 1$, then the integrand in \eqref{int1} is a product of a monomial and a Gaussian.
We will show in Section \ref{sec:AA::Integrals}, how we can efficiently  compute \eqref{int1} for $1 \leq n \leq N$ using recursions.

In summary, when $p_i$ is a polynomial, we can express \eqref{BFapp} using integrals that can be computed efficiently.
The number of integrals depend on the degree $N$, but do not scale with the width $\rho$ of the spatial filter.
We will later demonstrate that reasonably good approximations of \eqref{eq:BF_adaptive1} and \eqref{eq:BF_adaptive2} can be obtained using low-degree polynomials.

\subsection{Polynomial fitting}
\label{sec:AA::PolyFit}

\begin{figure}[t!]
	\centering
	\subfloat[]{\includegraphics[height=0.40\linewidth,keepaspectratio]{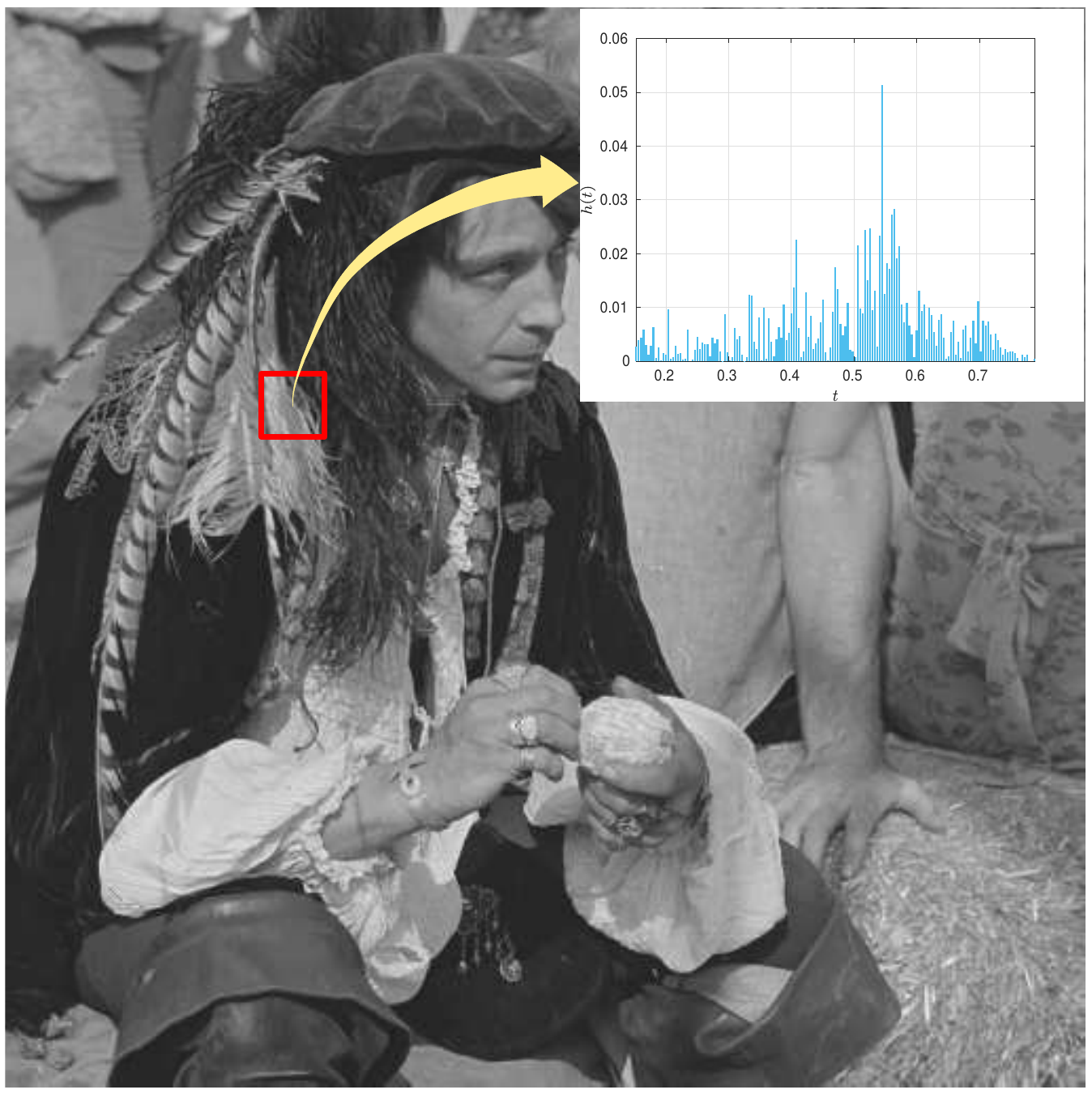}}
	\hspace{2mm}
	\subfloat[]{\includegraphics[height=0.40\linewidth,width=0.49\linewidth]{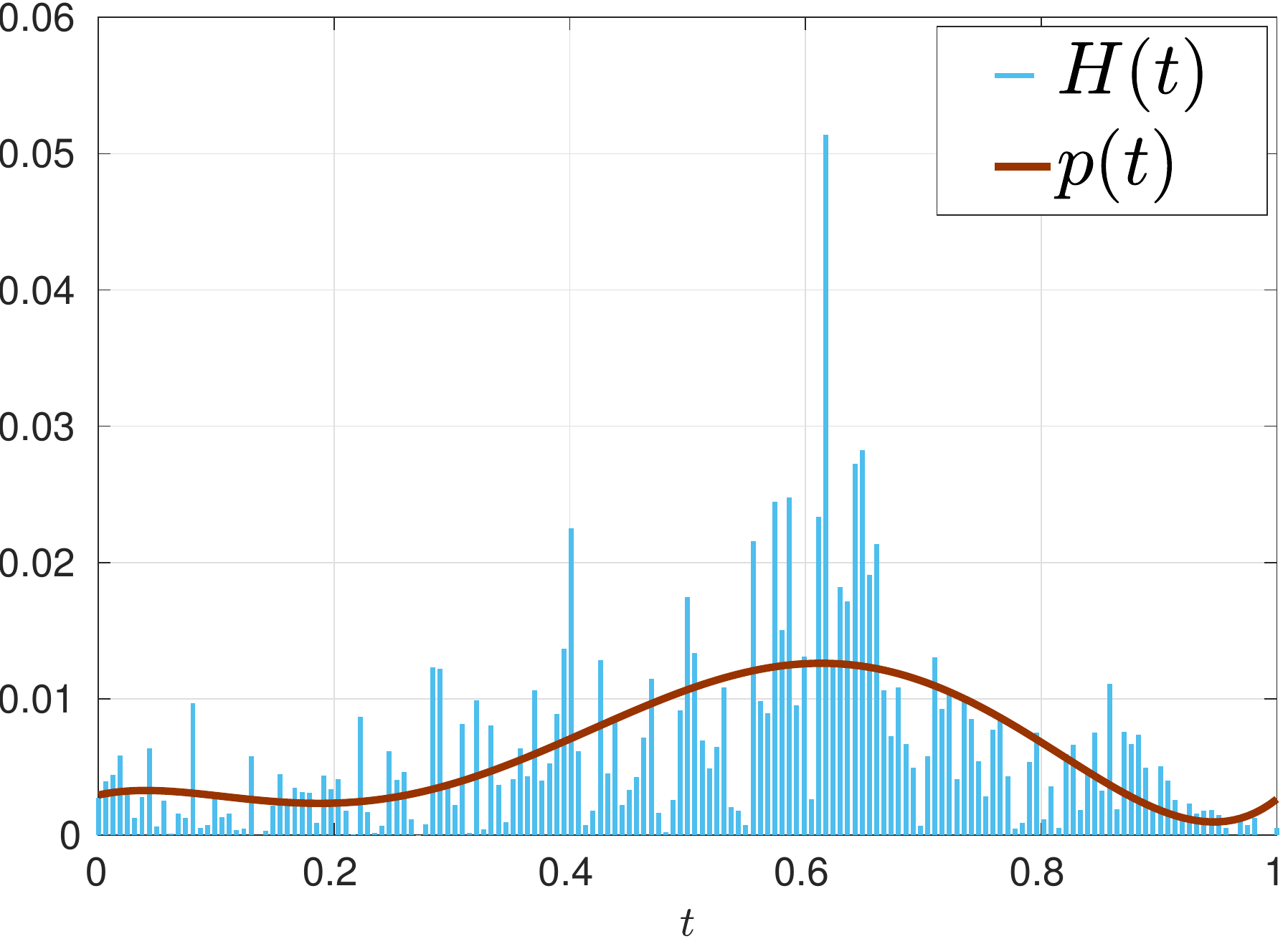}}
	\caption{Illustration of domain stretching and polynomial fitting. (a) Grayscale image with intensity range $[0,1]$. The histogram $h(t)$ for a $31 \times 31$ window (highlighted in red) is shown in the inset, where $t$ takes values in $[0.15,0.79]$. (b) The histogram $H(t)$ after affinely mapping the domain $[0.15,0.79]$ to $[0,1]$. Also shown is a fifth-degree polynomial $p(t)$ whose coefficients are obtained using moment matching (scaled for display purpose).}
	\label{fig:Histogram}
\end{figure}

We now provide technical details as to how the polynomial fitting can be performed in an efficient manner. Assume that a polynomial of degree $N$ is used to approximate the histogram $h_i$ at some  $i \in \I$. In particular, 
let the polynomial be given by \eqref{poly}, and we wish to approximate \eqref{hist} using this polynomial. To estimate the coefficients $c_0,\ldots,c_N$, we propose to match the first $N+1$ moments of $h_i$ and $p_i$ \cite{Munkhammar2017}. That is, we wish to find $c_0,\ldots,c_N$ such that, for $0 \leq k \leq N$,
\begin{equation}
\label{match}
m_k = \int_{\alpha_i}^{\beta_i} t^k p_i(t) \ dt = \sum_{t \in \Lambda_i} t^k h_i(t).
\end{equation}
Ideally, $p_i(t)$ should be a density function, i.e., a non-negative function with unit integral. However, since $p_i(t)$ would appear in the numerator and denominator of \eqref{BF1hist}, the latter normalization is not required. On the other hand, we simply choose to drop the non-negativity requirement; in fact, the approximation is empirically non-negative in most cases (see Figure \ref{fig:Histogram}).

Let $\m= (m_0,\ldots,m_N)$ and $\c=(c_0,\ldots,c_N)$. Then we can write \eqref{match} as 
\begin{equation}
\label{eq:lin}
\A \c =  \m,
\end{equation}
where $\A = \{\A_{m,n}: 1 \leq m,n\leq N+1\}$ is given by
\begin{equation}
\label{Hmn}
\A_{m,n} = \frac{1}{m+n-1} \big( \beta_i^{m+n-1} - \alpha_i^{m+n-1} \big).
\end{equation}
Needless to say, $\A$, $\c,$ and $\m$ depend on $i$, but we will avoid writing this explicitly to keep the notations simple.

Notice that the brute-force computation of $\alpha_i$ and $\beta_i$ given by \eqref{bounds}, and the moments $m_1,\ldots,m_N$, requires $O(\rho^2)$ operations. This is in fact the complexity of the filter that we set out to approximate. Fortunately, it turns out that there are faster ways of computing them.
\begin{proposition}
\label{prop:AlphaBeta}
There exists an algorithm for computing the bounds in \eqref{bounds} that has $O(1)$ complexity with respect to $\rho$.
\end{proposition}
In particular, the bounds can be computed by adapting the MAX-FILTER algorithm in \cite{Chaudhury2013} that has $O(1)$ complexity. This  was originally designed to compute the local maximum at each pixel, but it can trivially be modified to compute the local minimum. 

We next generalize an observation from \cite{Mozerov2015}, which allows us to efficiently compute the moments (see Appendix \ref{proof1} for the detailed argument).

\begin{proposition}
\label{prop:Moments}
For fixed $0 \leq k \le N$, the moments 
\begin{equation}
\label{hist_moments}
\sum_{t \in \Lambda_i} t^k h_i(t) \qquad (i \in \I)
\end{equation}
can be computed with $O(1)$ complexity with respect to $\rho$.
\end{proposition}

As a consequence of the above observations, we can compute $\A$ and $\m$ at every $i \in \I$ with $O(1)$ complexity. However, we still need to perform one matrix inversion at each pixel, which in fact would be slower than computing \eqref{eq:BF_adaptive1} and \eqref{eq:BF_adaptive2}. Fortunately, there is a simple solution to this problem. This is based on the observation that if $\alpha_i = 0$ and $\beta_i = 1$, then \eqref{Hmn} becomes
\begin{equation}
\label{HM}
\A_{m,n} = \frac{1}{m+n-1}.
\end{equation}
This is called the Hilbert matrix. Importantly, the inverse of the Hilbert matrix admits a closed-form expression \cite{Choi1983}.
\begin{theorem}
The Hilbert matrix \eqref{HM} is invertible and its inverse is given by
\begin{equation}
\label{inv}
\begin{aligned}
& (\A^{-1})_{m,n} = (-1)^{m+n} (m+n-1) \cdot \\
& \binom{N+m}{N+1-n} \binom{N+n}{N+1-m} \binom{m+n-2}{m-1}^2.
\end{aligned}
\end{equation}
\end{theorem}

In summary, if the domain $\Lambda_i$ of the local histogram can be stretched from $[\alpha_i, \beta_i]$ to $[0,1]$, then the moment matching problem at every $i \in \I$ can be solved using a single closed-form inversion. Now, the domain stretching can be done using  an affine transformation. In particular, define
\begin{equation}
\label{Lambda_tilde}
\Gamma_i = \left\{ \frac{t-\alpha_i}{\beta_i-\alpha_i} : t \in \Lambda_i \right\} \subseteq [0,1].
\end{equation}
Note that $\Gamma_i$ is a shifted and scaled copy of $\Lambda_i$. Next, define a copy of the original histogram on the new domain:
\begin{equation}
\label{mhist}
H_i(t) = h_i \big( \alpha_i + t(\beta_i - \alpha_i) \big) \qquad (t \in \Gamma_i).
\end{equation}
Let $\mu_k$ be the $k$-th moment of $H_i$:
\begin{equation}
\label{hist_moments_new}
\mu_k= \sum_{t \in \Gamma_i} t^k H_i(t).
\end{equation}
As discussed previously, we approximate $H_i$ using a polynomial $p_i$ on $[0,1]$. In particular, if we set
\begin{equation}
\label{polynom}
p_i(t) = c_0 + c_1 t + \cdots + c_N t^N,
\end{equation}
where 
\begin{equation*}
\begin{bmatrix}
           c_0 \\
           c_1 \\
           \vdots \\
           c_N
         \end{bmatrix}
= \A^{-1}
\begin{bmatrix}
           \mu_0 \\
           \mu_1 \\
           \vdots \\
           \mu_N
         \end{bmatrix},
\end{equation*}
and $\A^{-1}$ is given by \eqref{inv}, then the moments of $H_i$ and $p_i$ are equal. Moreover, we make the following observation concerning the moments of $H_i$ (see Appendix \ref{proof2}).
\begin{proposition}
\label{prop:mhist_moments}
For fixed $i \in \I$,  we can compute $(\mu_k)$ from $(m_k)$. In particular, $\mu_0=m_0$, and for $1 \leq k \le N$,
\begin{equation}
\label{mhist_moments}
\mu_k = \left(\beta_i-\alpha_i\right)^{-k} \sum_{r=0}^k \binom{k}{r} (-\alpha_i)^{k-r} m_r.
\end{equation}
\end{proposition}
While the original formulations \eqref{BF1hist} and \eqref{BF2hist} are in terms of $h_i$,
we can also express them in terms of $H_i$. Before doing so, it will be convenient to define for each $i \in \I$, the following parameter and kernel:
\begin{equation}
\label{sub}
\lambda_i = \big(2 \sigma(i)^2 \big)^{-1} (\beta_i - \alpha_i)^2
\ \text{ and } 
\ \psi_i(t) = \exp \left( - \lambda_i t^2 \right).
\end{equation}
Moreover, define $\theta_0 : \I \rightarrow \mathbb{R}$ to be
\begin{equation*}
\theta_0(i) = \big( \beta_i-\alpha_i \big)^{-1} \big( \theta(i)-\alpha_i \big).
\end{equation*}
The following result allows us to express the original filter \eqref{eq:BF_adaptive1} in terms of the above reductions (the derivation is provided in  Appendix \ref{proof3}).
\begin{proposition}
\label{prop:BFmhist}
Define $F: \I \rightarrow \mathbb{R}$ to be
\begin{equation}
\label{ftilde}
F(i) = \frac{\sum_{t \in \Gamma_i} t H_i(t) \psi_i \big( t-\theta_0(i) \big)}{\sum_{t \in \Gamma_i} H_i(t) \psi_i \big( t-\theta_0(i) \big)}. \\
\end{equation}
Then, for $i \in \I$,
\begin{equation}
\label{BFmhist}
g(i) = \alpha_i + ( \beta_i - \alpha_i) F(i).
\end{equation}
\end{proposition}
In short, we first rescale the original histogram, then average using the new histogram, and finally rescale the output to get back the original result.  
As described earlier, we approximate $H_i$ using \eqref{polynom}, and replace the sums with integrals in \eqref{ftilde}. More precisely, we approximate the numerator and denominator 
of \eqref{ftilde} using
\begin{equation*}
\int_0^1 \! t p_i(t) \psi_i \big( t-\theta_0(i) \big) \ dt = \sum_{k=0}^N c_k \! \int_0^1 \! t^{k+1} \psi_i \big( t-\theta_0(i) \big) \ dt ,
\end{equation*}
and
\begin{equation*}
\int_0^1 \! p_i(t) \psi_i \big( t-\theta_0(i) \big) \ dt = \sum_{k=0}^N c_k \! \int_0^1 \! t^k \psi_i \big( t-\theta_0(i) \big)\ dt,
\end{equation*}
where we recall that $\psi_i$ is given by \eqref{sub}. In other words, for fixed $i \in \I$, we are required to compute integrals of the form
\begin{equation}
\label{integrals}
I_k = \int_0^1 t^k \exp \left( -\lambda (t-t_0)^2 \right) dt,
\end{equation}
for $k=0,\ldots,N+1$, where $t_0=\theta_0(i)$ and $\lambda = \lambda_i$.

In terms of \eqref{integrals}, the proposed approximation of \eqref{eq:BF_adaptive1}, $\hat{g} : \I \rightarrow \mathbb{R}$, is defined as follows: If $\alpha_i = \beta_i$, then 
$\hat{g}(i)=f(i)$, else
\begin{equation}
\label{eq:BFfinal}
\hat{g}(i) = \alpha_i + (\beta_i-\alpha_i) \frac{c_0 I_1+\cdots+c_{N}I_{N+1}} {c_0 I_0+\cdots+c_N I_N}.
\end{equation}
As before, it is understood that $t_0$, $\lambda,$ and $I_k$ depend on $i$.

In Figure \ref{fig:Histogram}, we explain the domain stretching and polynomial fitting procedures using an example.

\subsection{Recursive computation of integrals}
\label{sec:AA::Integrals}

We now explain how $I_0,\ldots,I_{N+1}$ can be efficiently computed using recursions.
Notice that ratio of the integrands of $I_k$ and $I_{k+1}$ is $t$, which (up to an affine scaling) is the derivative of the argument in the exponential. 
More specifically, consider the following equation: 
\begin{align*}
& \frac{d}{dt}  \Big( t^{k-1}  \exp \big( -\lambda (t-t_0)^2 \big) \Big) \\
& =  \left(k-1+ 2 \lambda t_0 t  - 2 \lambda t^2 \right) t^{k-2} \exp \left( -\lambda (t-t_0)^2 \right).
\end{align*}
By integrating this over $[0,1]$, we arrive at the relation
\begin{equation*}
2 \lambda I_k = 2 \lambda t_0 I_{k-1} + (k-1) I_{k-2} -\exp \left( -\lambda (1-t_0)^2 \right) .
\end{equation*}
This gives us the following recursive formula:
\begin{equation}
\label{eq:F_forward}
I_k = t_0 I_{k-1} + \frac{k-1}{2 \lambda} I_{k-2} - \frac{1}{2 \lambda} \exp \left(-\lambda (1-t_0)^2 \right).
\end{equation}
We can use this to compute $I_2,\ldots,I_{N+1}$ starting with $I_0$ and $I_1$. However,
\begin{equation*}
I_0 = \int_0^1 \exp \left( -\lambda (t-t_0)^2 \right) \ dt,
\end{equation*}
which can be computed using two evaluations of the standard error function \cite{Press2007}:
\begin{equation}
\label{eq:I_0}
I_0 = \frac{1}{2} \sqrt{\frac{\pi}{\lambda}} \left[ \mathrm{erf}\big( \sqrt{\lambda}(1-t_0) \big) - \mathrm{erf}\big( -\sqrt{\lambda} t_0 \big) \right].
\end{equation}
On the other hand, 
\begin{align}
& I_1 = \int_0^1  t \exp \left( -\lambda (t-t_0)^2 \right) \ dt \nonumber \\
& = t_0 I_0 + \frac{1}{2\lambda} \Big( \exp \left( -\lambda t_0^2 \right) -\exp \left( -\lambda (1-t_0)^2 \right) \Big). \label{eq:I_1}
\end{align}
In summary, we first compute $I_0$ and $I_1$ using \eqref{eq:I_0} and \eqref{eq:I_1}, and then we use \eqref{eq:F_forward} to compute $I_2,\ldots,I_{N+1}$.

\subsection{Algorithm and implementation}
\label{imp}

\begin{algorithm}[t!]
	\label{alg:Proposed}
	\DontPrintSemicolon
	\KwIn{Image $f:\I \rightarrow \mathbb{R}$; parameters $\sigma : \I \rightarrow \mathbb{R},\ \theta : \I \rightarrow \mathbb{R},\ \rho$ and $N$.}
	\KwOut{Image $\hat{g}:\I \rightarrow \mathbb{R}$ given by \eqref{eq:BFfinal}.}
	Set $\omega(j)$ using \eqref{eq:sp_kernel}. \;
	Compute $\{\alpha_i ,\beta_i : i \in \I\}$. \label{line:bounds} \;
	Populate $\A^{-1}$ using \eqref{inv}. \;
	Set $\I_0 = \{i \in \I:\alpha_i \neq \beta_i\}$.\; \label{line:threshold}
	For $i \notin \I_0$, set $\hat{g}(i) = f(i)$.\;
	Compute  $\gamma_k = \omega \ast f^k$ for $k=0,\ldots,N$.\; \label{line:conv}
	\For{$i \in \I_0$}
	{
		$m_k =\gamma_k(i)$ for $k=0,\ldots,N$.\;
		Compute $(\mu_k)$ from $(m_k)$ using \eqref{mhist_moments}. \; 
		$\boldsymbol{\mu} = (\mu_0,\ldots,\mu_N)$.\;
		$\c = \A^{-1} \boldsymbol{\mu}$. \label{line:linear} \;
		$t_0 = (\theta(i)-\alpha_i)/(\beta_i-\alpha_i)$. \;
		$\lambda = 0.5 (\beta_i-\alpha_i)^2/\sigma(i)^2$. \;
		Compute $I_0,I_1$ using \eqref{eq:I_0} and \eqref{eq:I_1}. \label{line:I_direct} \;
		$T_1 = c_0 I_1$. \;
		$T_2 = c_0 I_0$. \;
		\For{$k = 2,\ldots,N+1$}
		{
			Compute $I_k$ using \eqref{eq:F_forward}. \label{line:I_recursive} \;
			$T_1 = T_1 + c_{k-1} I_{k}$. \;
			$T_2 = T_2 + c_{k-1} I_{k-1}$. \;
		}
		$\hat{g}(i) = \alpha_i + (\beta_i-\alpha_i) (T_1/T_2)$. \;
	}
	\caption{Fast adaptive bilateral filtering.}
\end{algorithm}

The steps in the proposed approximation are summarized in Algorithm \ref{alg:Proposed}. 
The computation in line \ref{line:bounds} is performed using the algorithm in \cite{Chaudhury2013}, which is $O(1)$ and very fast.
The main computations are essentially in lines \ref{line:conv}, \ref{line:linear}, \ref{line:I_direct}, and \ref{line:I_recursive}.
Note that $\gamma_0(i) = 1$ for all $i$, hence we need just $N$ convolutions to compute $(\gamma_k)$ in line \ref{line:conv}.
As we shall see later, it is sufficient to set $N=5$ for most applications.
The convolutions can be performed using a recursive $O(1)$ Gaussian filter \cite{Deriche1993,Young1995}.
However, we used the Matlab  inbuilt ``imfilter'' which is faster than our implementation of \cite{Young1995} for practical values of $\rho$ (though its run-time scales with $\rho$).
In fact, the Gaussian convolutions dominate the total runtime of the algorithm.

Recall that in line \ref{line:linear} we do not need to actually invert $\A$ since a closed-form formula is available.
Thus only a matrix-vector multiplication is required.
In fact, since $\A^{-1}$ is common for all pixels, we can compute $\c$ for all $i$ in a single step by multiplying $\A^{-1}$ with the matrix containing the values of $\boldsymbol{\mu}$ for all $i$.
This is precisely the advantage of shifting and scaling the histogram domain as discussed in Section \ref{sec:AA::PolyFit}.

To compute $I_0$ (line \ref{line:I_direct}) we need just two evaluations of the error function, whereas for $I_1$ we need two evaluations of the exponential function.
Further, note that the exponential in the third term in  \eqref{eq:F_forward} is already computed in \eqref{eq:I_1} for $I_1$, and hence need not be computed again.
On the overall, we need just two evaluations of the error and the exponential functions, i.e., four transcendental function evaluations per pixel
(this should be compared with the brute-force implementation, which requires $O(\rho^2)$ transcendental function evaluations per pixel).
Thus, as claimed earlier, our algorithm has $O(1)$ complexity w.r.t. $\rho$.

\section{Comparisons And Discussion}
\label{sec:comp}

We recall that there are no existing algorithms in the literature for fast adaptive bilateral filtering. A relevant comparison is that between Algorithm \ref{alg:Proposed} and the brute-force implementation of \eqref{eq:BF_adaptive1}.
A practical problem in this regard is that we need rules to set the center $\theta(i)$ and width $\sigma(i)$ at every pixel.
Of course, these rules will depend on the application at hand.
In Section \ref{sec:app}, we describe three applications of adaptive bilateral filtering, where these rules will be provided.
We will compare the approximation accuracy of our fast algorithm with the brute-force implementation at that point.

On the other hand, to demonstrate the acceleration achieved by the proposed algorithm, we can use any values for $\theta(i)$ and $\sigma(i)$. In particular, we will demonstrate the speedup obtained using Algorithm \ref{alg:Proposed} for the classical bilateral filter, where we set $\theta(i) = f(i)$ and $\sigma(i)$ to a fixed value $\sigma_0$ for all $i$. 

As mentioned earlier, the present fast algorithm was inspired by the histogram-based approximation in \cite{Mozerov2015}. 
For completeness, we compare Algorithm \ref{alg:Proposed} with \cite{Mozerov2015} for classical bilateral filtering.
This also gives us a particularly simple way to assess the accuracy of the proposed algorithm.
In particular, we will show that our algorithm is more accurate (in terms of PSNR) compared to \cite{Mozerov2015}. 

To objectively compare two images, we use the peak signal-to-noise ratio (PSNR).
Following \cite{Durand2002,Yang2009,Zhang2012,Mozerov2015}, we measure the approximation accuracy of Algorithm \ref{alg:Proposed} by observing the PSNR between the exact (brute-force) and approximate filterings (i.e. $\hat{g}$ and $g$ in Section \ref{sec:method}).
The approximation is considered satisfactory if the PSNR is at least $40$ dB \cite{Porikli2008,Zhang2012,Mozerov2015}.

\begin{figure}[t!]
	\raggedright
	\hspace{0.055\linewidth}
	\subfloat[Input ($512 \times 512$).]{\includegraphics[height=0.40\linewidth,keepaspectratio]{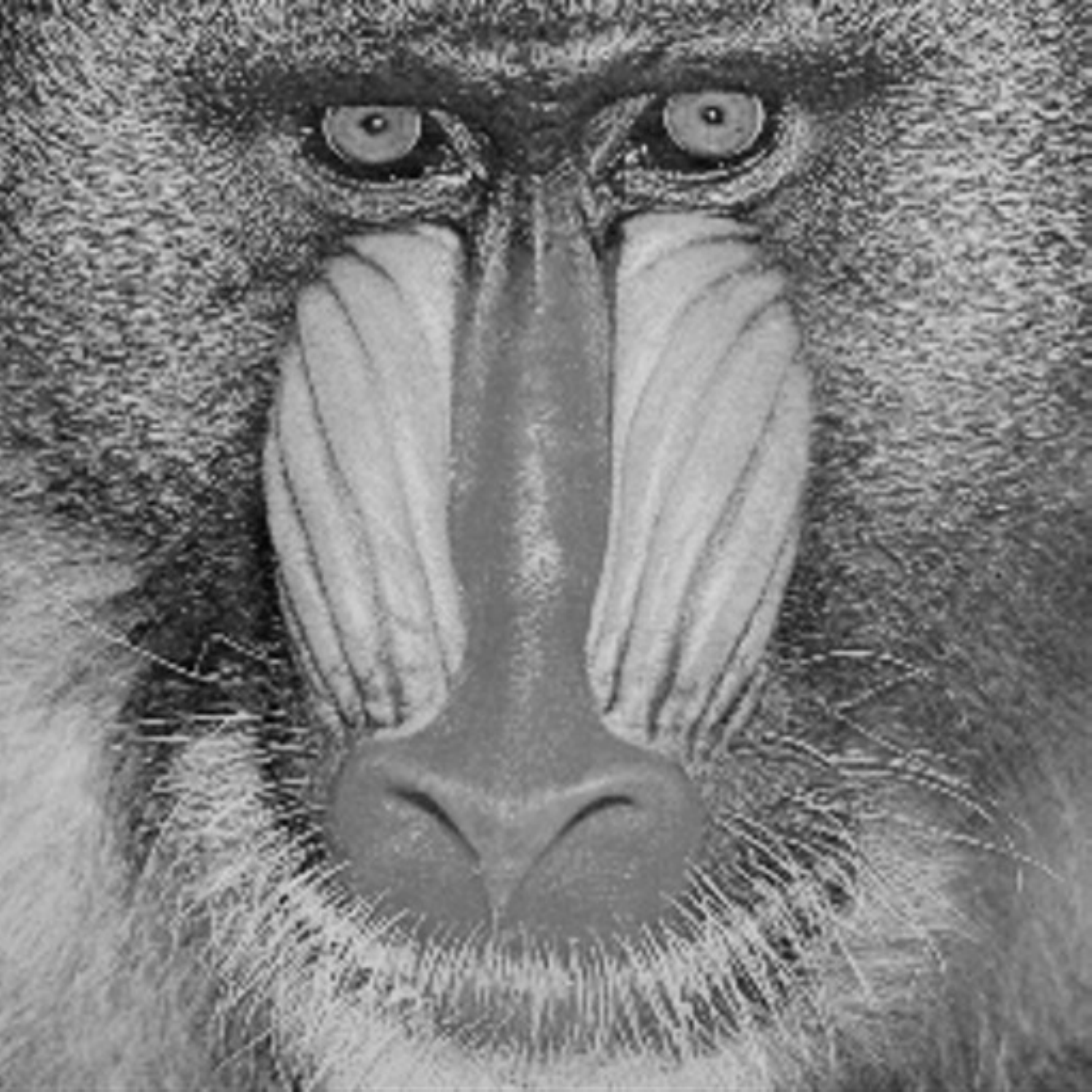}}
	\hspace{0.1mm}
	\subfloat[Brute-force filtering.]{\includegraphics[height=0.40\linewidth,keepaspectratio]{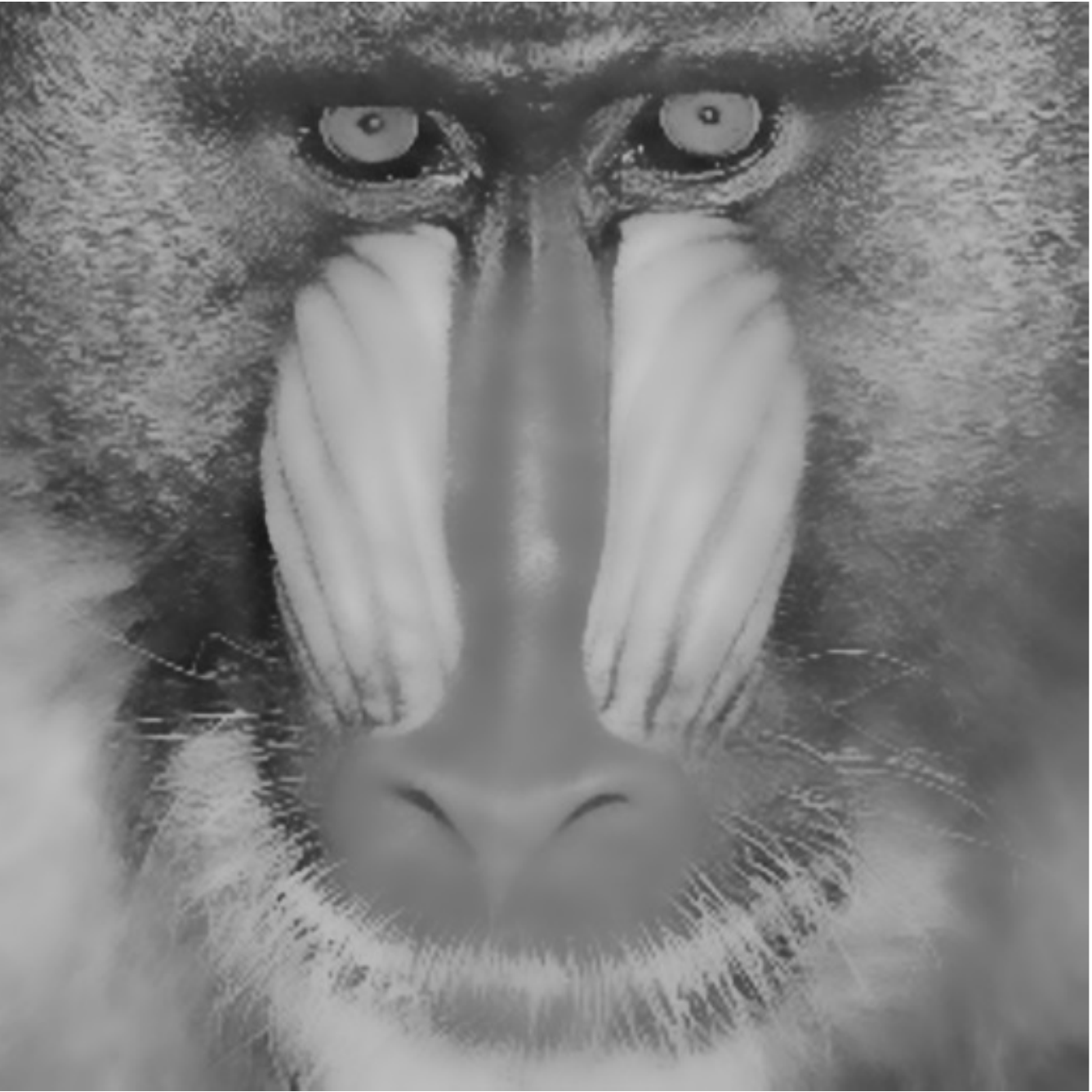}} \\
	\centering
	\subfloat[$N=2$, $40.83$ dB.]{\includegraphics[height=0.40\linewidth,keepaspectratio]{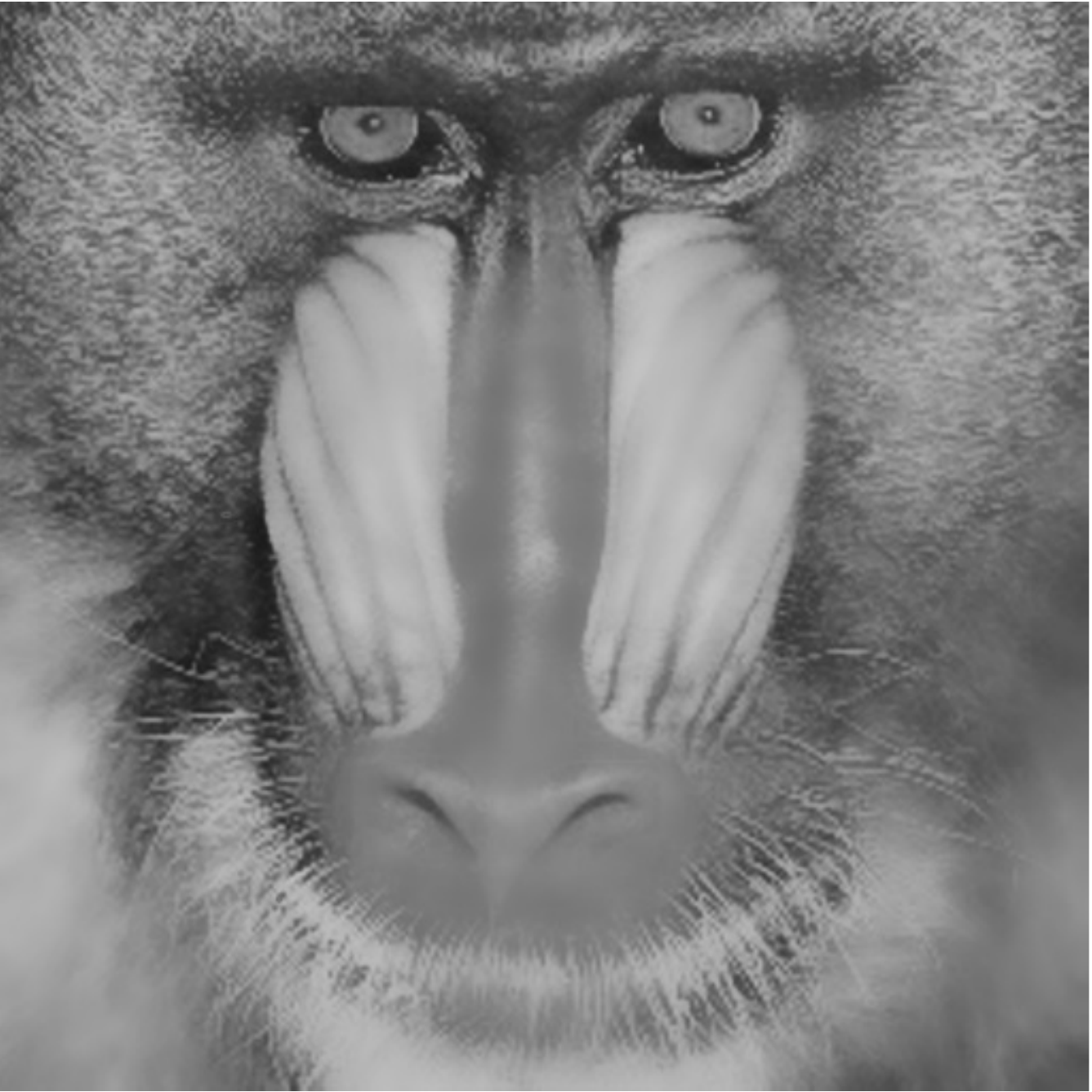}}
	\hspace{0.1mm}
	\subfloat[Error.]{\includegraphics[height=0.40\linewidth,keepaspectratio]{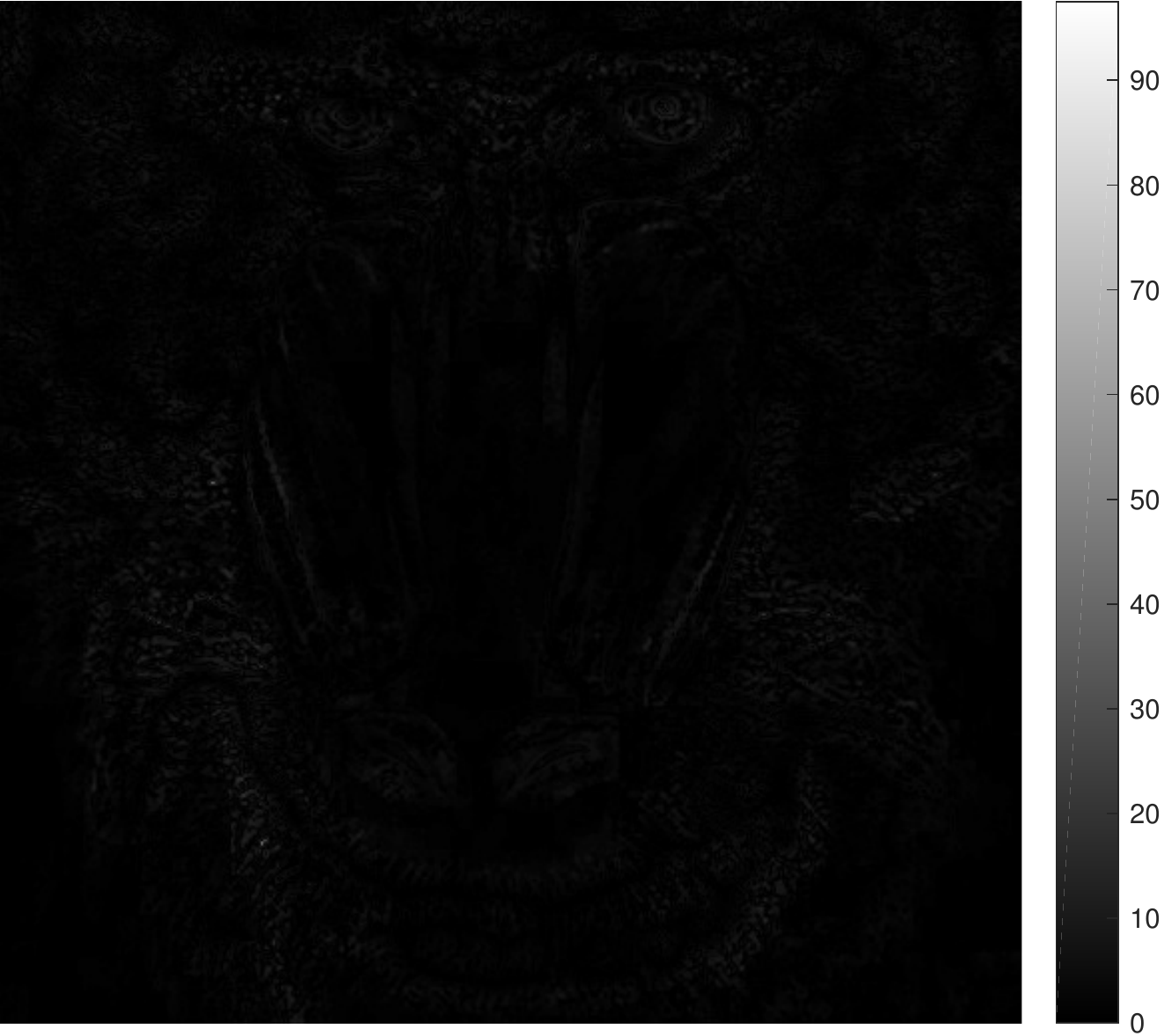}} \\
	\subfloat[$N=0$, $28.15$ dB.]{\includegraphics[height=0.40\linewidth,keepaspectratio]{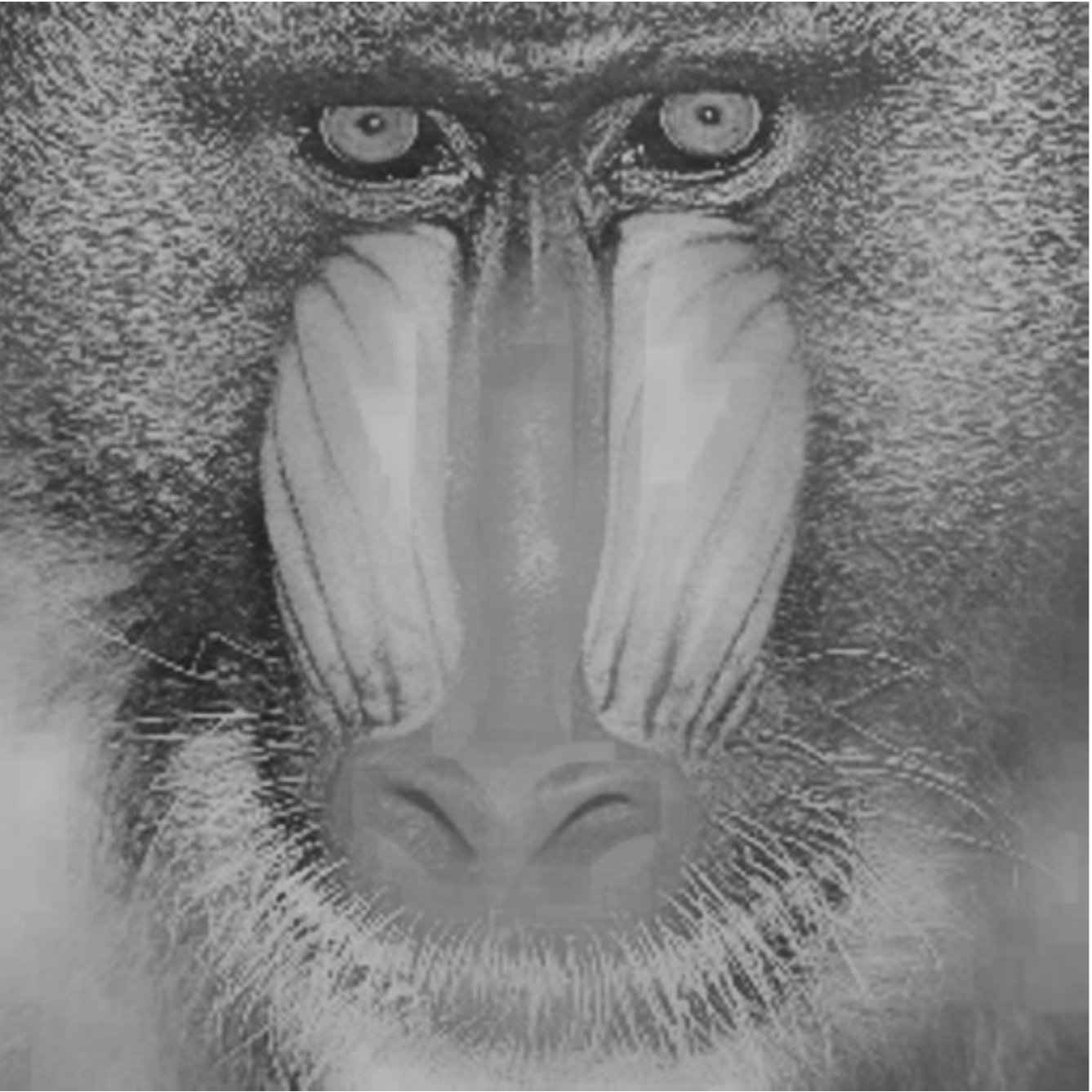}}
	\hspace{0.1mm}
	\subfloat[Error.]{\includegraphics[height=0.40\linewidth,keepaspectratio]{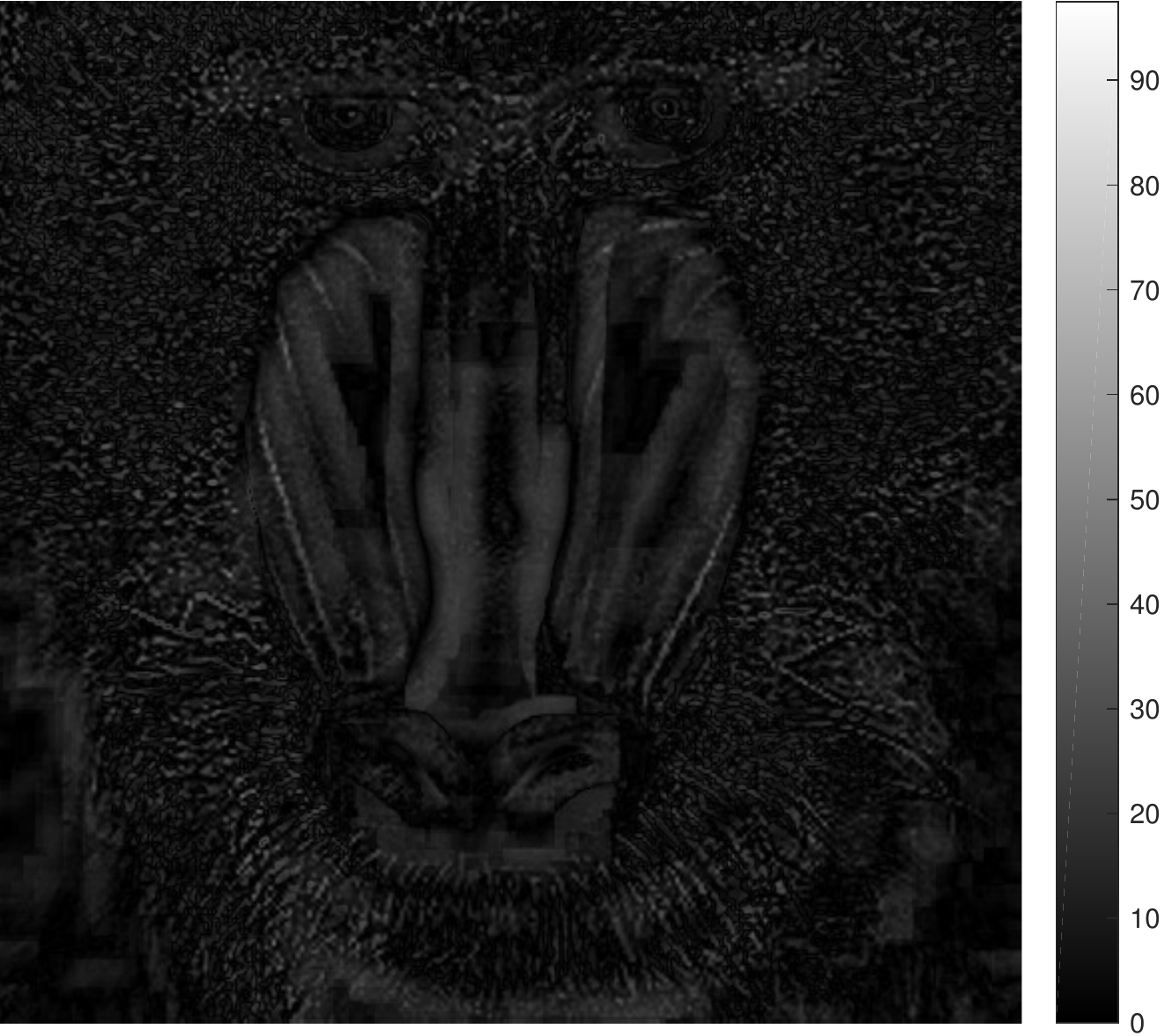}} \\
	\subfloat[\cite{Mozerov2015}, $21.55$ dB.]{\includegraphics[height=0.40\linewidth,keepaspectratio]{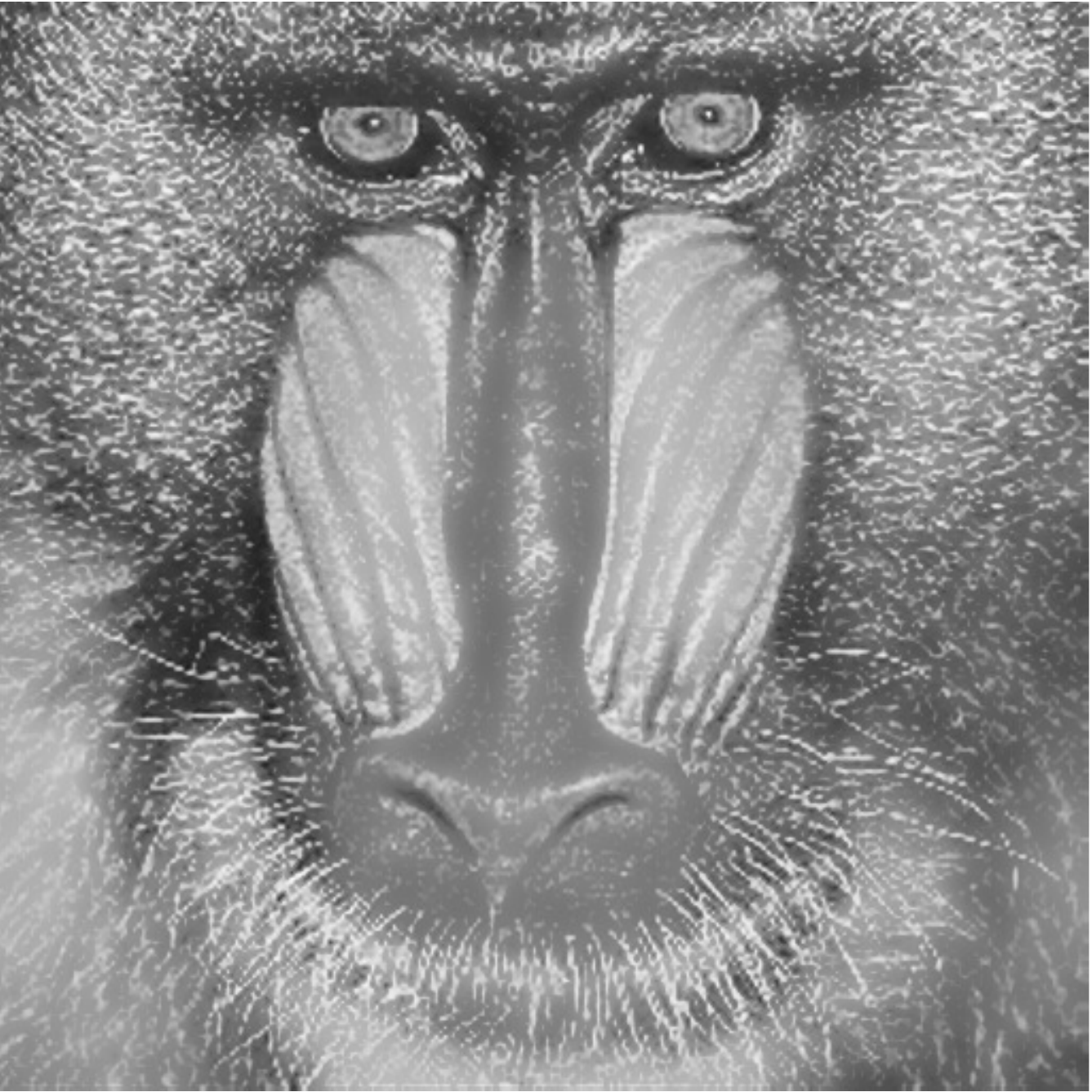}}
	\hspace{0.1mm}
	\subfloat[Error.]{\includegraphics[height=0.40\linewidth,keepaspectratio]{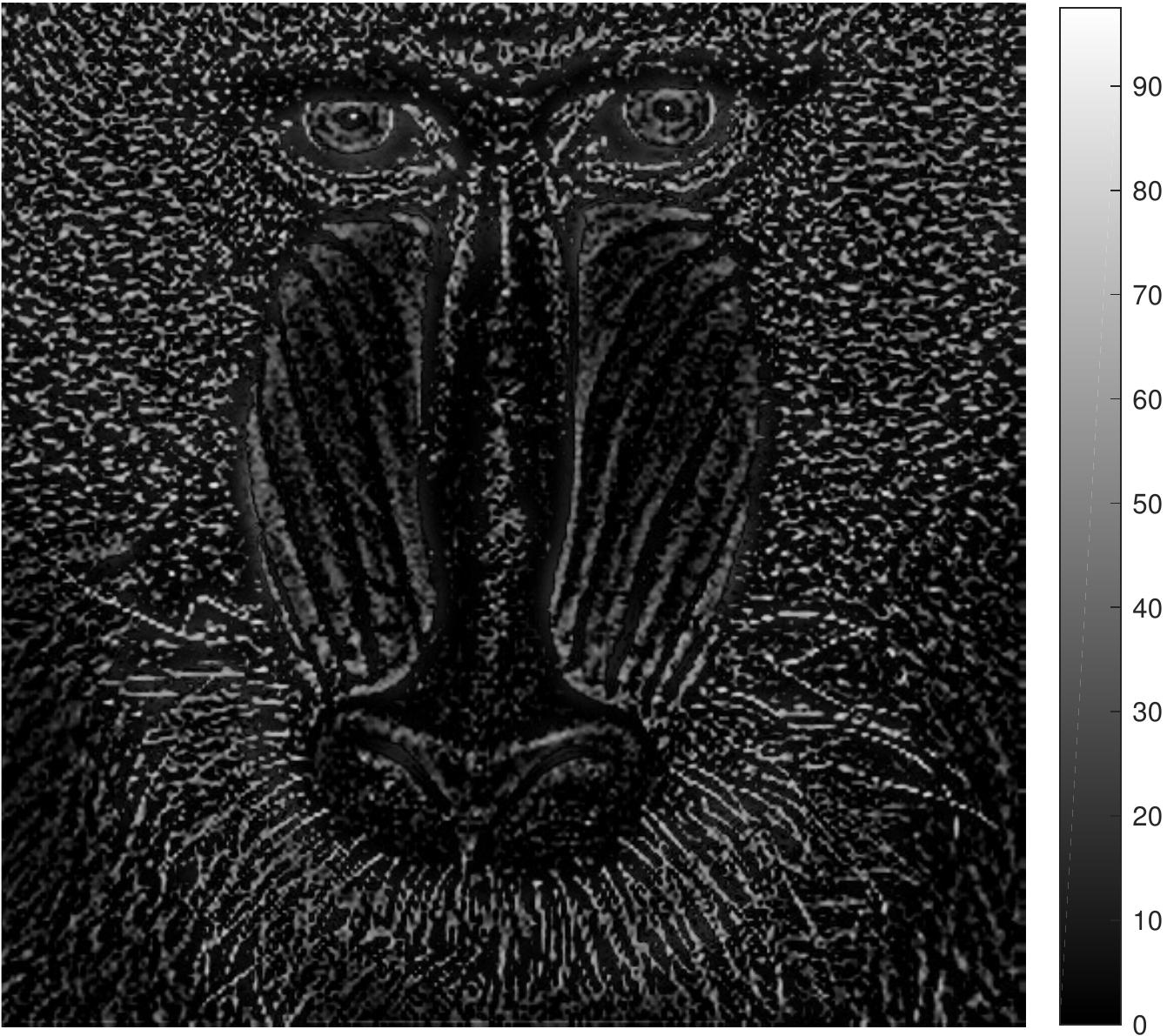}}
	\caption{Comparison of the proposed approximation with that from \cite{Mozerov2015} for classical bilateral filtering. The parameters are $\rho = 5$ and $\sigma_0 = 40$. (a) Input image; (b) Brute-force bilateral filtering; (c) Approximation using $N=2$; (d) Error between (c) and (b); (d) Approximation using $N=0$; (e) Error between (d) and (b); (f) Approximation using \cite{Mozerov2015}; (g) Error between (f) and (b). The error image gives the absolute value of the pixelwise difference (the dynamic range is scaled to $[0,98]$ for visual clarity).}
	\label{fig:VisualComp}
\end{figure}

To generate all the results reported in this paper, we have used Matlab 9.1 on a Linux system running on a 3.4 GHz CPU with 32 GB memory.
The values of $\sigma(i)$ used for the experiments are on a scale of $[0,255]$, the dynamic range of an $8$-bit image.
Before applying Algorithm \ref{alg:Proposed}, we rescaled the dynamic range to $[0,1]$ and $\sigma(i)$ to $\sigma(i)/255$ (recall that the input image $f(i)$ is assumed to take values in $[0,1]$).
The output of the algorithm is finally rescaled back to $[0,255]$.

In Figure \ref{fig:VisualComp}, we compare the results of Algorithm \ref{alg:Proposed} and \cite{Mozerov2015} for the parameter settings $\rho=5$ and $\sigma_0 = 40$. The results of Algorithm \ref{alg:Proposed} are shown for two different approximation orders (cf. caption for details). For the method in \cite{Mozerov2015}, we set $\Delta=0$, which results in the narrowest valid uniform distribution (cf. \cite{Mozerov2015} for definition of $\Delta$). Note that we actually fit a polynomial of degree $0$ (a constant function) to the local histogram when $N=0$; this is done by matching the zeroth moments of the constant function and the histogram. On the other hand, the first moment is used for matching in \cite{Mozerov2015}. As a result, the two approaches result in different heights of the constant function. However, the height appears in the numerator and denominator of \eqref{app1} and is thus canceled out. Therefore the height of the constant function may be set to any non-zero value. Thus, the only difference between our method (when $N=0$) and \cite{Mozerov2015} is that the intervals on which the constant functions are fitted are different. These are $[\alpha_i,\beta_i]$ and $[f(i),2\bar{f}(i)-f(i)]$ for our method and \cite{Mozerov2015} respectively, where $\bar{f}(i)$ is the smoothed version of $f(i)$. Notice in Figures \ref{fig:VisualComp}(e) and (g) that using the exact values of $\alpha_i$ and $\beta_i$ alone results in few dBs improvement. The result using $N=2$ is shown to confirm our intuition that a higher-degree polynomial should result in better approximation. We have reported some additional PSNR values for different $\rho$ and $N$ in Table \ref{tab:PSNR}. As expected, the PSNR progressively increases with $N$.

\begin{table}[t!]
\centering
\caption{$\mathrm{PSNR}$ values (in dB) for bilateral filtering ($\sigma_0 = 40$) obtained using Algorithm \ref{alg:Proposed}.}

\begin{adjustbox}{max width=\linewidth}
\begin{tabular}{|c|c|c|c|c|c|c|c|}
\hline
$N$				& $0$		& $1$		& $2$		& $3$		& $4$		& $5$		& $6$		\\ \hline
$\rho=3$		& $29.03$	& $35.16$	& $43.26$	& $50.28$	& $58.51$	& $67.58$	& $76.61$	\\ \hline
$\rho=5$		& $28.15$	& $32.87$	& $40.83$	& $47.56$	& $55.69$	& $64.99$	& $73.24$	\\ \hline
$\rho=10$		& $27.49$	& $30.86$	& $38.21$	& $44.31$	& $52.19$	& $61.12$	& $68.04$	\\ \hline
\end{tabular}
\end{adjustbox}

\label{tab:PSNR}
\end{table}

We now compare the timing of  Algorithm \ref{alg:Proposed} with the brute-force implementation of \eqref{eq:BF_adaptive1}. The timings in Table \ref{tab:Time} are for the $512 \times 512$ image in Figure \ref{fig:VisualComp}. Notice that for a fixed $N$, there is very little variation in timing for different $\rho$. Our algorithm is thus $O(1)$ as claimed. In contrast, the timing for the brute-force implementation increases with $\rho$ as expected. Notice in Table \ref{tab:Time} that, even when $N$ is as high as $5$, the speedup is by at least $20\times$ and often as much as $60\times$.

\begin{table}[t!]
\centering
\caption{Comparison of the timings of Algorithm \ref{alg:Proposed} and the brute-force implementation for different $\rho$ and $N$.}

\begin{adjustbox}{max width=\linewidth}
\begin{tabular}{|c|c|c|c|c|c|}
\hline
$\rho$			& $3$		& $5$		& $7$		& $9$		& $11$		\\ \hline
$N=1$			& $75$ ms	& $73$ ms	& $74$ ms	& $69$ ms	& $70$ ms	\\ \hline
$N=3$			& $116$ ms	& $119$ ms	& $125$ ms	& $120$ ms	& $125$ ms	\\ \hline
$N=5$			& $171$	ms	& $179$ ms	& $184$ ms	& $189$ ms	& $186$ ms	\\ \hline
Brute-force		& $3.90$ s	& $6.87$ s	& $8.60$ s	& $10.9$ s	& $12.4$ s	\\ \hline
\end{tabular}
\end{adjustbox}

\label{tab:Time}
\end{table}

\section{Applications}
\label{sec:app}

To demonstrate the potential of the proposed fast algorithm, we consider three applications of the adaptive bilateral filter: (1) image sharpening and noise removal, (2) JPEG deblocking, and (3) texture filtering.
The first application is directly based on the method in \cite{Zhang2008}, whereas the second is a minor modification of the method in \cite{Zhang2009}.
Thus, our essential contribution is the speedup obtained using our algorithm.
The third application is novel in the sense that we propose a new method for texture removal, and we show that our results are competitive with state-of-the-art methods.

\subsection{Image sharpening and noise removal}

\begin{figure}[t!]
	\raggedright
	\subfloat[Input ($1024 \times 1024$).]{\includegraphics[height=0.45\linewidth,keepaspectratio]{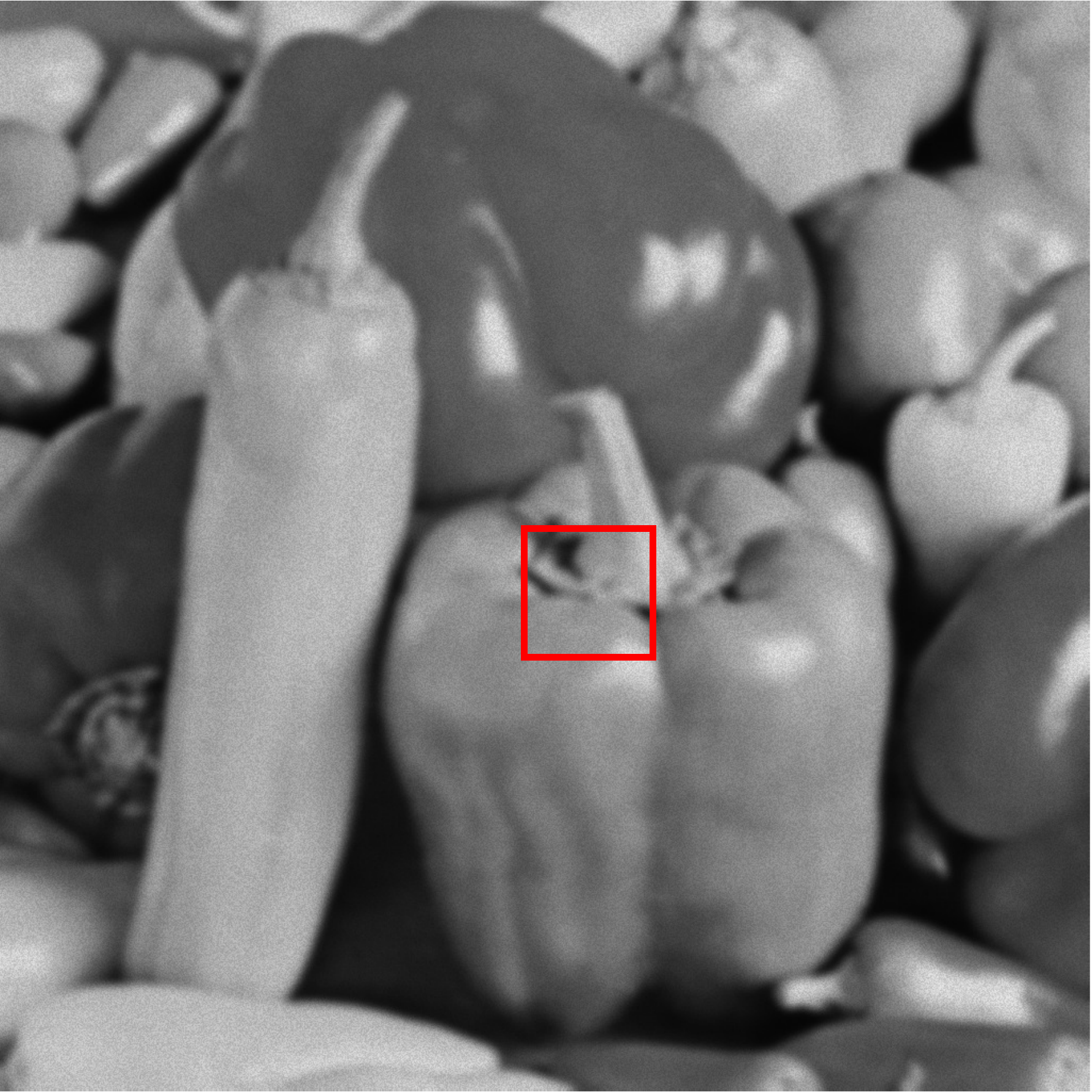}}
	\hspace{0.1mm}
	\subfloat[Classical ($\sigma_0=30$), $30$ s.]{\includegraphics[height=0.45\linewidth,keepaspectratio]{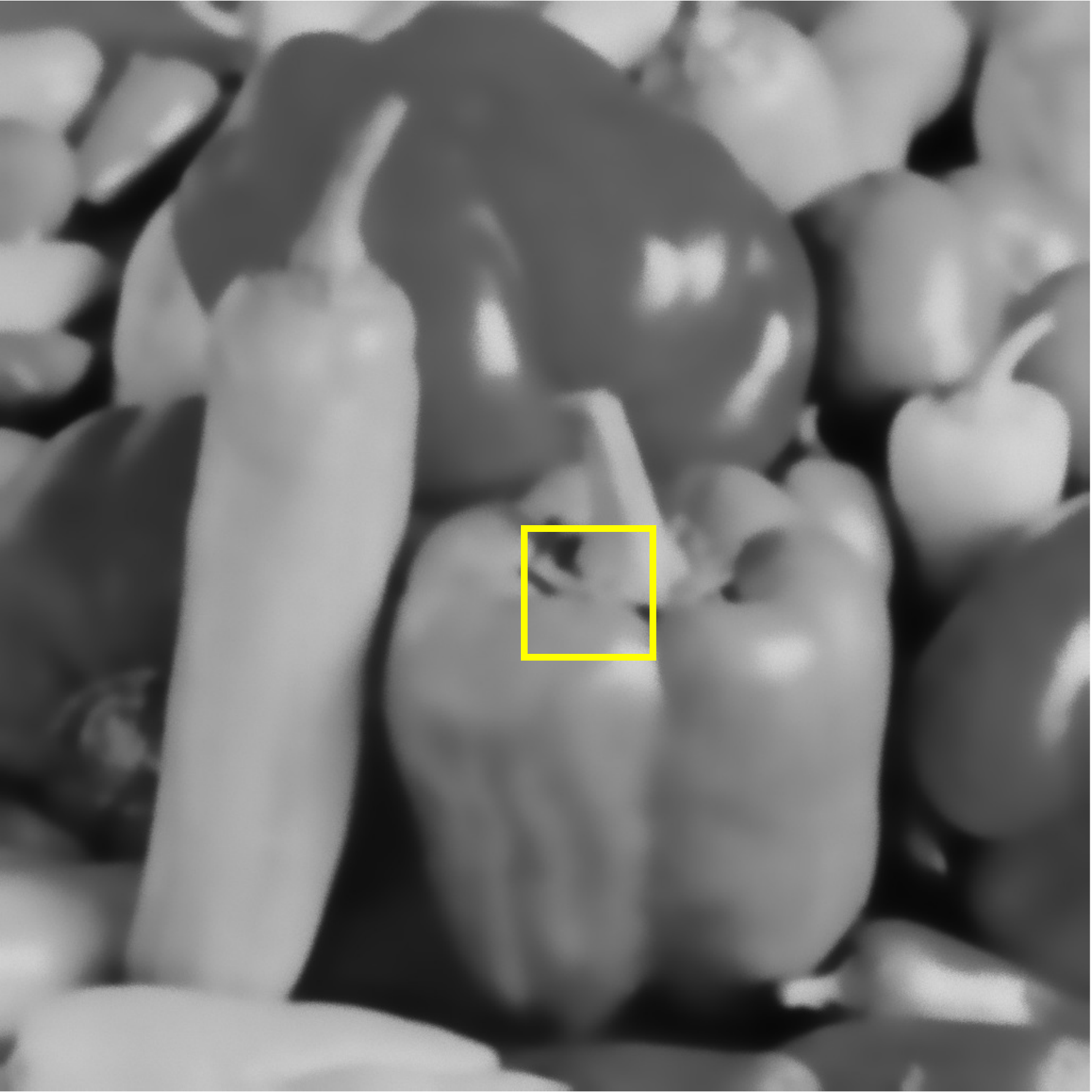}} \\
	\subfloat[Adaptive (proposed), $0.74$ s.]{\includegraphics[height=0.45\linewidth,keepaspectratio]{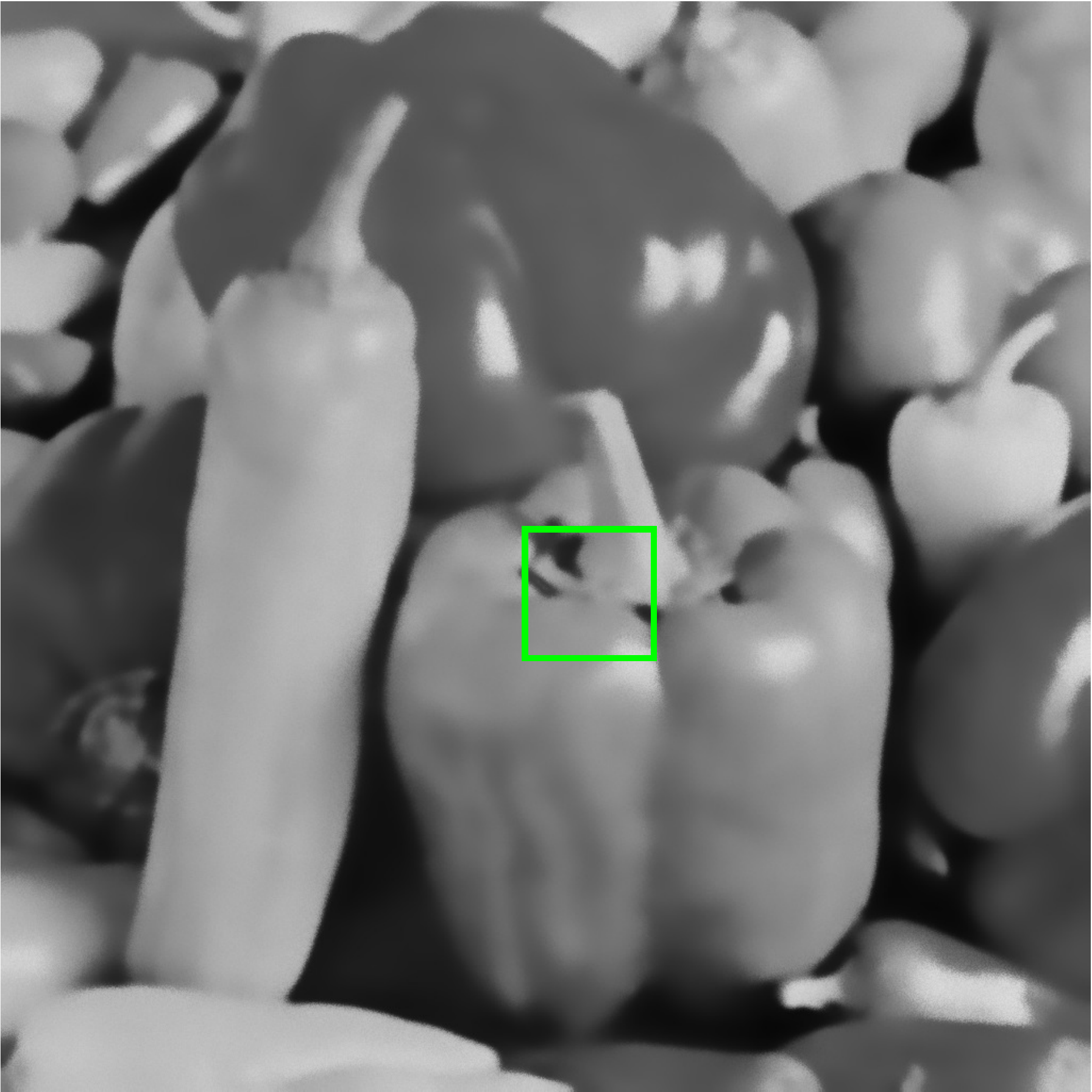}}
	\hspace{0.1mm}
	\subfloat[Adaptive (brute-force), $31$ s.]{\includegraphics[height=0.45\linewidth,keepaspectratio]{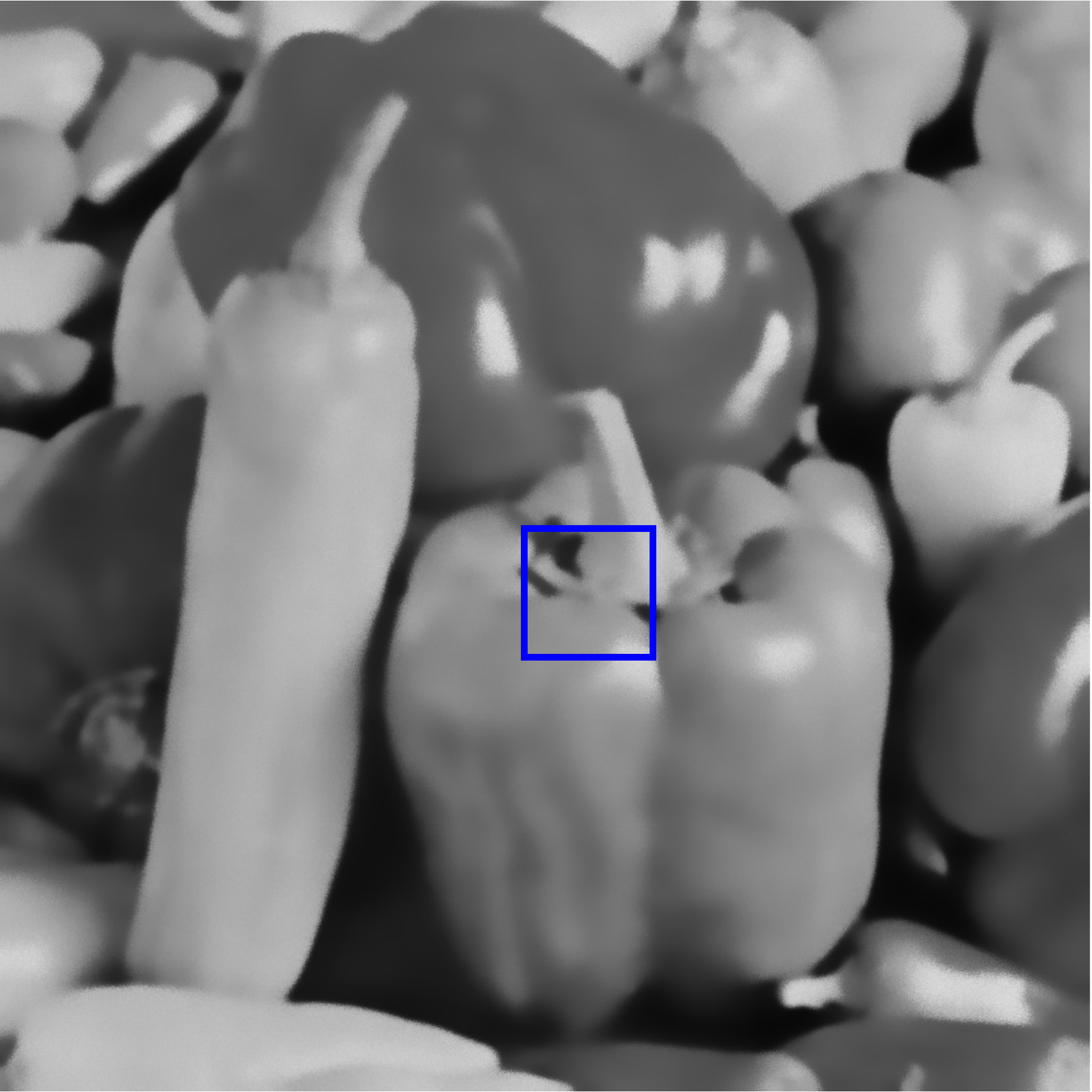}} \\
	\centering
	\subfloat[]{\includegraphics[height=0.44\linewidth,keepaspectratio]{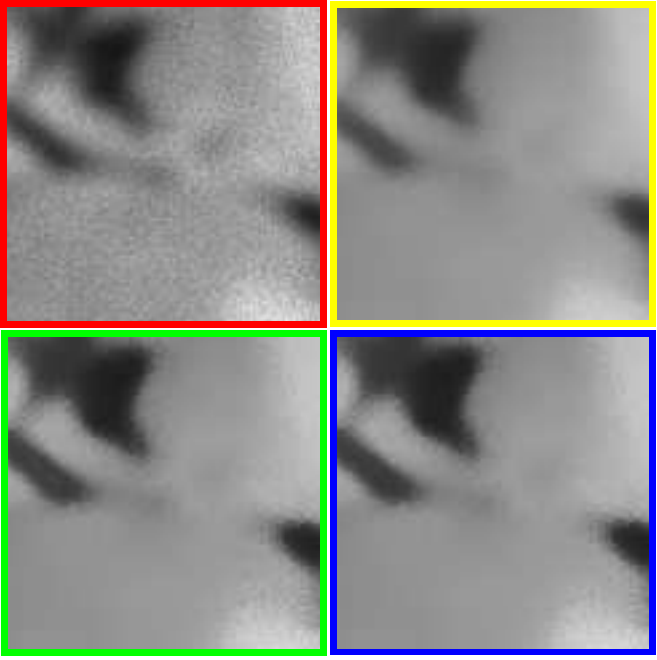}}
	\hspace{0.1mm}
	\subfloat[$\sigma(i)$.]{\includegraphics[height=0.45\linewidth,keepaspectratio]{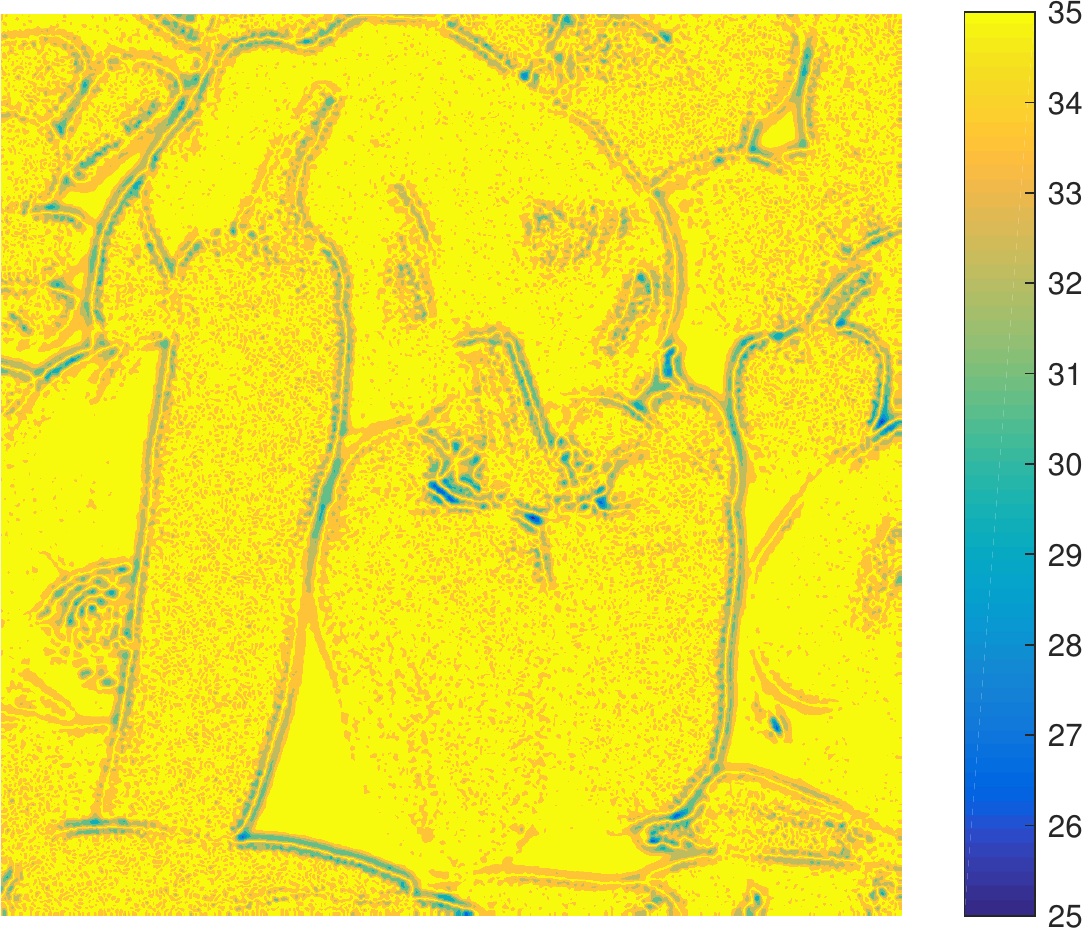}}
	\caption{Image sharpening and noise removal results for classical and adaptive bilateral filtering. The map of $\sigma(i)$ used for adaptive bilateral filtering is shown in (f). We used $\rho=5$ for all results, and $N=5$ in (c). The PSNR between (c) and (d) is $59.8$ dB.}
	\label{fig:Sharpening1}
\end{figure}

\begin{figure}[t!]
	\centering
	\subfloat[Input.]{\includegraphics[width=0.30\linewidth,keepaspectratio]{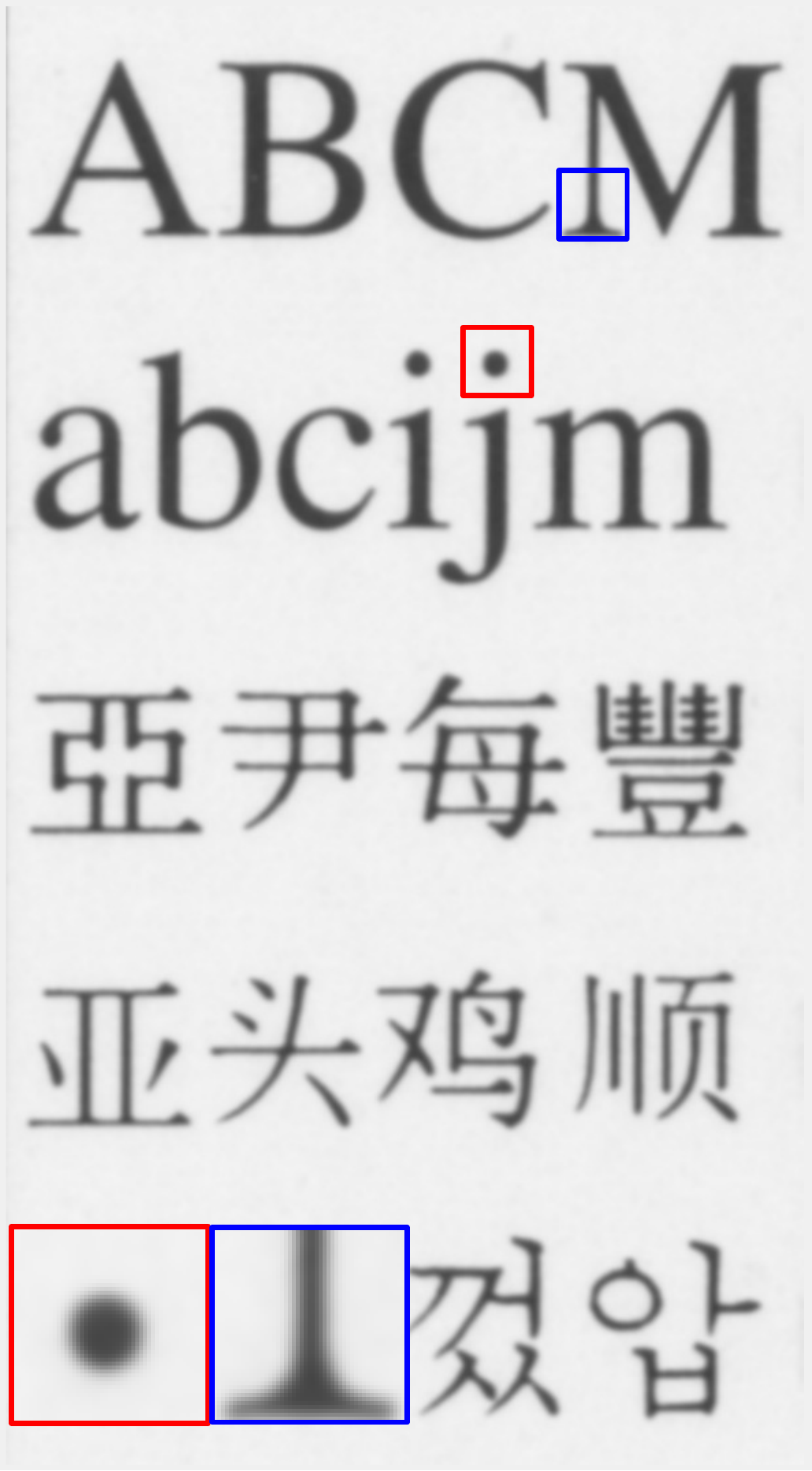}}
	\hspace{0.1mm}
	\subfloat[Proposed, $0.41$ s.]{\includegraphics[width=0.30\linewidth,keepaspectratio]{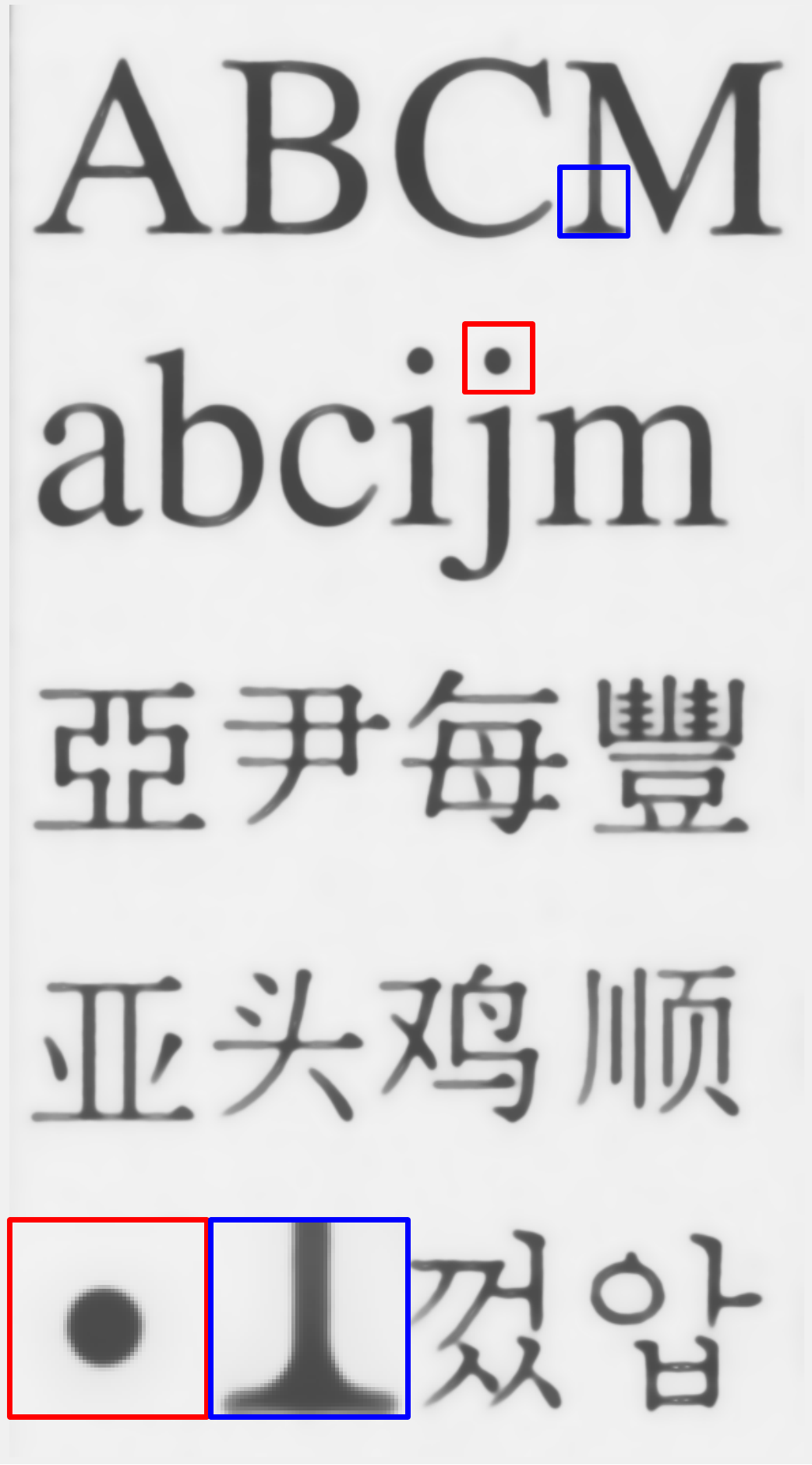}}
	\hspace{0.1mm}
	\subfloat[Brute-force, $11.6$ s.]{\includegraphics[width=0.30\linewidth,keepaspectratio]{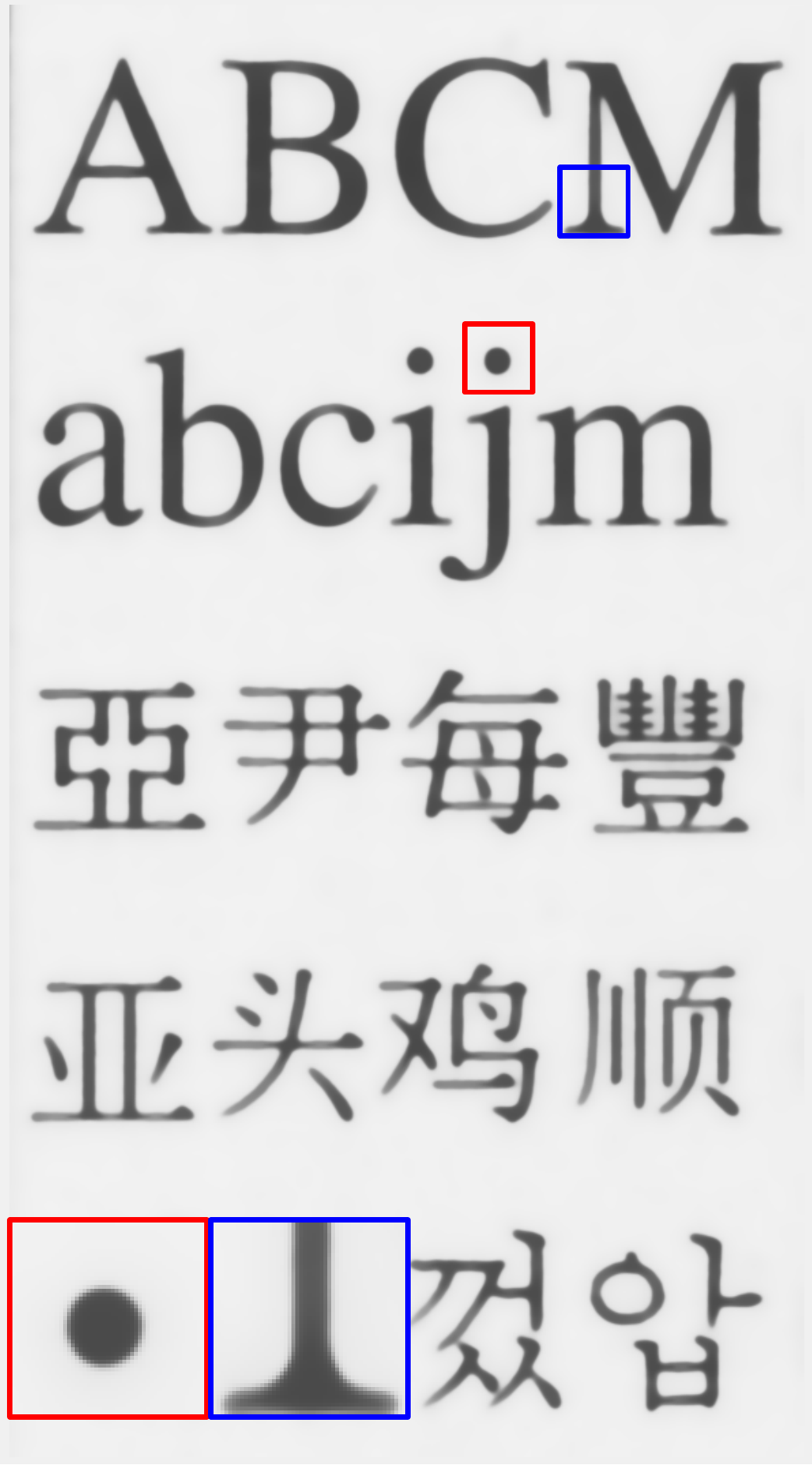}}
	\caption{Image sharpening and noise removal results ($\rho=5$, $N=5$) for the image ($1066 \times 584$) used in \cite{Zhang2008}. The PSNR between (b) and (c) is $54.66$ dB.}
	\label{fig:Sharpening3}
\end{figure}

The adaptive bilateral filter was originally used in \cite{Zhang2008} for sharpening along with denoising of fine grains. We apply our fast algorithm for this task. 

Following \cite{Zhang2008}, we set $\theta(i) = f(i) + \zeta(i)$ where $f(i)$ is the input image. The offset $\zeta(i)$ is set to be
\begin{equation*}
\zeta(i) = f(i) - \bar{f}(i),
\end{equation*}
where $\bar{f}(i)$ is the average intensity in the $\Omega$-neighborhood of $i$.
To determine $\sigma(i)$, we high-pass filter the input image using the Laplacian-of-Gaussian filter, and 
apply an affine mapping on the absolute value of the resulting image to get $\sigma(i)$.
The affine function maps low values of its input to high values of $\sigma(i)$ and vice-versa.

A typical result of sharpening and noise removal in shown in Figure \ref{fig:Sharpening1}.
The input image and the values of $\sigma(i)$ are shown in Figures \ref{fig:Sharpening1}(a) and \ref{fig:Sharpening1}(f). 
The image in Figure \ref{fig:Sharpening1}(c) is obtained by applying Algorithm \ref{alg:Proposed} on the input image, using the prescribed values of $\theta(i)$ and $\sigma(i)$.
Notice that the output image contains visibly sharper boundaries and less noise grains than the input.
Importantly, it is visually indistinguishable from the output of brute-force implementation shown in Figure \ref{fig:Sharpening1}(d).
Indeed, the PSNR between the images in (c) and (d) is quite high ($59.8$ dB).
However, note that our method is about $40$ times faster.
The image in Figure \ref{fig:Sharpening1}(b) is obtained by applying the classical bilateral filter on the input image.
Note that bilateral filtering removes noise as expected, but it cannot sharpen the edges.
In \cite{Zhang2008}, the adaptive bilateral filter was used for sharpening a text image.
We show the result for the same image in Figure \ref{fig:Sharpening3}, where the speedup using Algorithm \ref{alg:Proposed} is about $28 \times$.

\subsection{JPEG deblocking }

\begin{figure}[t!]
    \centering
	\subfloat[Input image.]{\includegraphics[height=0.35\linewidth,keepaspectratio]{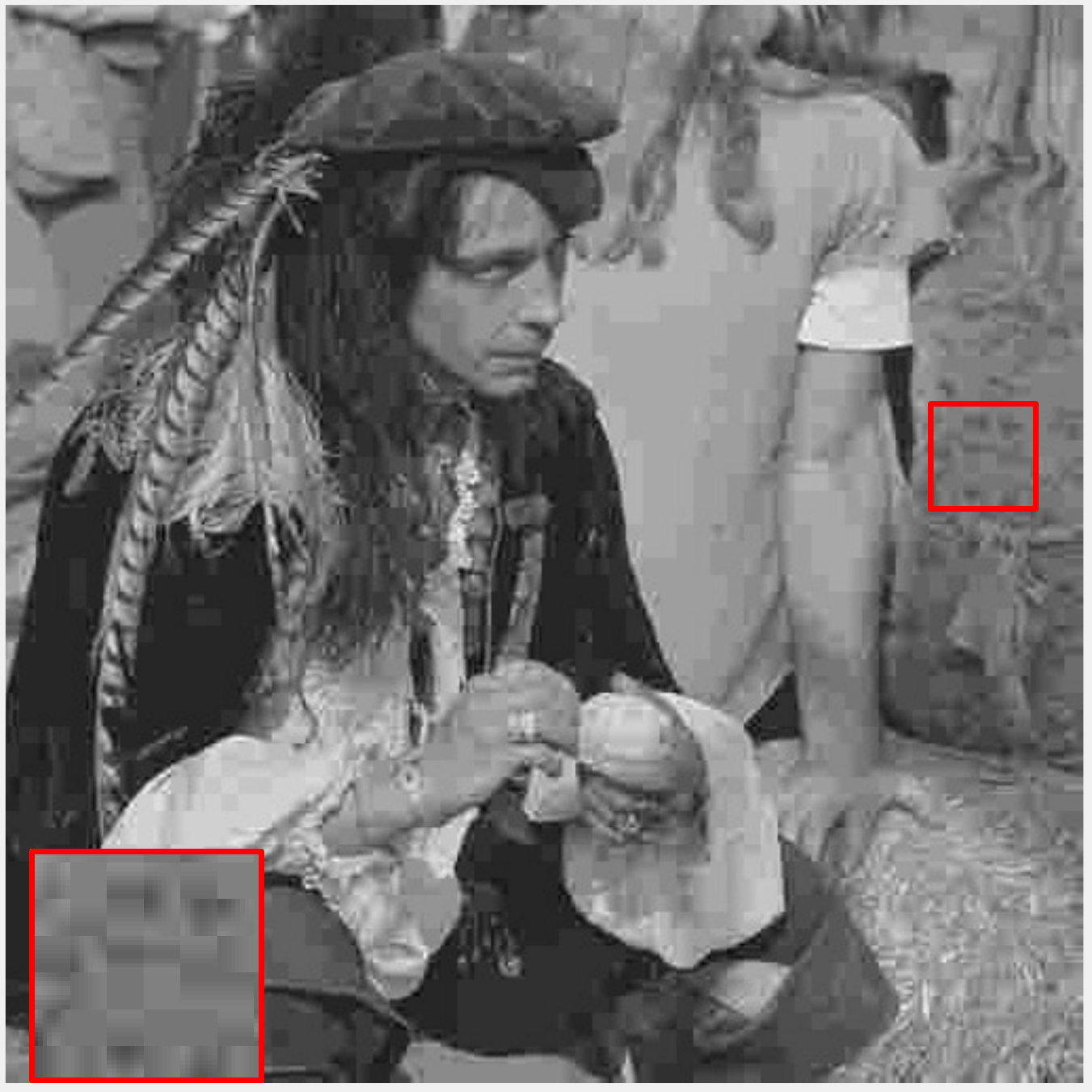}}
    \hspace{0.1mm}
    \subfloat[$\sigma$ map.]{\includegraphics[height=0.35\linewidth,keepaspectratio]{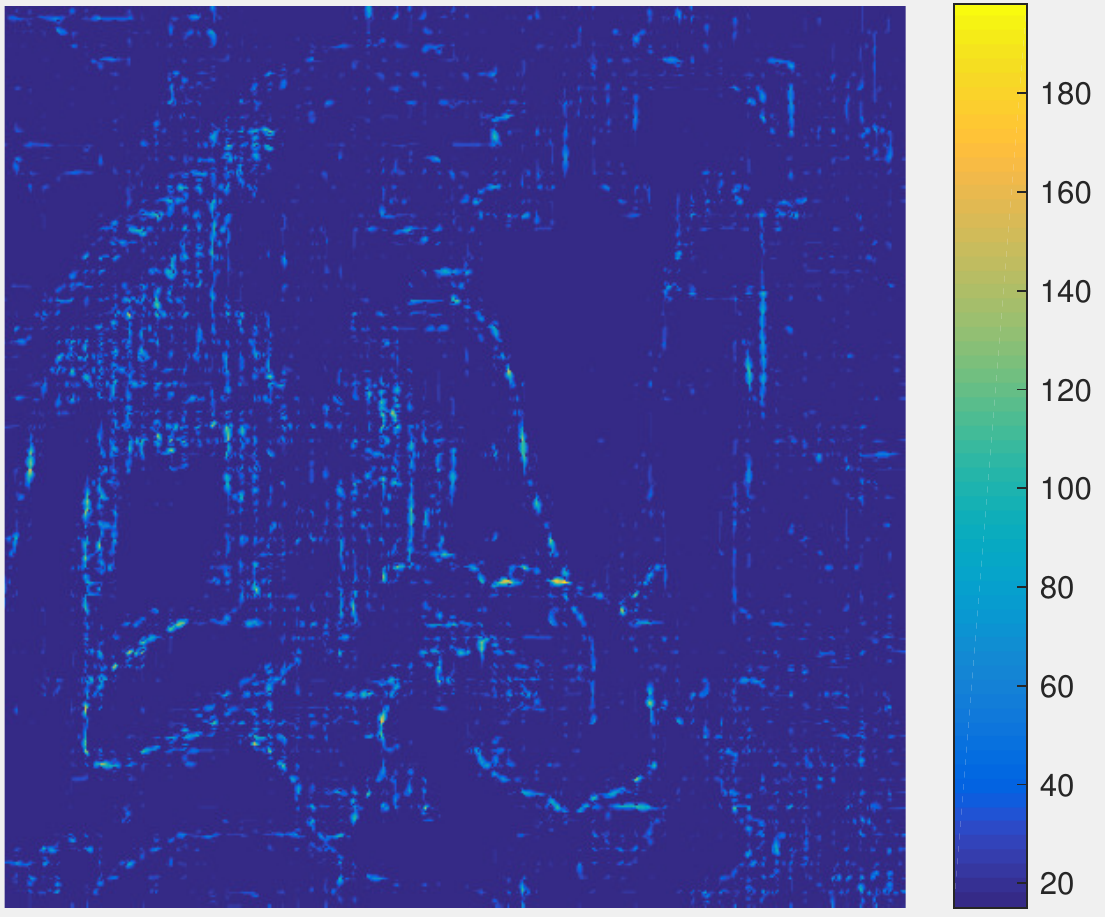}}
    \caption{Computation of $\sigma(i)$ for deblocking.}
    \label{fig:Artifact1}
\end{figure}

It was shown in \cite{Zhang2009} that adaptive bilateral filtering can be used to remove blocking artifacts from JPEG images,
where the width of the spatial kernel was also adapted.
However, adapting the spatial kernel width is beyond the scope of the proposed algorithm.
We will rather show that changing the range kernel width alone can satisfactorily remove blocking.
The motivation in \cite{Zhang2009} for using a locally adaptive range width is as follows.
In JPEG, the image is tiled using $8 \times 8$ blocks, and the blocks are compressed using DCT. 
Blocking artifacts occur due to discontinuities at the boundary of adjacent  blocks.
To effectively smooth out the discontinuity at a boundary pixel $i$, the authors in \cite{Zhang2009} proposed to set $\sigma(i)$ in proportion with the intensity difference across the boundary.

For completeness, we briefly describe how $\sigma(i)$ is computed in \cite{Zhang2009}.
The image is first divided into non-overlapping $8 \times 8$  blocks corresponding to the blocks used in JPEG compression.
Then a discontinuity map $B(i)$ is calculated at each pixel as follows.
At a vertical or horizontal edge of a given block, $B(i)$ is set to be the intensity difference across the respective edge (corner pixels are handled appropriately).
The corresponding values for the four center pixels in the block are set to zero.
For the remaining pixels in the block, $B(i)$ is set using linear interpolation.
Finally, $\sigma(i)$ is set as
\begin{equation*}
\sigma(i) = \max (\sigma_0,B(i)).
\end{equation*}
We refer the reader to \cite{Zhang2009} for further details.
An example is shown in Figure \ref{fig:Artifact1}.
The center $\theta(i)$ is set as $f(i)$, where $f(i)$ is the input JPEG image.

\begin{figure}[t!]
    \centering
    \subfloat[{\scriptsize Uncompressed.}]{\includegraphics[height=0.245\linewidth,keepaspectratio]{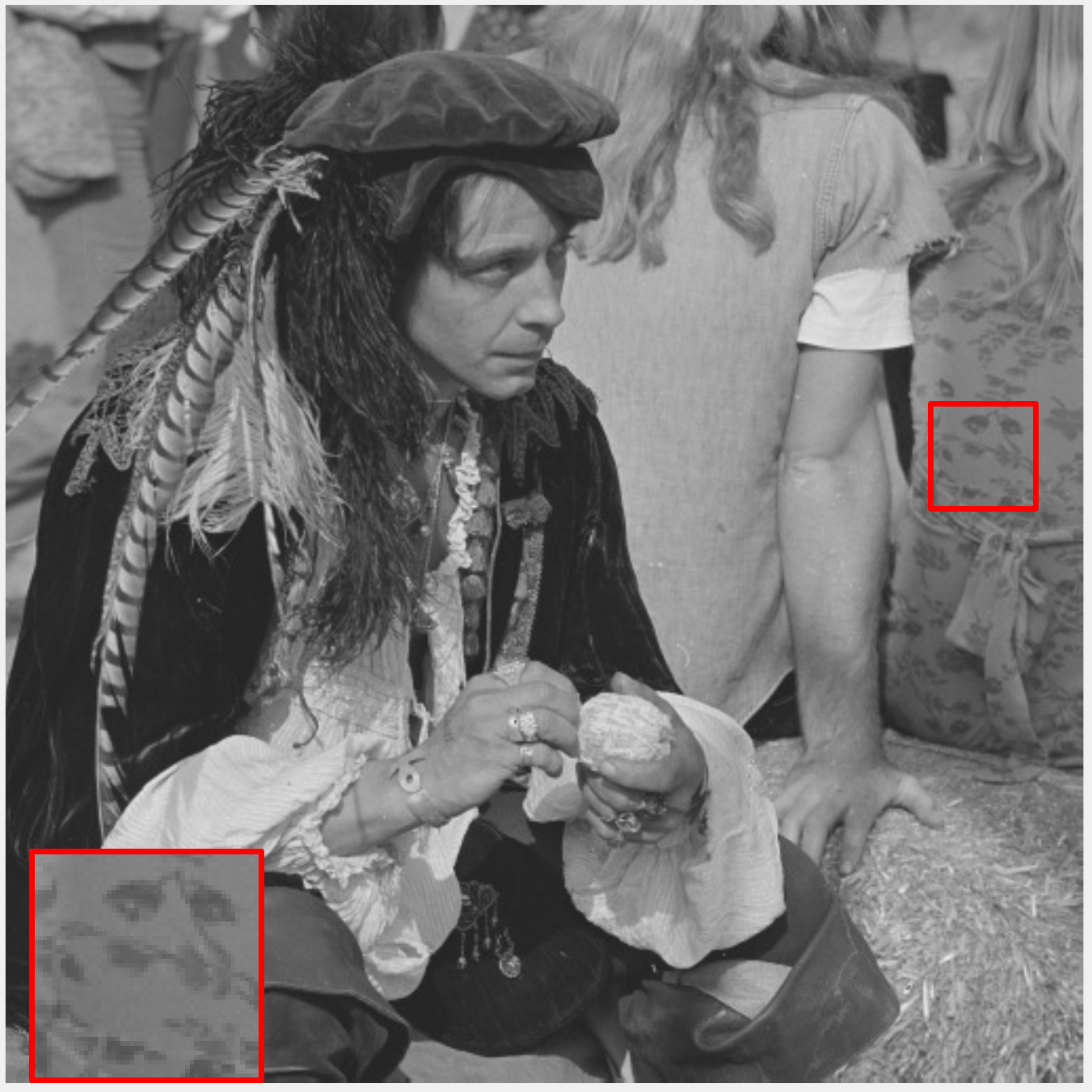}}
    \hfill
    \subfloat[{\scriptsize Input.}]{\includegraphics[height=0.245\linewidth,keepaspectratio]{deblocking_pirate_input.pdf}}
    \hfill
    \subfloat[{\scriptsize Algorithm \ref{alg:Proposed}.}]{\includegraphics[height=0.245\linewidth,keepaspectratio]{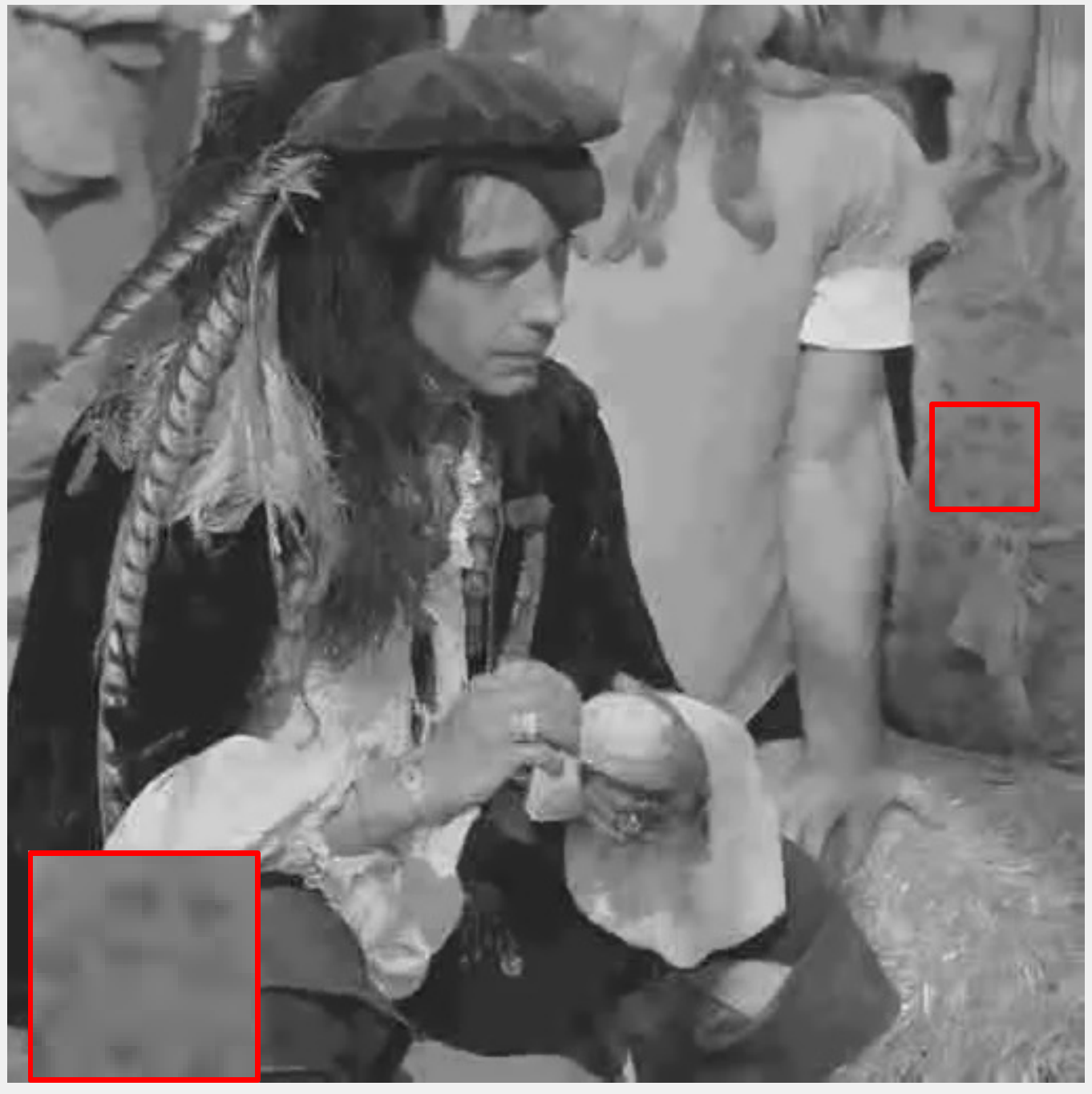}}
    \hfill
    \subfloat[{\scriptsize Brute force.}]{\includegraphics[height=0.245\linewidth,keepaspectratio]{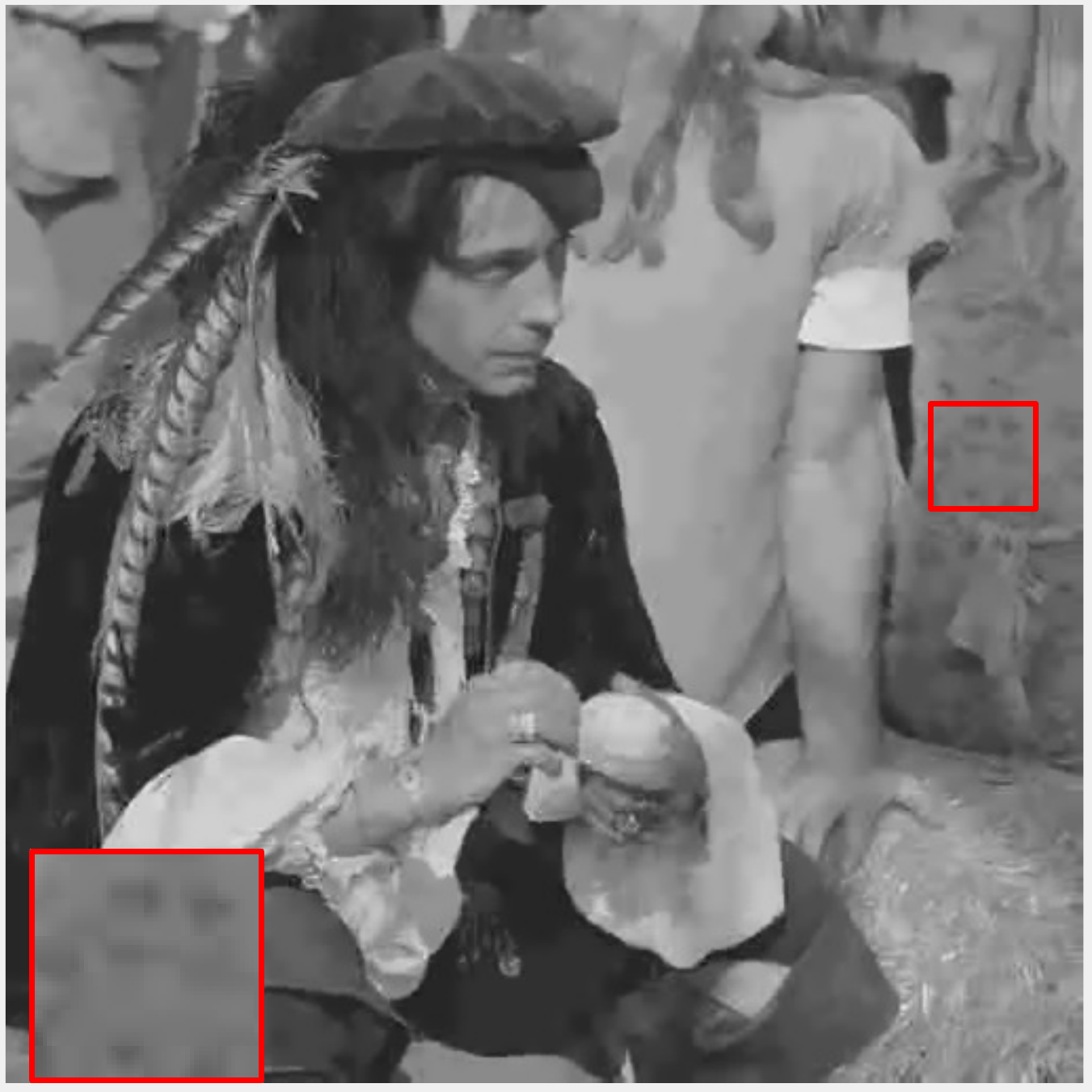}}\\
    \subfloat[{\scriptsize \cite{Li2014}.}]{\includegraphics[height=0.245\linewidth,keepaspectratio]{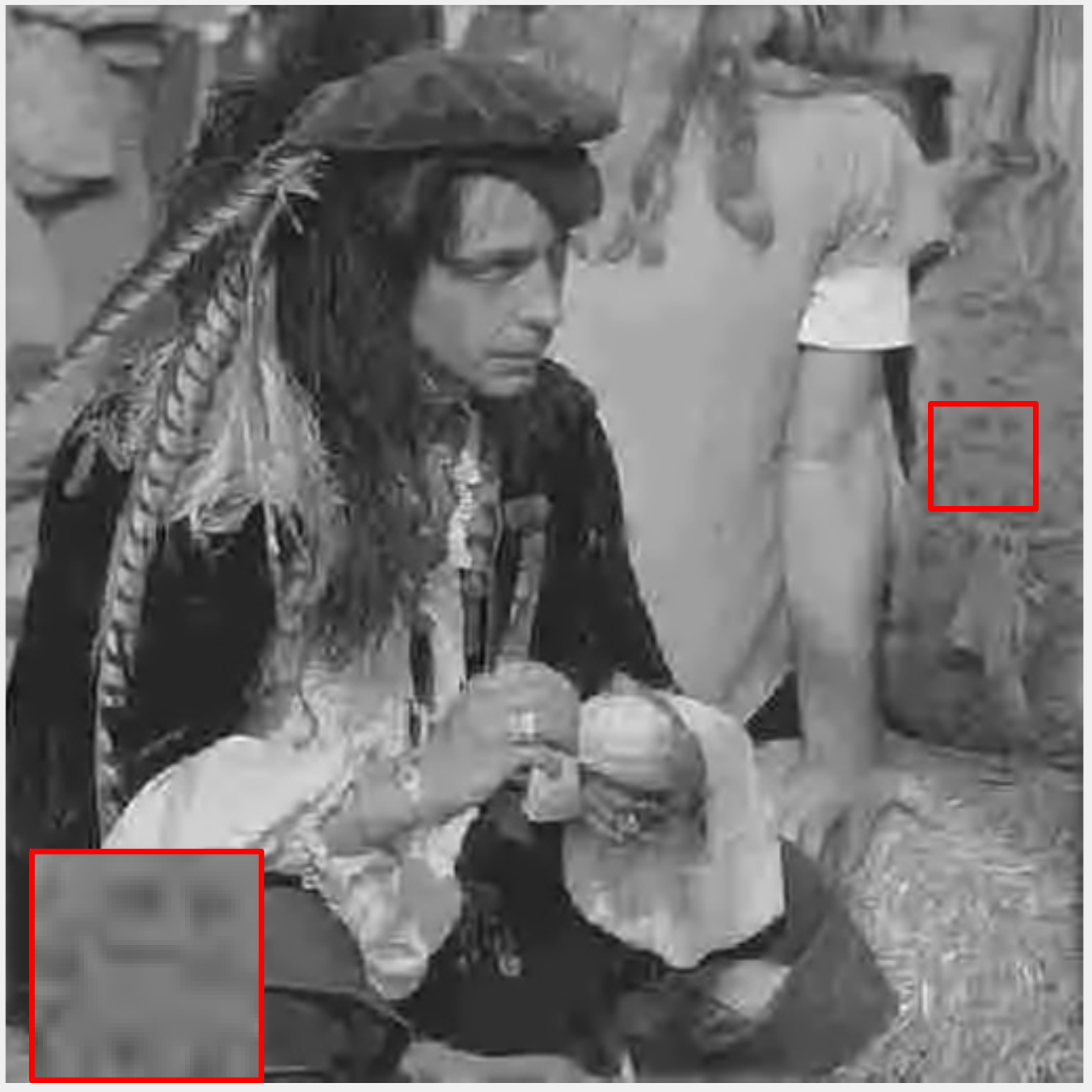}}
    \hfill
    \subfloat[{\scriptsize \cite{Zhang2016_concolor}.}]{\includegraphics[height=0.245\linewidth,keepaspectratio]{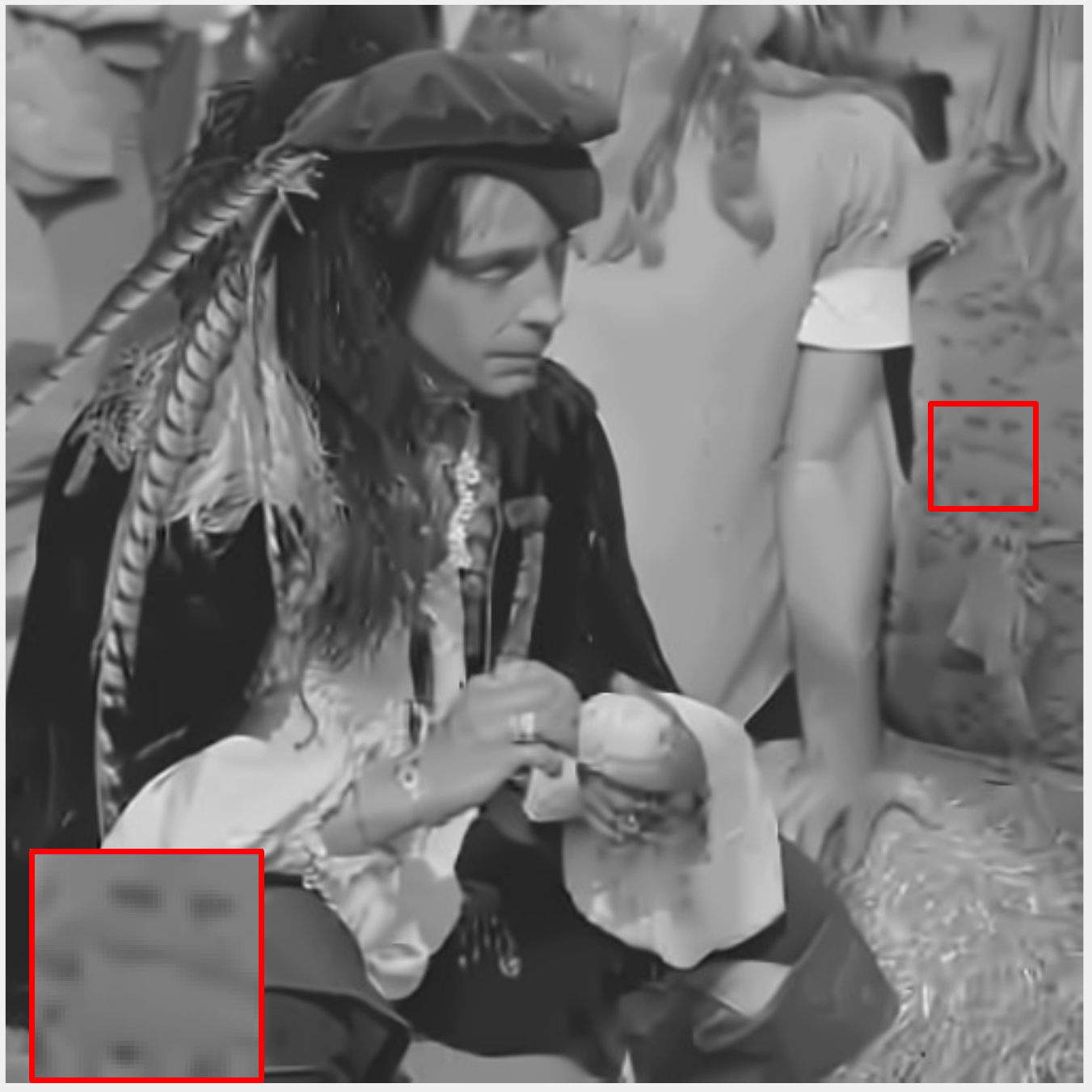}}
    \hfill
    \subfloat[{\scriptsize \cite{Zhao2017}.}]{\includegraphics[height=0.245\linewidth,keepaspectratio]{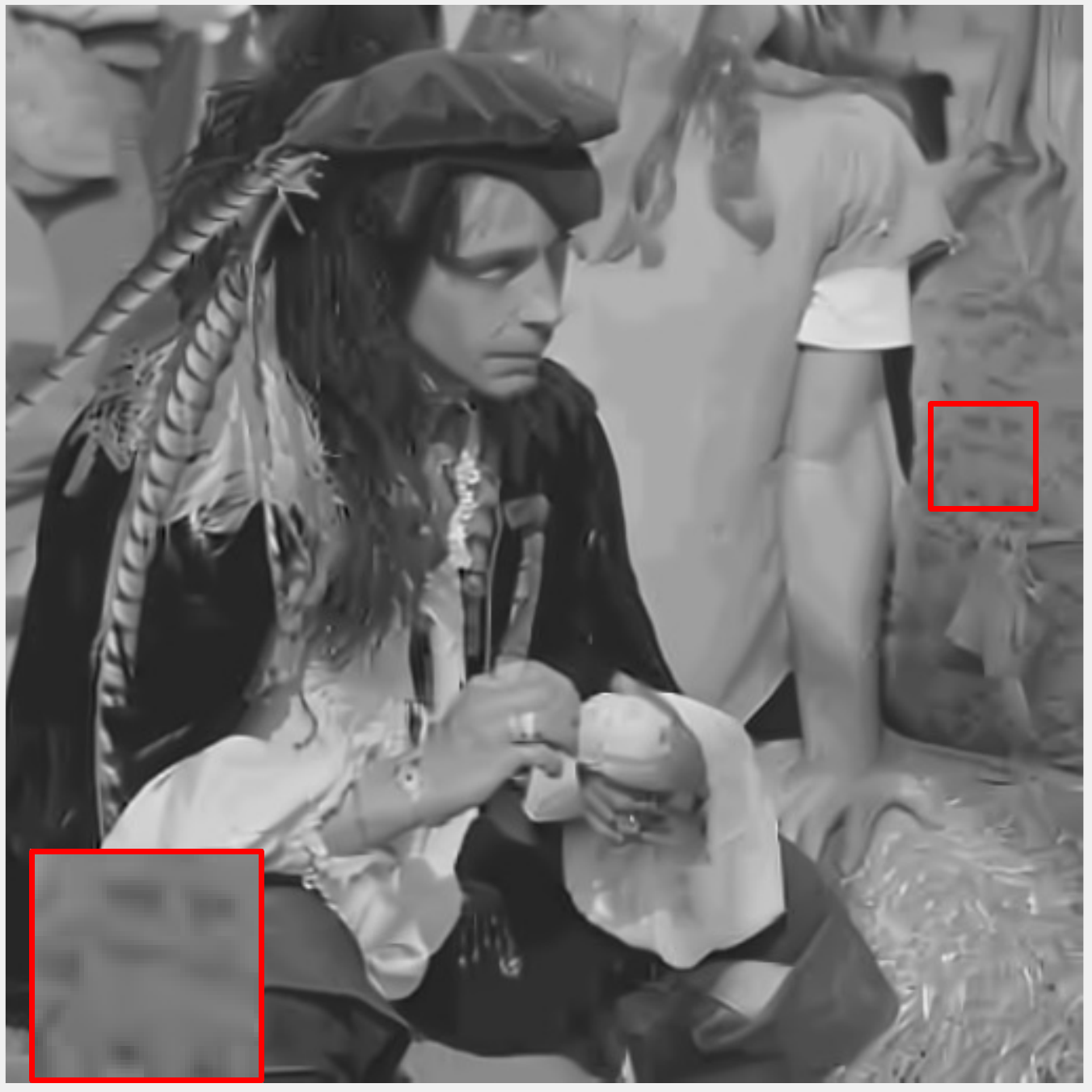}}
    \hfill
    \subfloat[{\scriptsize \cite{Golestaneh2014}.}]{\includegraphics[height=0.245\linewidth,keepaspectratio]{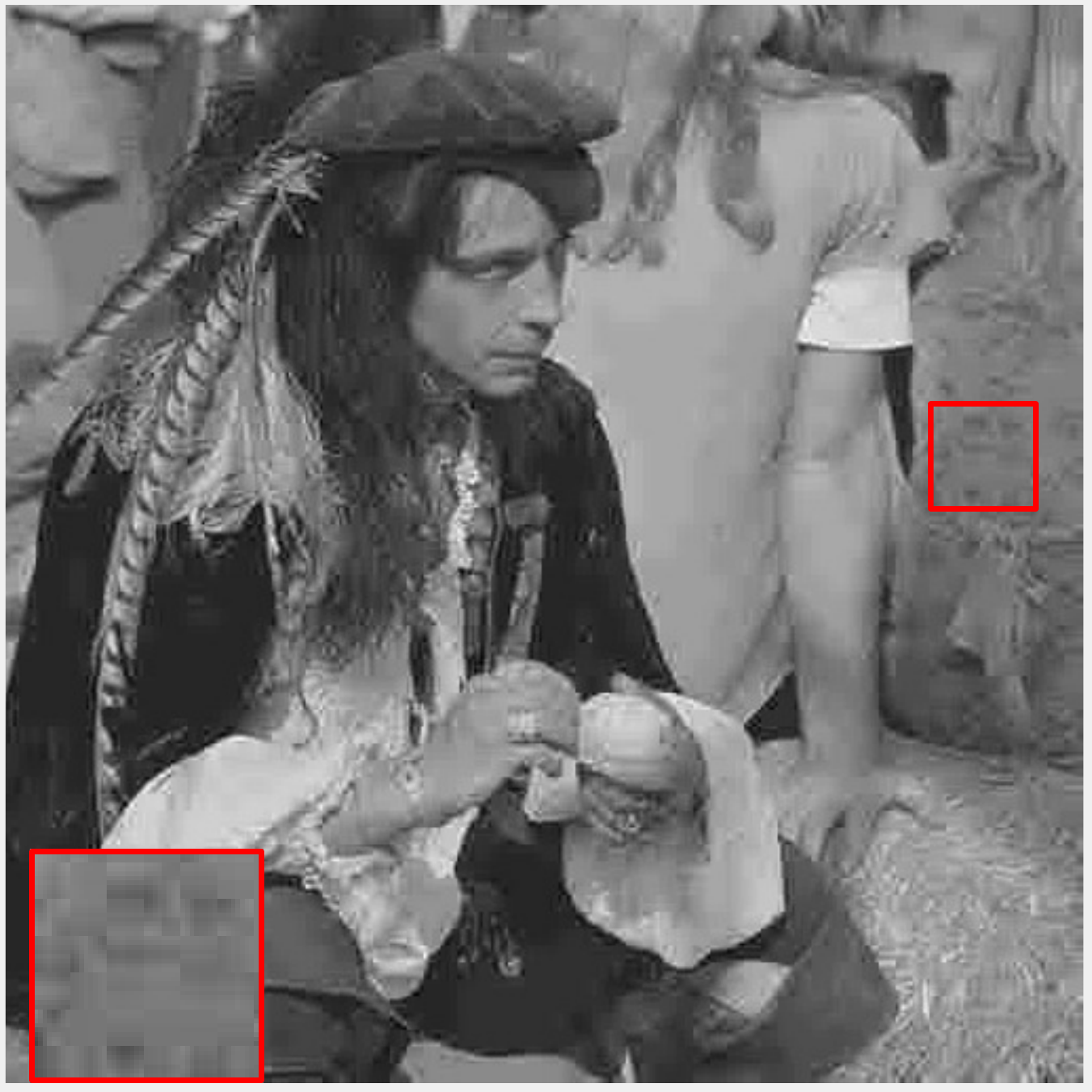}}
    \caption{JPEG deblocking results (image size $512 \times 512$). The (Runtime, PSNR, SSIM) values relative to the uncompressed image (a) are as follows (PSNR in dB):
    (b) $-$, $28.24$, $0.7612$;
    (c) $0.16$s, $28.52$, $0.7694$;
    (d) $3.3$s, $28.56$, $0.7709$;
    (e) $0.027$s, $27.60$, $0.7667$;
    (f) $20$ min, $29.32$, $0.7969$;
    (g) $89$s, $29.25$, $0.7934$;
    (h) $2.2$s, $28.04$, $0.7511$.
    The PSNR between (c) and (d) is $48.9$ dB.
    Please zoom in for a clearer view.}
    \label{fig:Artifact2}
\end{figure}

\begin{figure}[t!]
    \centering
    \subfloat[{\scriptsize Uncompressed.}]{\includegraphics[height=0.245\linewidth,keepaspectratio]{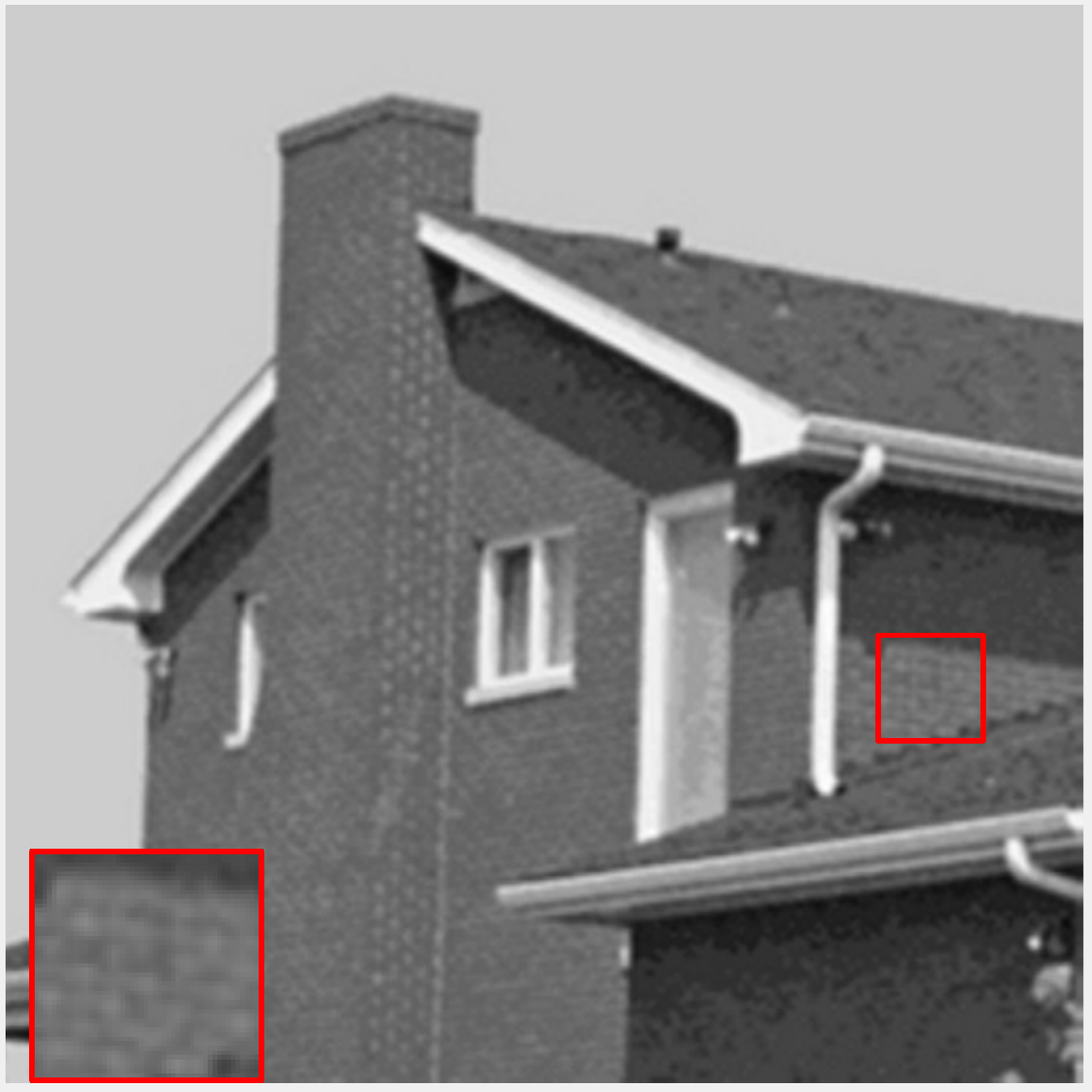}}
    \hfill
    \subfloat[{\scriptsize Input.}]{\includegraphics[height=0.245\linewidth,keepaspectratio]{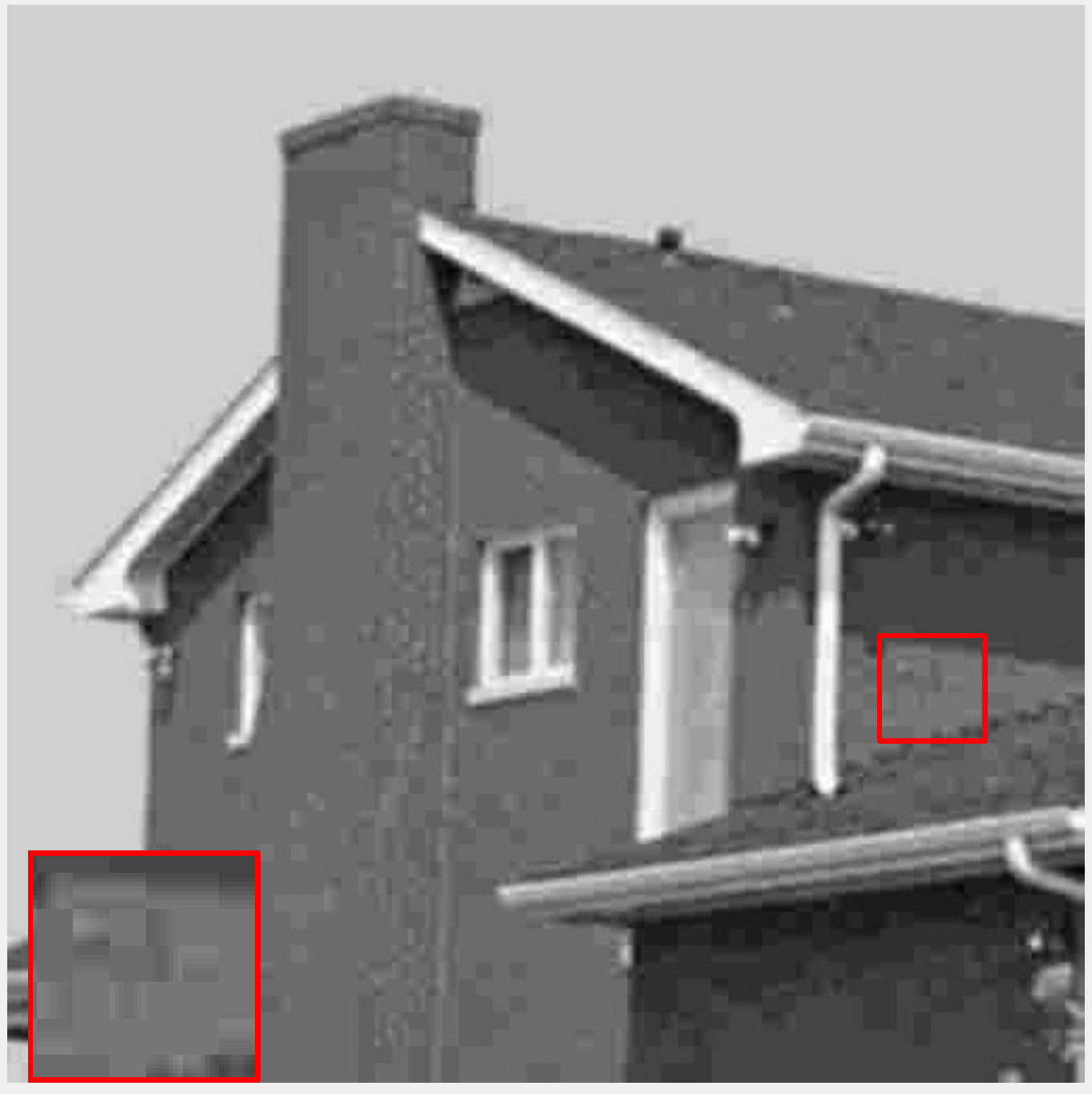}}
    \hfill
    \subfloat[{\scriptsize Algorithm \ref{alg:Proposed}.}]{\includegraphics[height=0.245\linewidth,keepaspectratio]{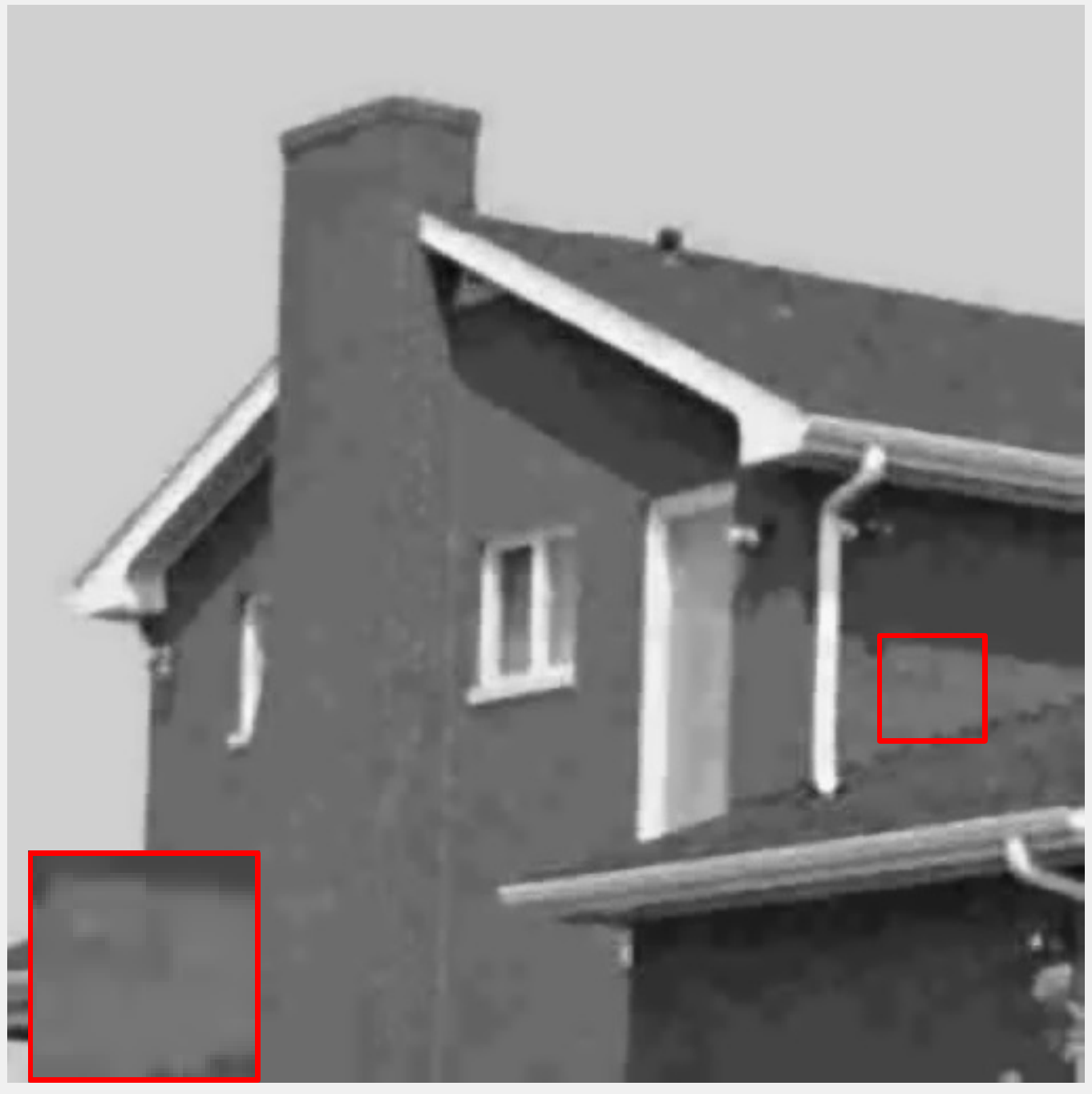}}
    \hfill
    \subfloat[{\scriptsize Brute force.}]{\includegraphics[height=0.245\linewidth,keepaspectratio]{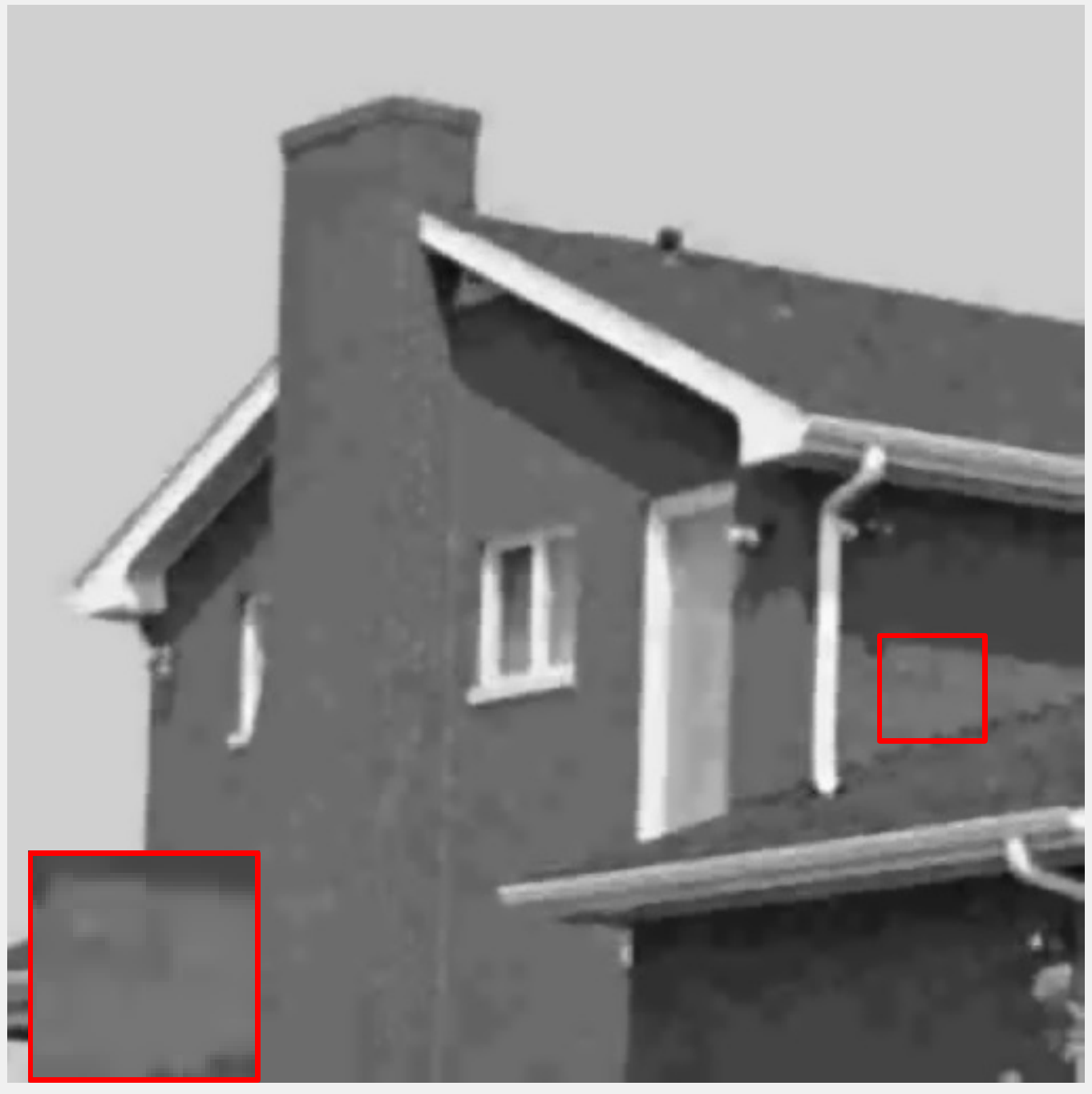}}\\
    \subfloat[{\scriptsize \cite{Li2014}.}]{\includegraphics[height=0.245\linewidth,keepaspectratio]{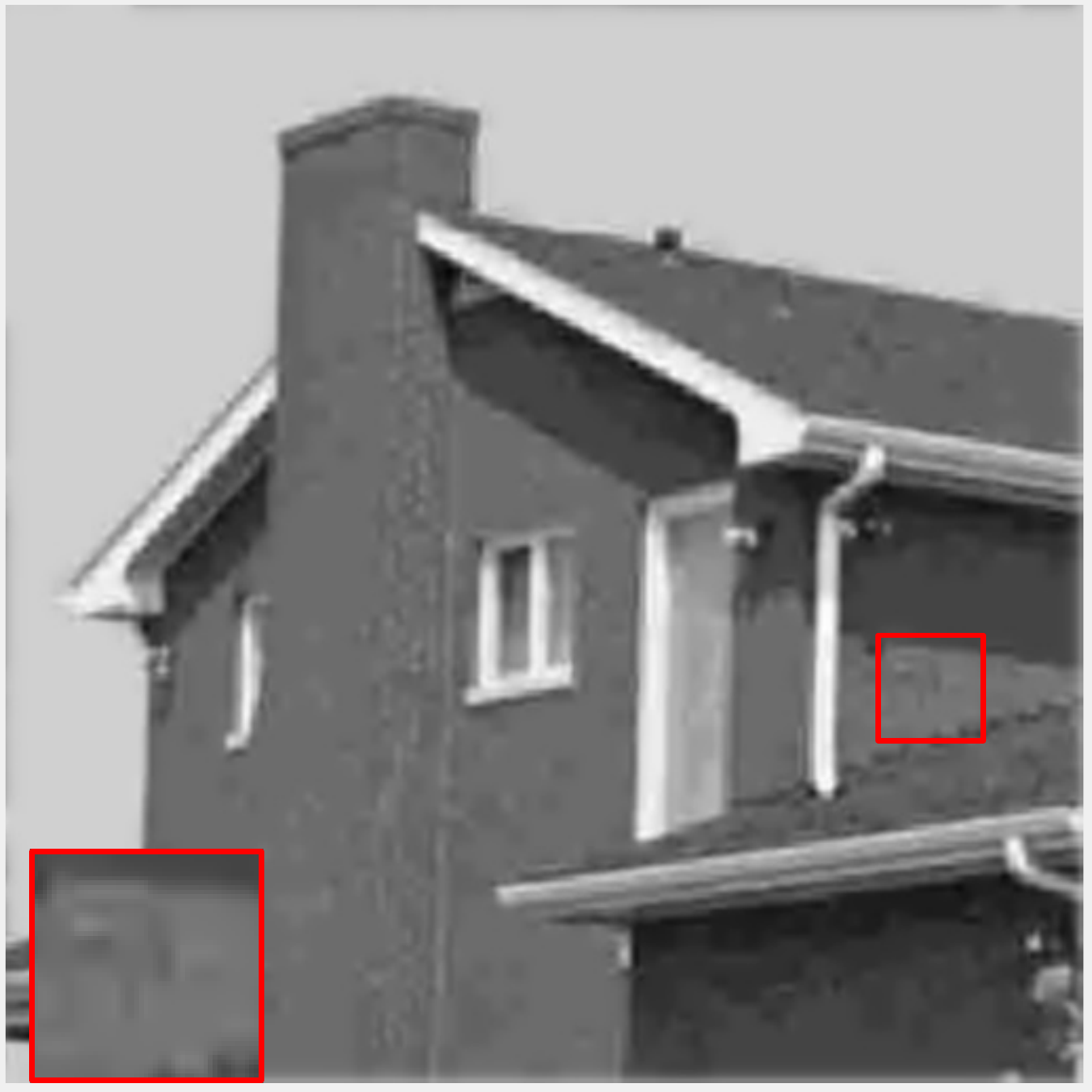}}
    \hfill
    \subfloat[{\scriptsize \cite{Zhang2016_concolor}.}]{\includegraphics[height=0.245\linewidth,keepaspectratio]{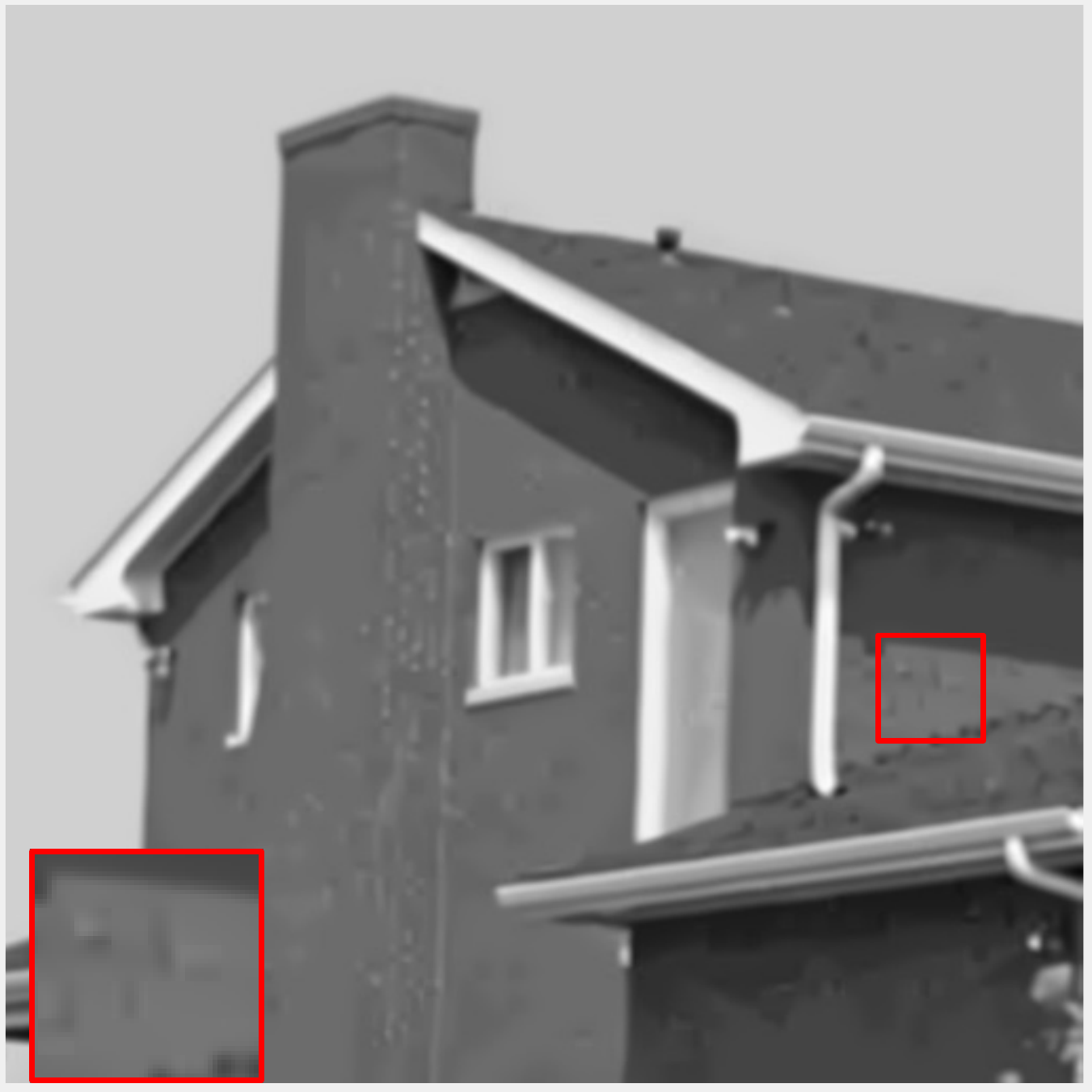}}
    \hfill
    \subfloat[{\scriptsize \cite{Zhao2017}.}]{\includegraphics[height=0.245\linewidth,keepaspectratio]{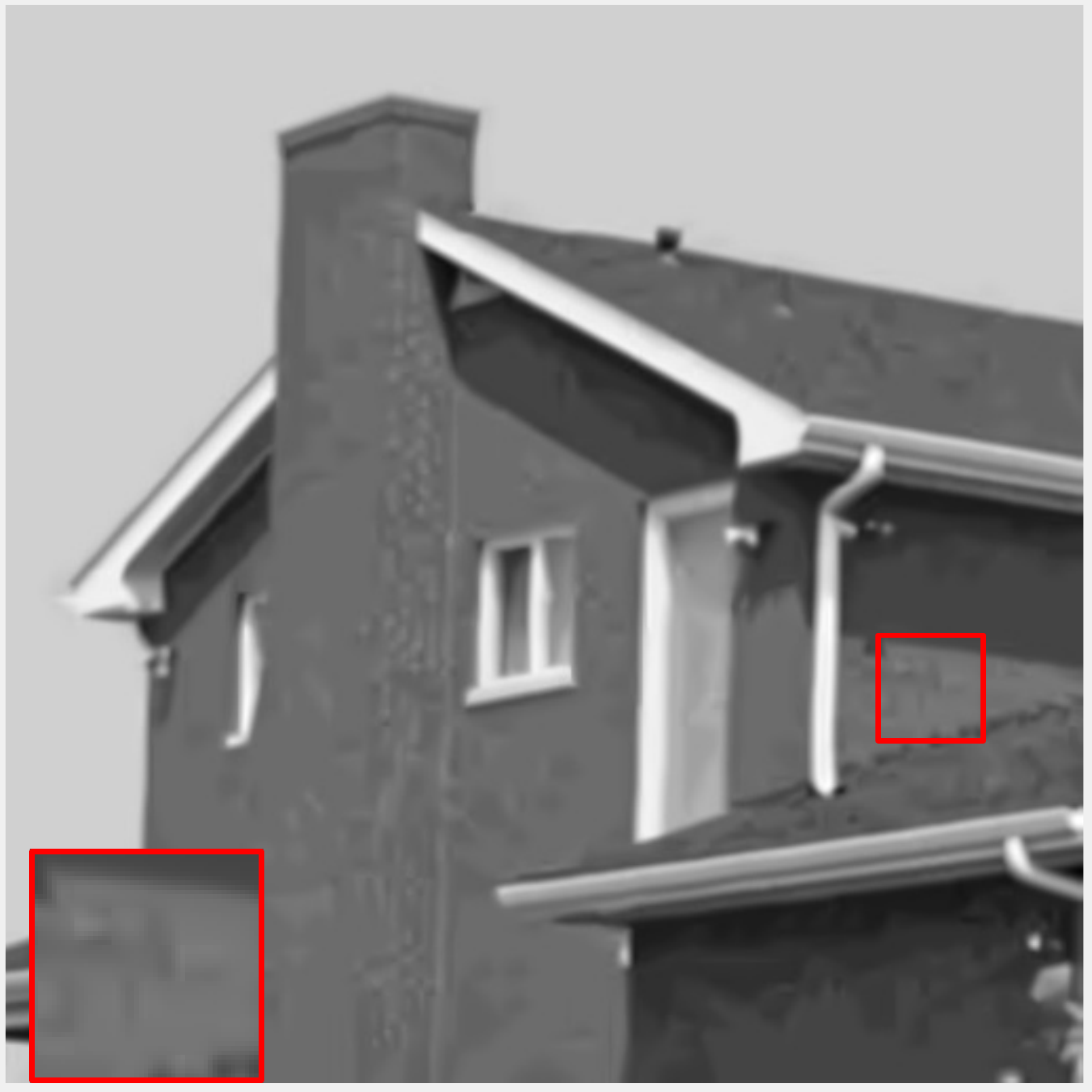}}
    \hfill
    \subfloat[{\scriptsize \cite{Golestaneh2014}.}]{\includegraphics[height=0.245\linewidth,keepaspectratio]{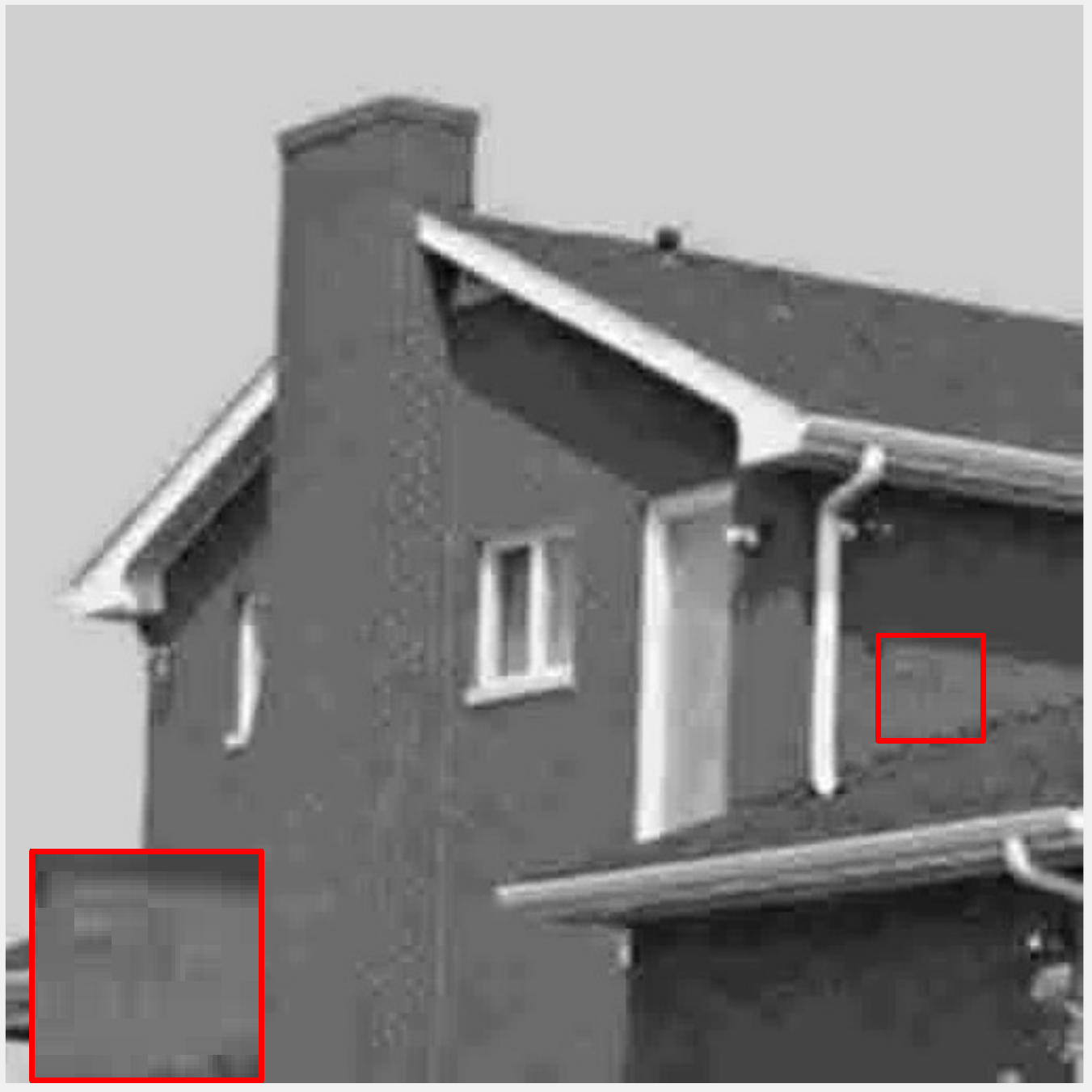}}
    \caption{JPEG deblocking results (image size $512 \times 512$). The (Runtime, PSNR, SSIM) values relative to the uncompressed image (a) are as follows (PSNR in dB):
    (b) $-$, $33.77$, $0.8937$;
    (c) $0.15$s, $34.44$, $0.9127$;
    (d) $3.3$s, $34.49$, $0.9136$;
    (e) $0.030$s, $31.29$, $0.9094$;
    (f) $20$ min, $35.94$, $0.9266$;
    (g) $90$s, $35.59$, $0.9238$;
    (h) $2.1$s, $33.76$, $0.8962$;
    The PSNR between (c) and (d) is $50.8$ dB.
    Please zoom in for a clearer view.}
    \label{fig:Artifact3}
\end{figure}

The deblocking outputs of this method are shown in Figures \ref{fig:Artifact2} and \ref{fig:Artifact3}.
For completeness we compare the results with \cite{Li2014}, \cite{Zhang2016_concolor}, \cite{Zhao2017}, and \cite{Golestaneh2014}
\footnote{Codes: \cite{Li2014}: \url{http://yu-li.github.io/}\\
\cite{Zhang2016_concolor}: \url{https://github.com/jianzhangcs/CONCOLOR}\\
\cite{Zhao2017}: \url{https://github.com/coolbay/Image-deblocking-SSRQC}}.
We note that \cite{Li2014} was originally intended for use in a contrast-enhancement pipeline, but its results are acceptable when used for deblocking.
The methods in \cite{Zhang2016_concolor} and \cite{Zhao2017} used an optimization model, hence their performance is the best among all the methods considered.
However both of them are quite slow.
Adaptive bilateral filtering (Algorithm \ref{alg:Proposed}) shown in (c) gives an acceptable performance at a much higher speed, while its PSNR is less by only about $0.8 \mbox{--} 1.5$ dB than \cite{Zhang2016_concolor} and \cite{Zhao2017}.
We note that Algorithm \ref{alg:Proposed} is few orders faster than \cite{Zhang2016_concolor,Zhao2017}.
The method in \cite{Li2014} is faster than ours, but the deblocking using Algorithm \ref{alg:Proposed} is superior.
Compared to \cite{Golestaneh2014}, Algorithm \ref{alg:Proposed} is superior in terms of both runtime and performance.
Additionally, it is about $20 \times$ faster than brute-force adaptive bilateral filtering (shown in (d)).
We found that adaptive bilateral filtering performs well for low compression rates.
Accordingly, we have used a JPEG quality factor of $10\%$ for compressing the images in Figures \ref{fig:Artifact2} and \ref{fig:Artifact3}.

\subsection{Texture filtering}

\begin{figure}[t!]
    \centering
	\subfloat[Input image.]{\includegraphics[height=0.30\linewidth,keepaspectratio]{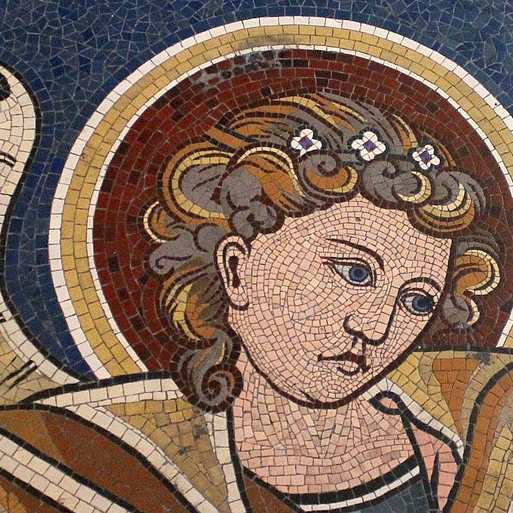}}
    \hspace{0.1mm}
    \subfloat[$\sigma$ map.]{\includegraphics[height=0.30\linewidth,keepaspectratio]{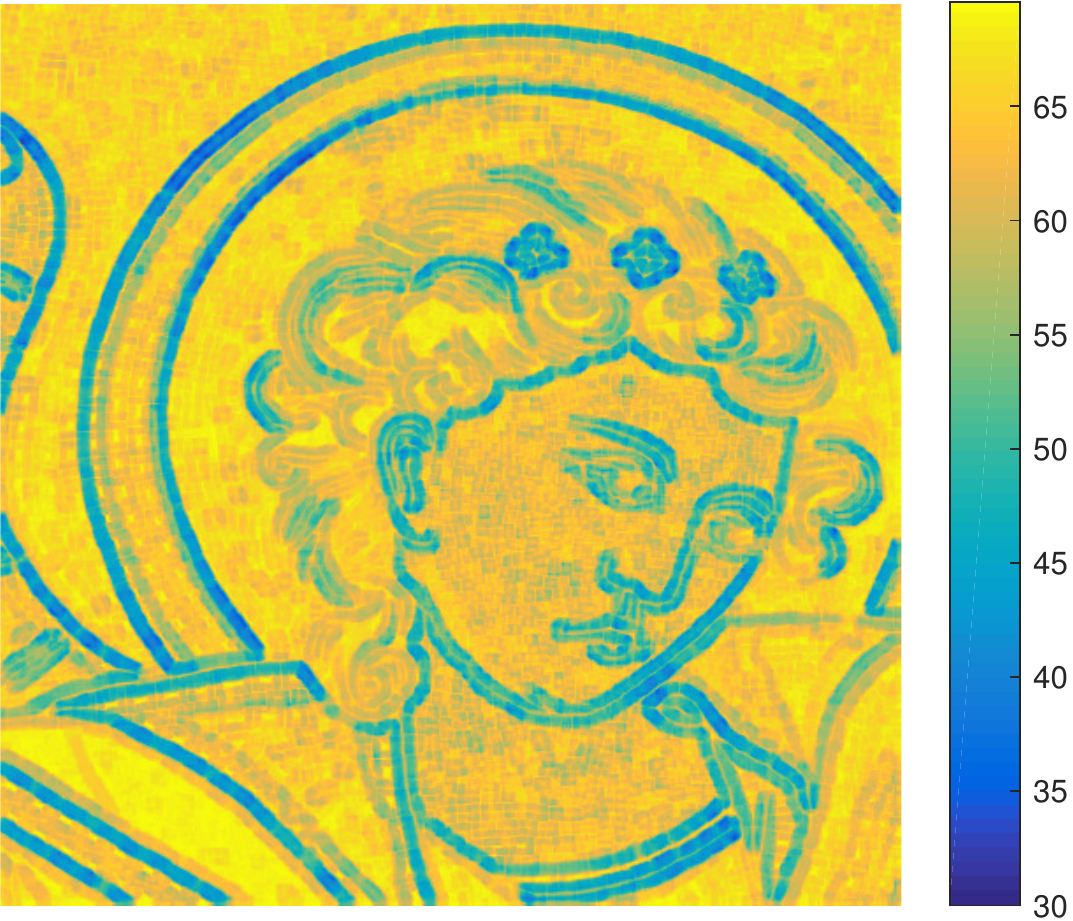}}
    \caption{Computation of $\sigma(i)$ for texture filtering (see text for details).}
    \label{fig:Texture1}
\end{figure}

\begin{figure*}[t!]
	\captionsetup[subfigure]{labelformat=empty}
    \centering
	\subfloat[$600 \times 450$.]{\includegraphics[width=0.124\linewidth,keepaspectratio]{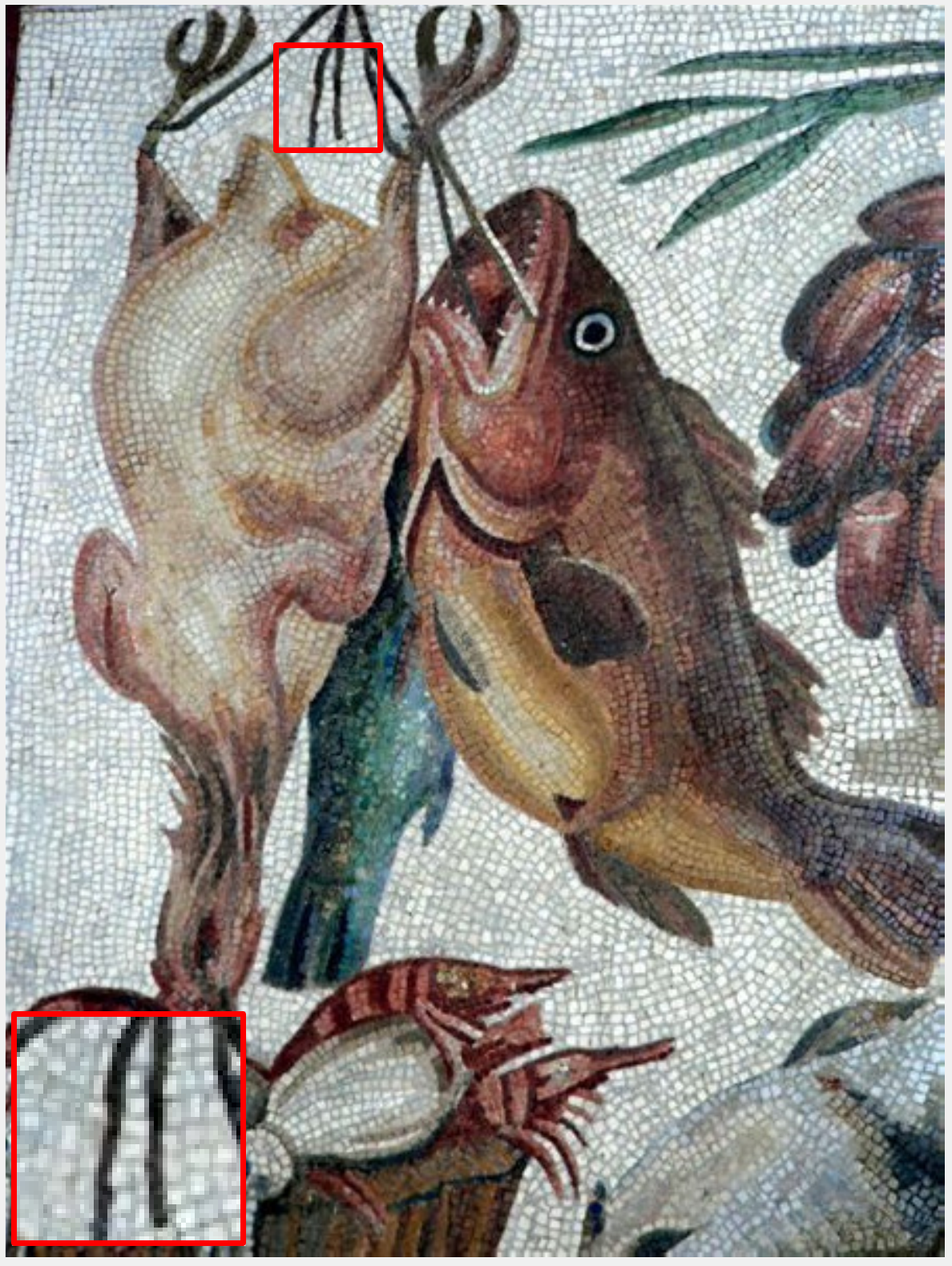}}
    \hfill
    \subfloat[$1.9$ s.]{\includegraphics[width=0.124\linewidth,keepaspectratio]{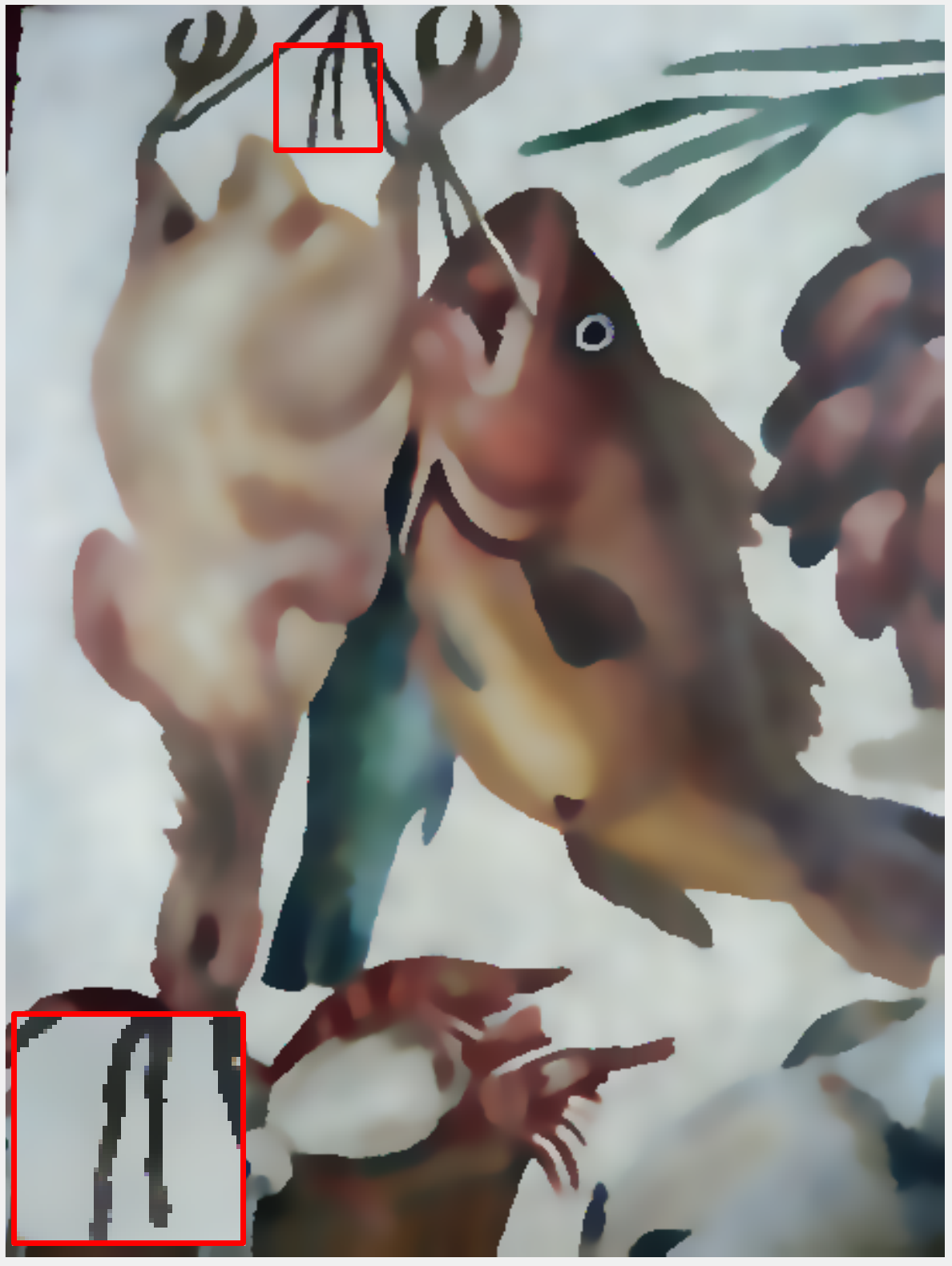}}
    \hfill
    \subfloat[$11$ s.]{\includegraphics[width=0.124\linewidth,keepaspectratio]{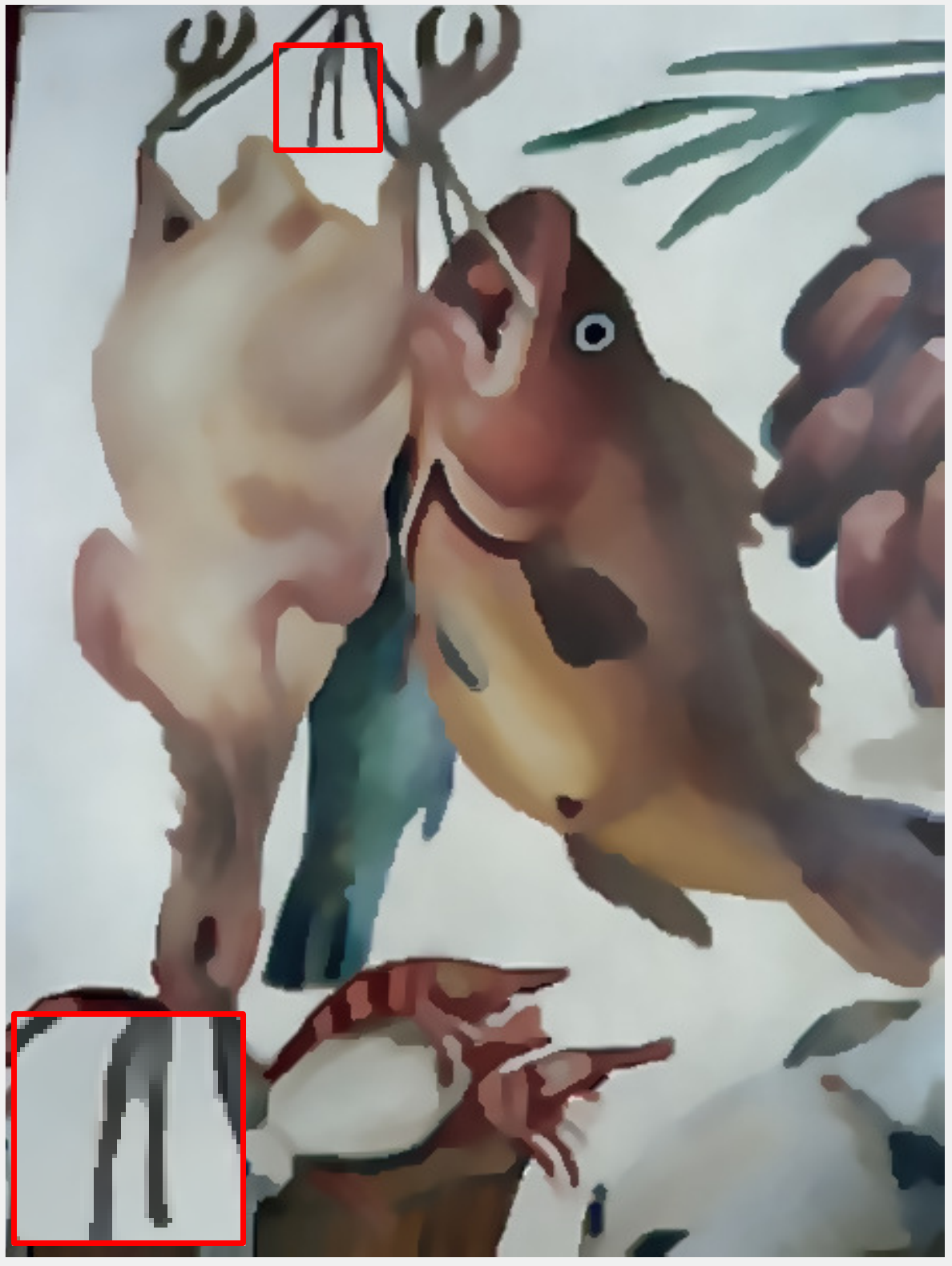}}
    \hfill
	\subfloat[$4.1$ s.]{\includegraphics[width=0.124\linewidth,keepaspectratio]{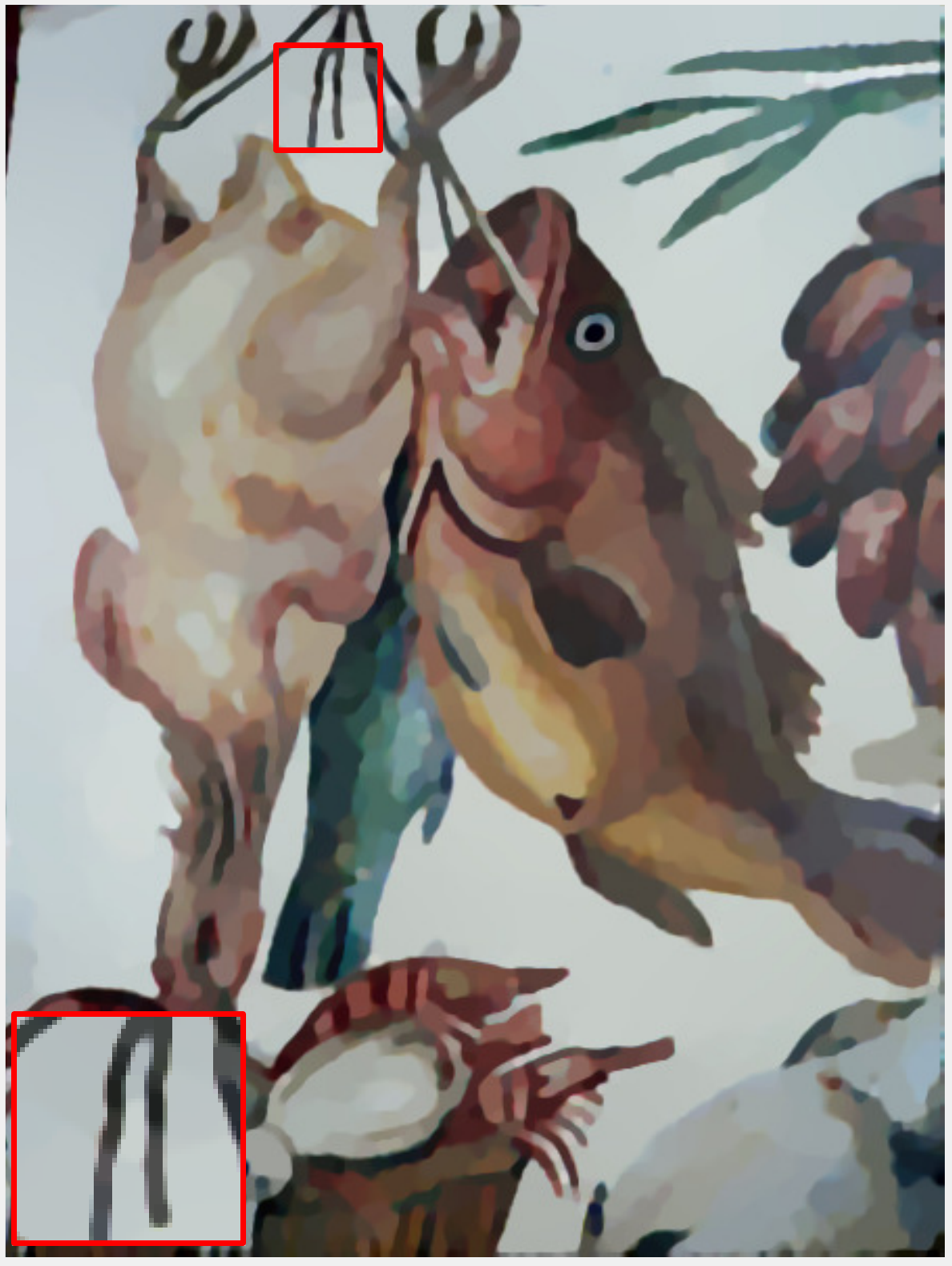}}
	\hfill
	\subfloat[$50$ s.]{\includegraphics[width=0.124\linewidth,keepaspectratio]{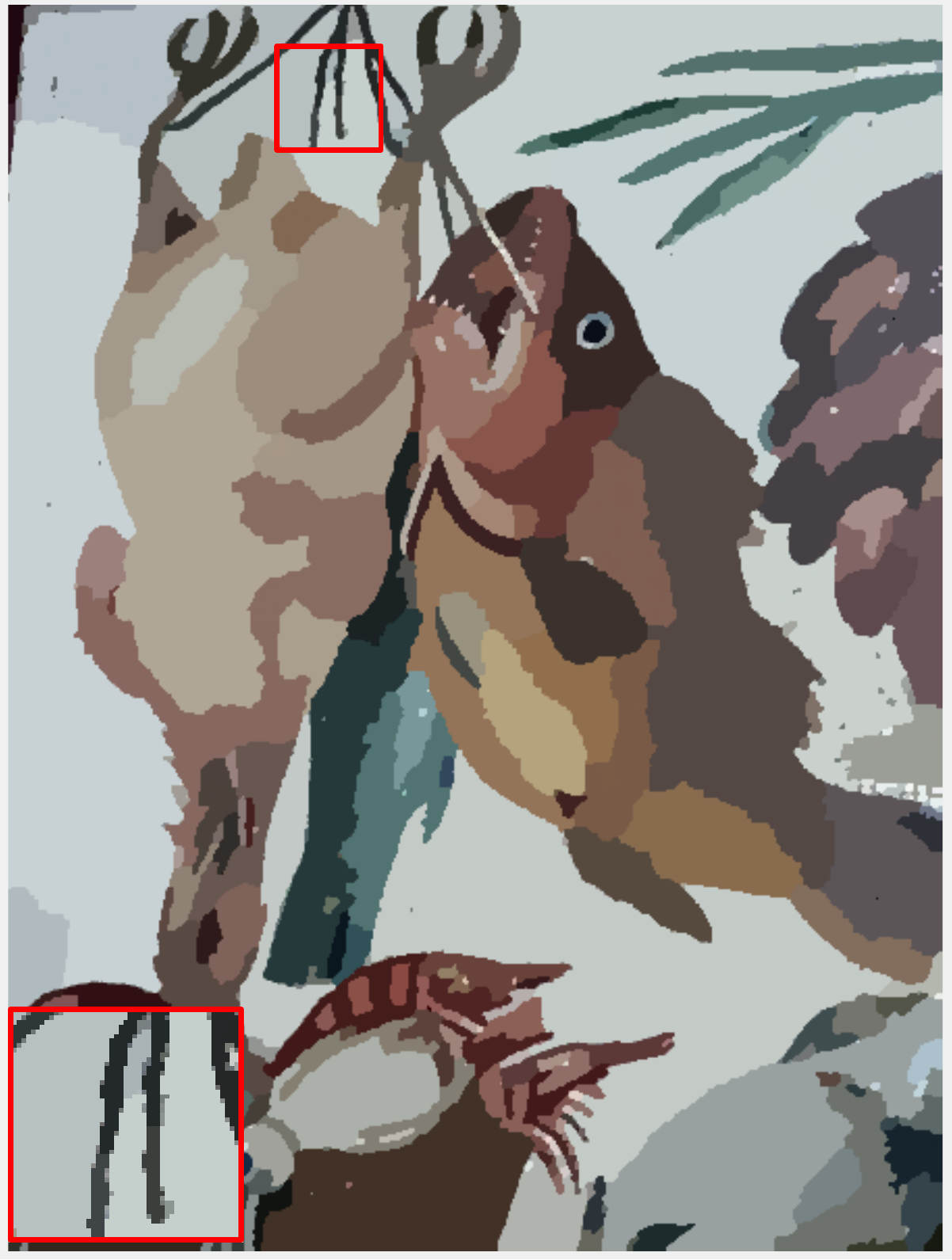}}
	\hfill
    \subfloat[$0.69$ s.]{\includegraphics[width=0.124\linewidth,keepaspectratio]{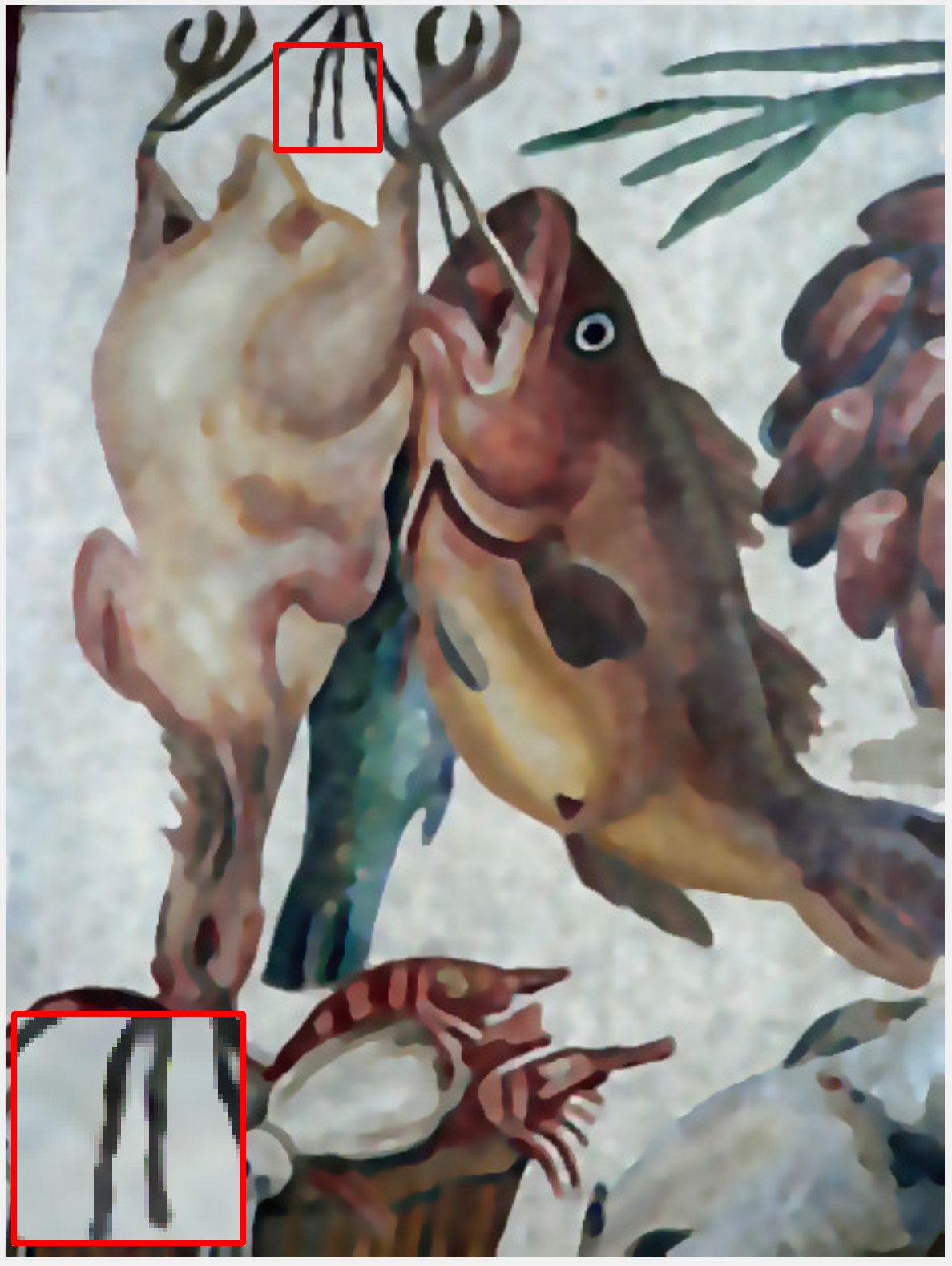}}
    \hfill
    \subfloat[ $7.9$ s.]{\includegraphics[width=0.124\linewidth,keepaspectratio]{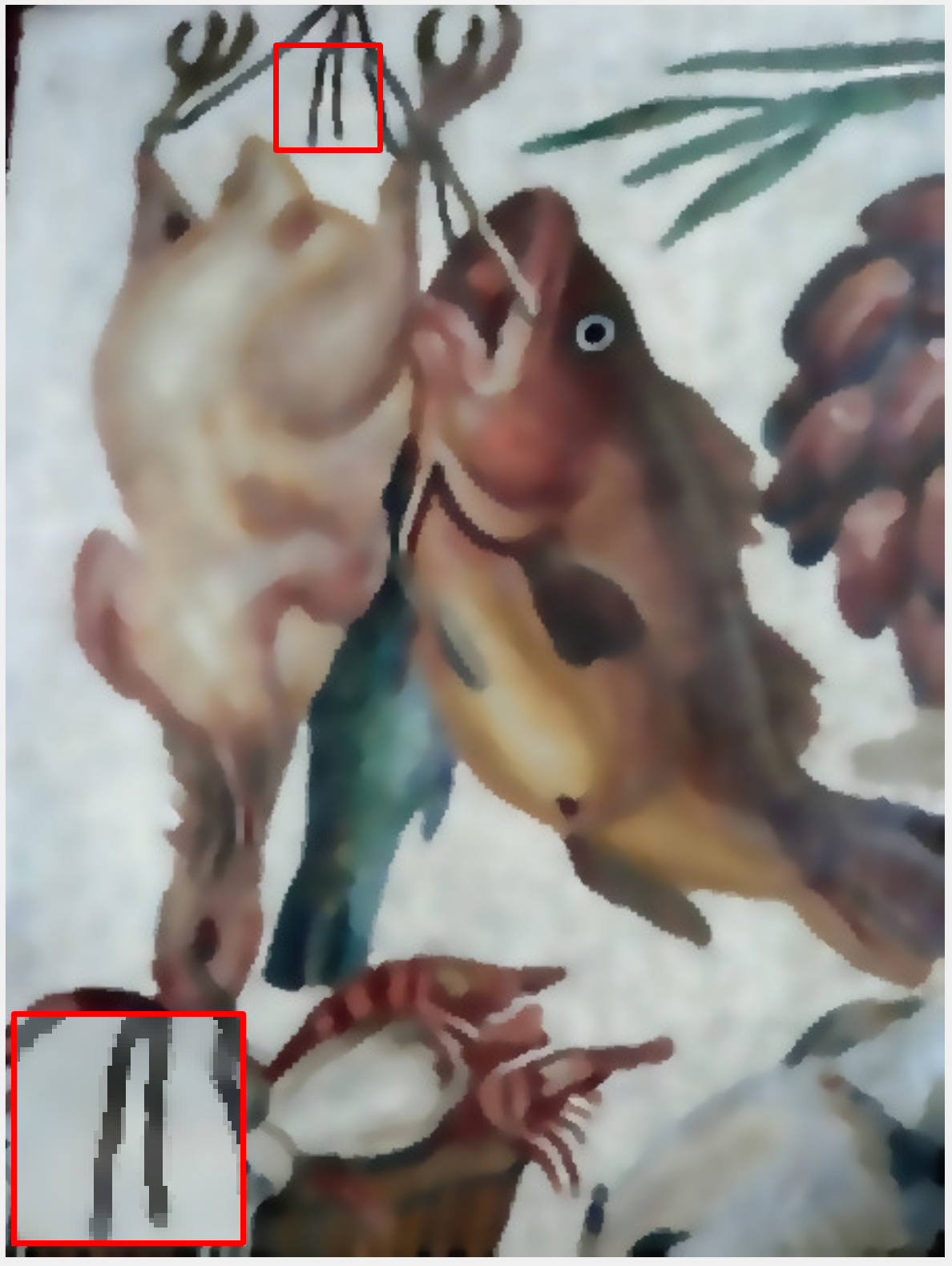}}
    \hfill
    \subfloat[$45$ s.]{\includegraphics[width=0.124\linewidth,keepaspectratio]{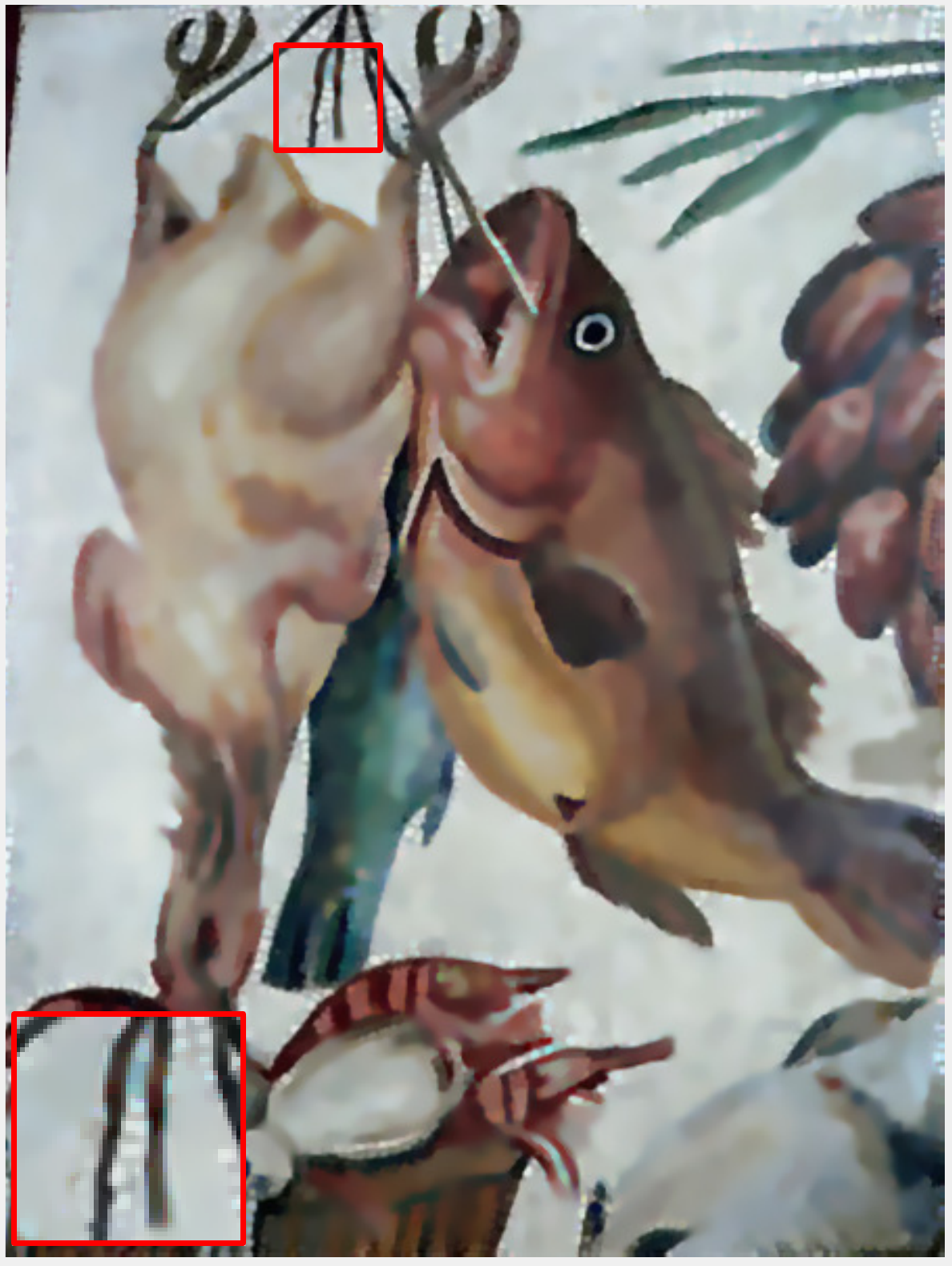}}

	\subfloat[$512 \times 512$.]{\includegraphics[width=0.124\linewidth,keepaspectratio]{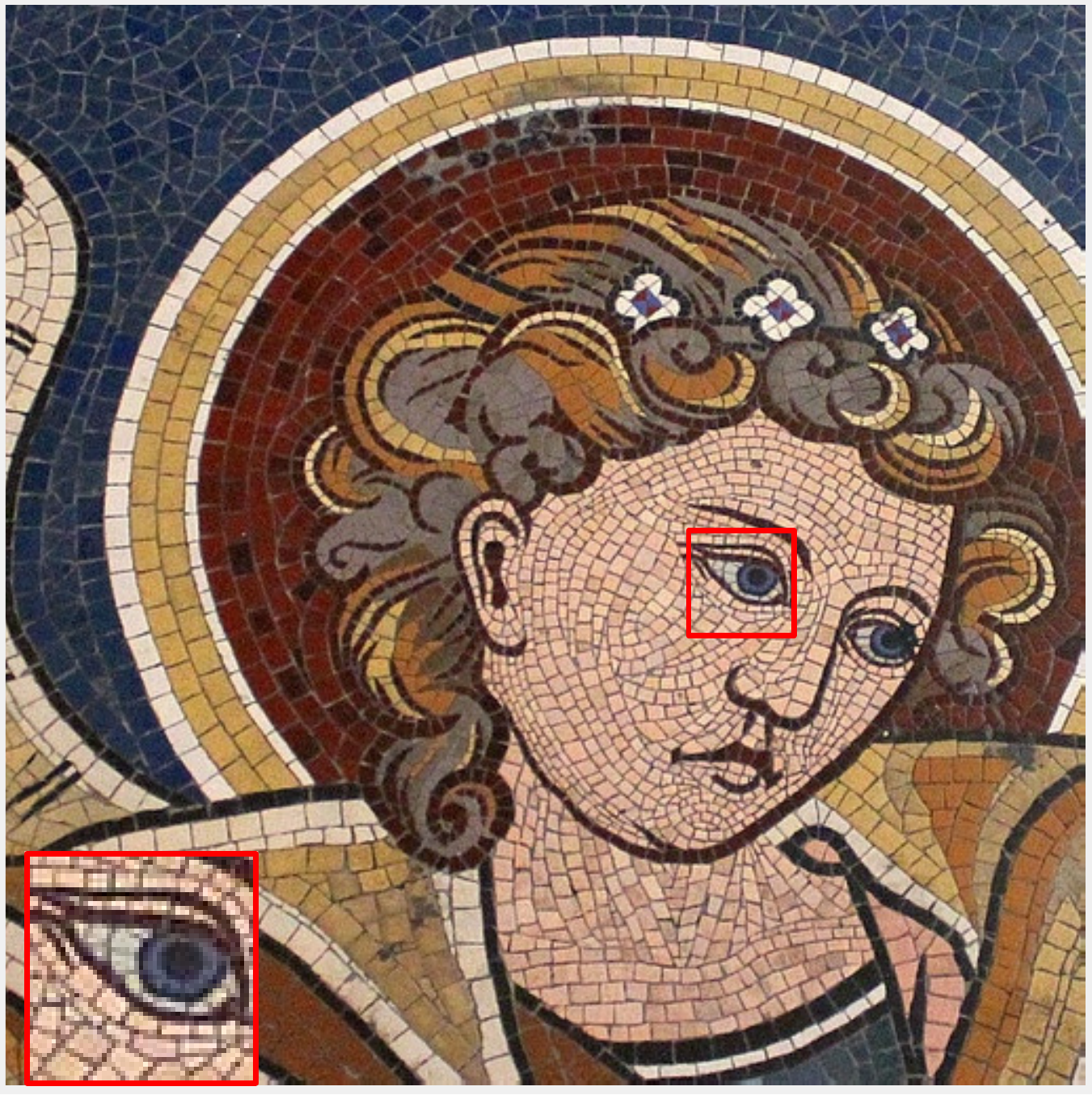}}
    \hfill
    \subfloat[$1.7$ s.]{\includegraphics[width=0.124\linewidth,keepaspectratio]{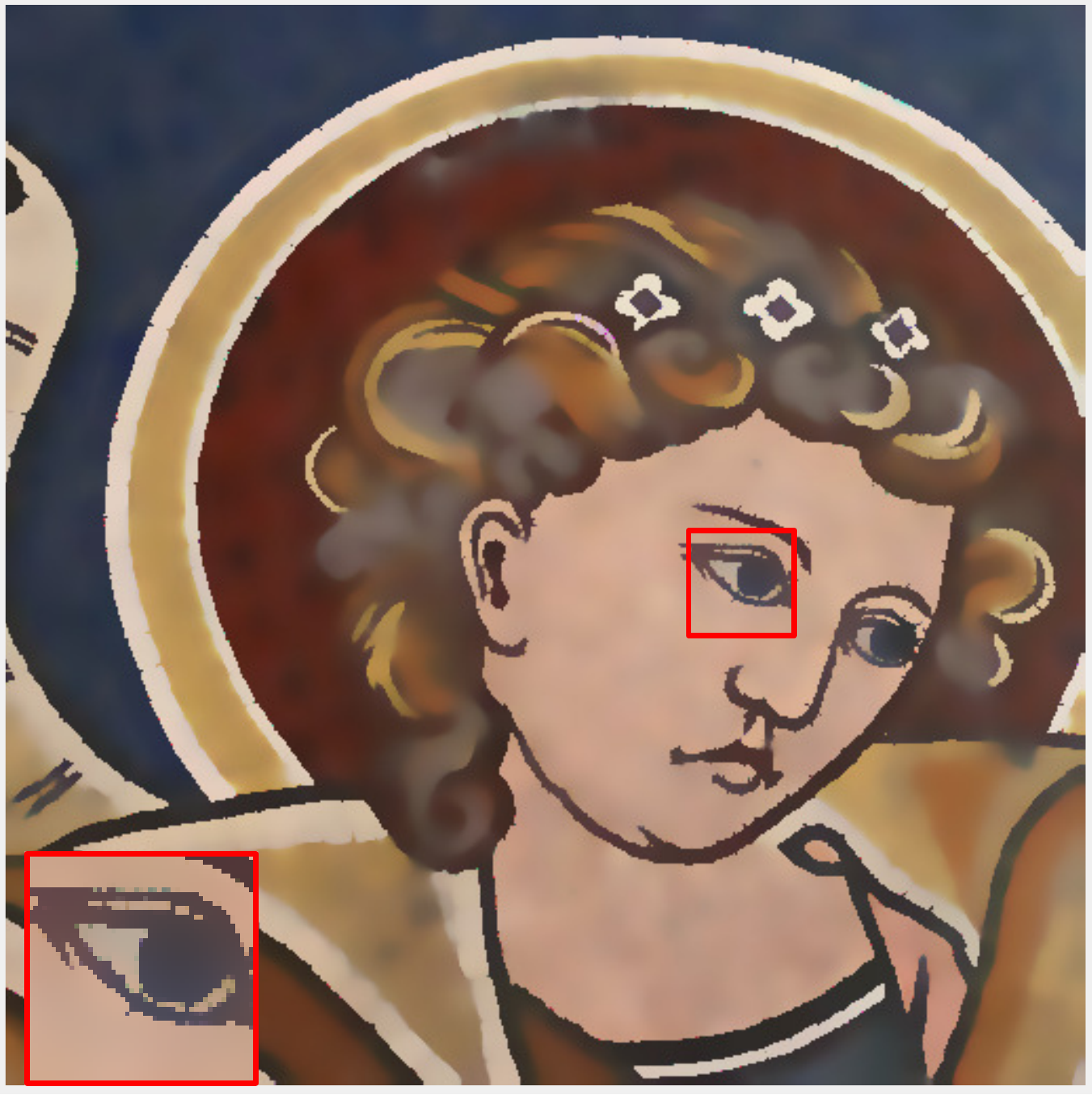}}
    \hfill
    \subfloat[$12$ s.]{\includegraphics[width=0.124\linewidth,keepaspectratio]{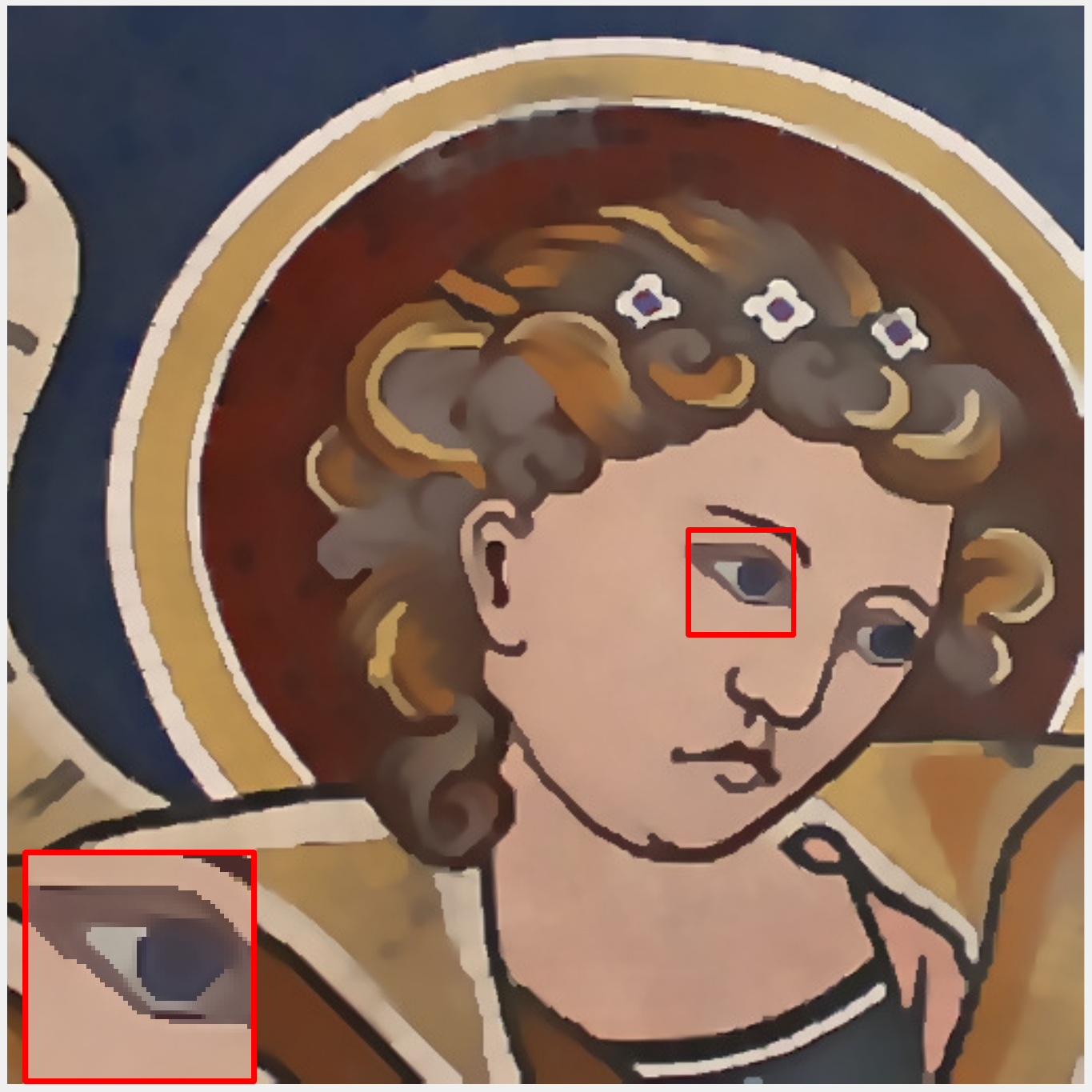}}
    \hfill
	\subfloat[$4.4$ s.]{\includegraphics[width=0.124\linewidth,keepaspectratio]{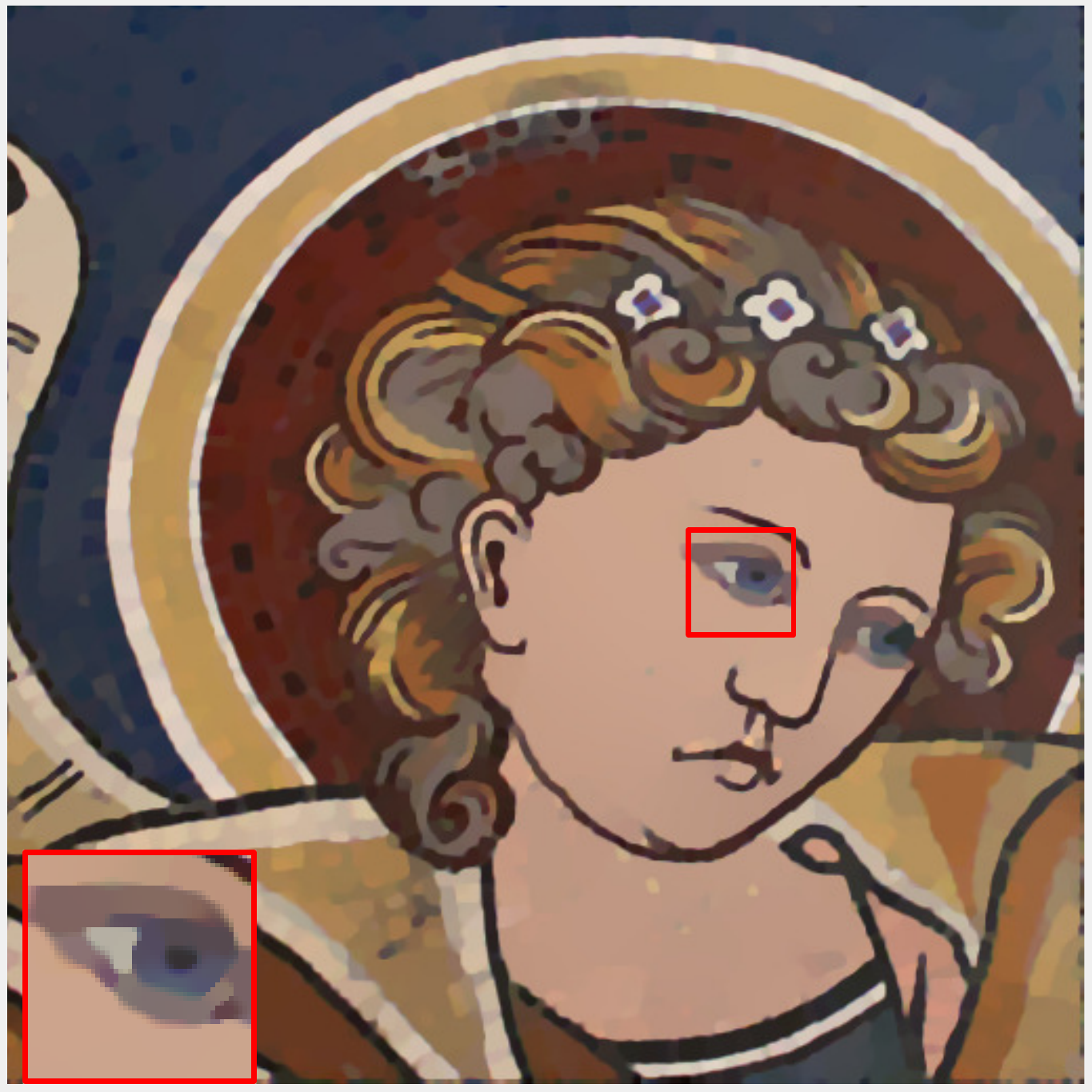}}
	\hfill
	\subfloat[$45$ s.]{\includegraphics[width=0.124\linewidth,keepaspectratio]{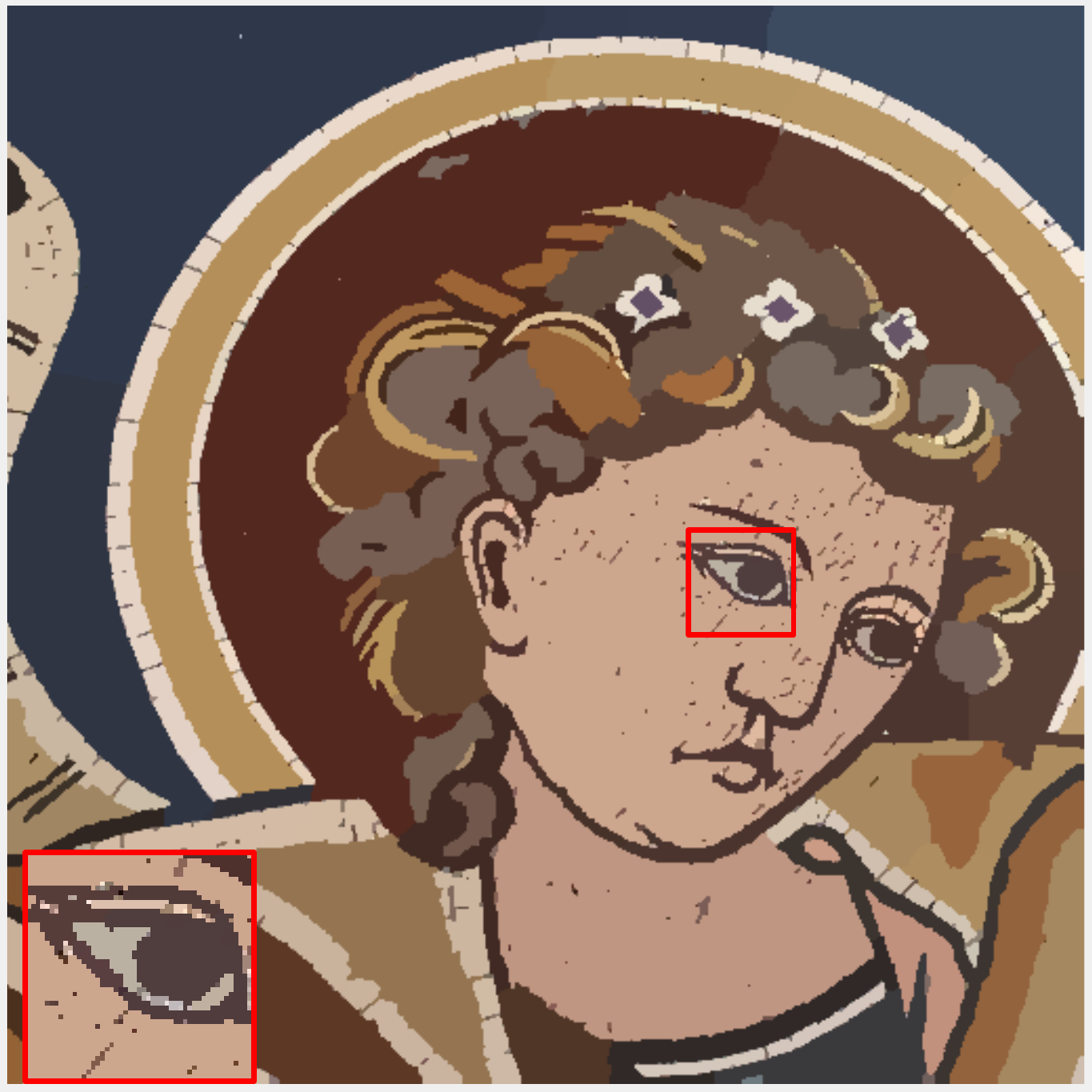}}
	\hfill
    \subfloat[$0.52$ s.]{\includegraphics[width=0.124\linewidth,keepaspectratio]{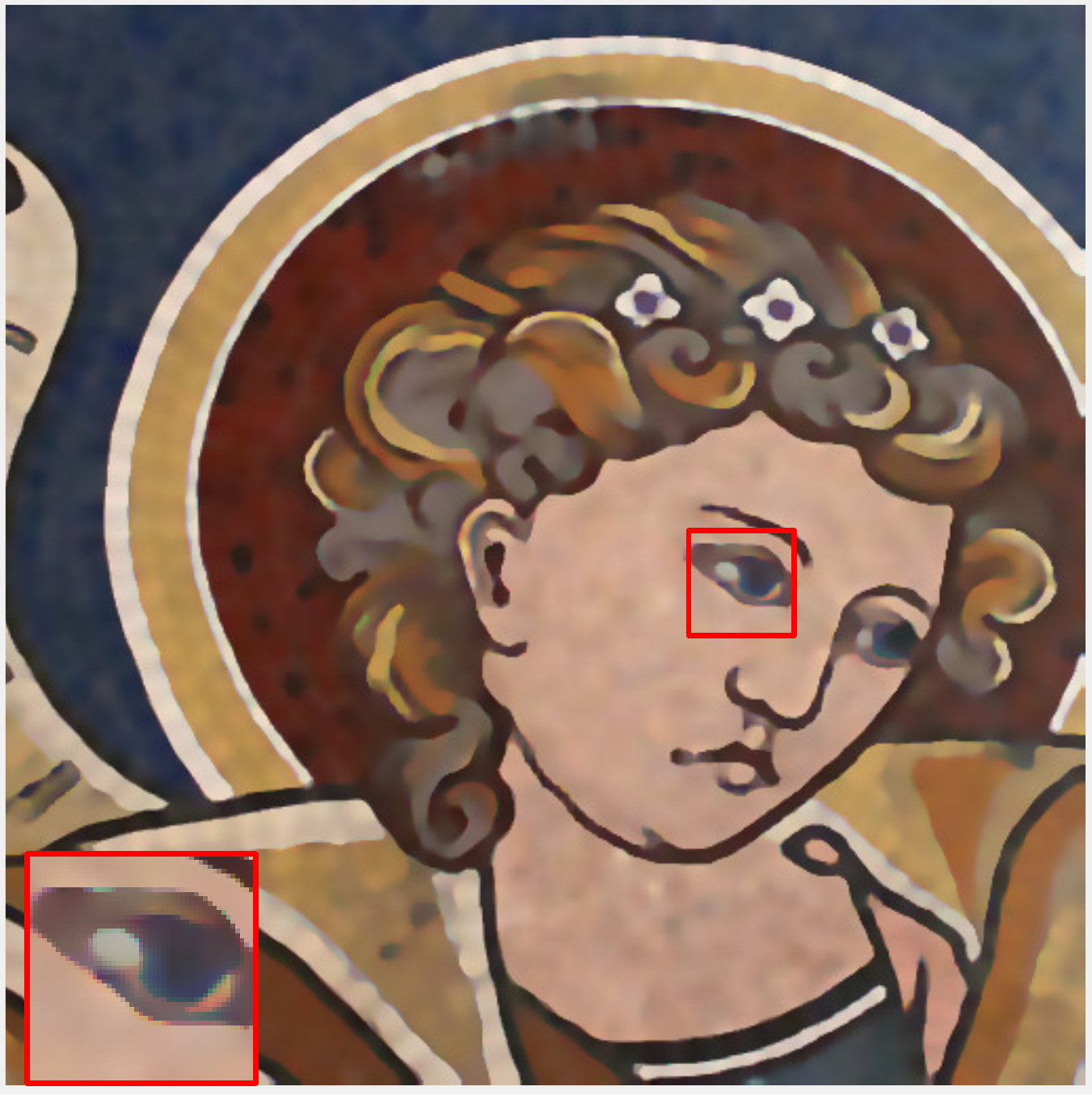}}
    \hfill
    \subfloat[$6.4$ s.]{\includegraphics[width=0.124\linewidth,keepaspectratio]{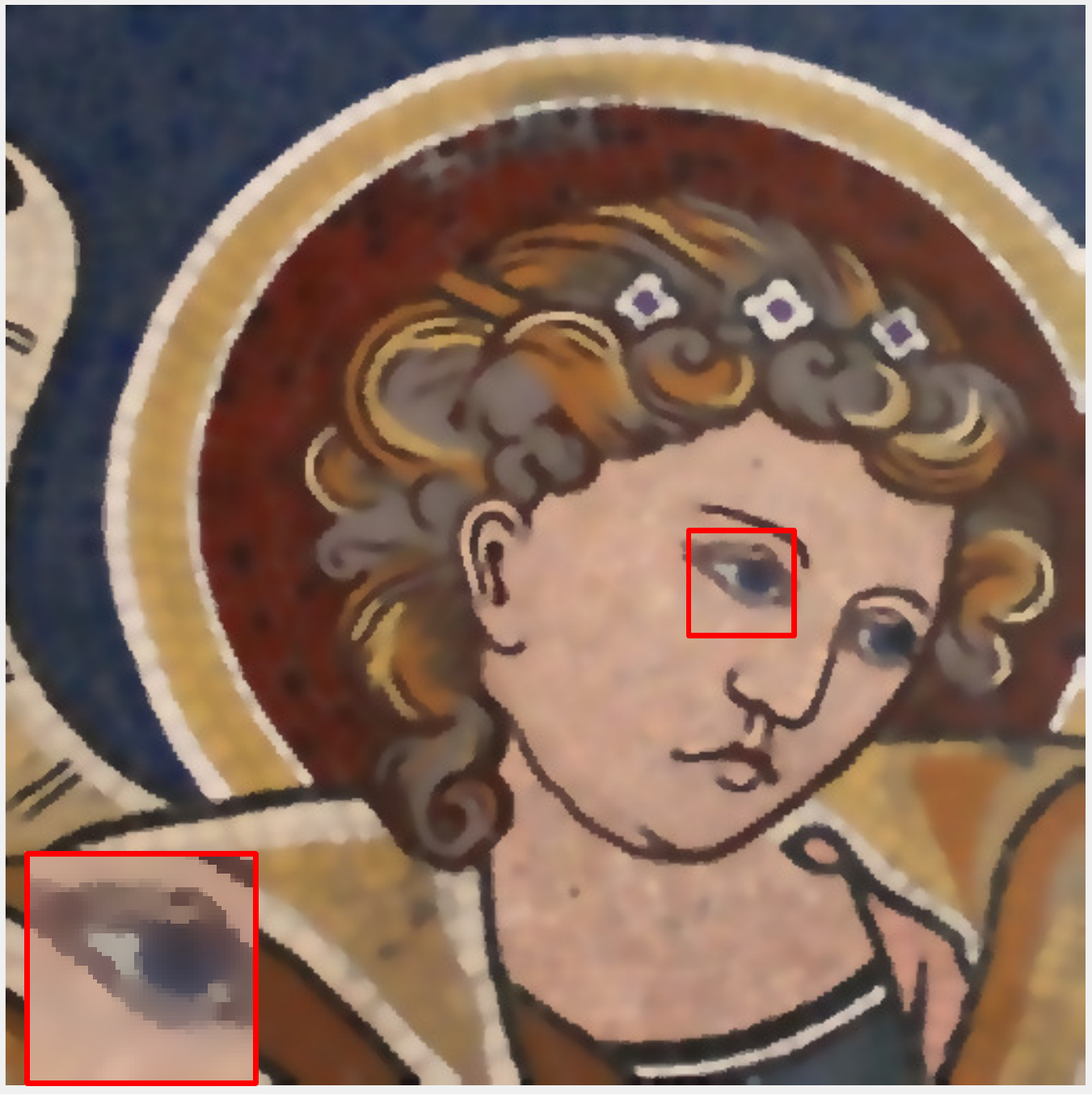}}
    \hfill
    \subfloat[$48$ s.]{\includegraphics[width=0.124\linewidth,keepaspectratio]{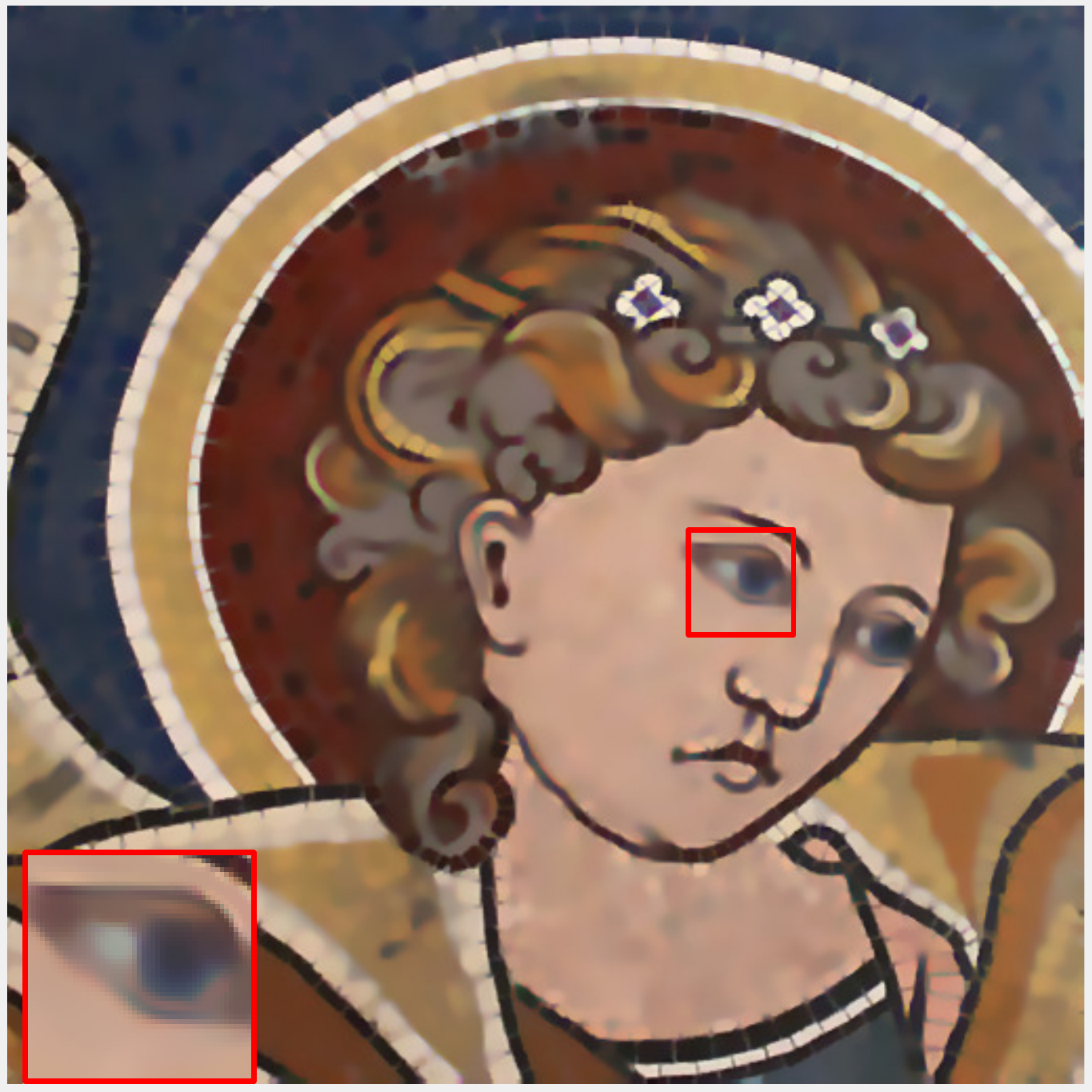}}

	\subfloat[$500 \times 650$.]{\includegraphics[width=0.124\linewidth,keepaspectratio]{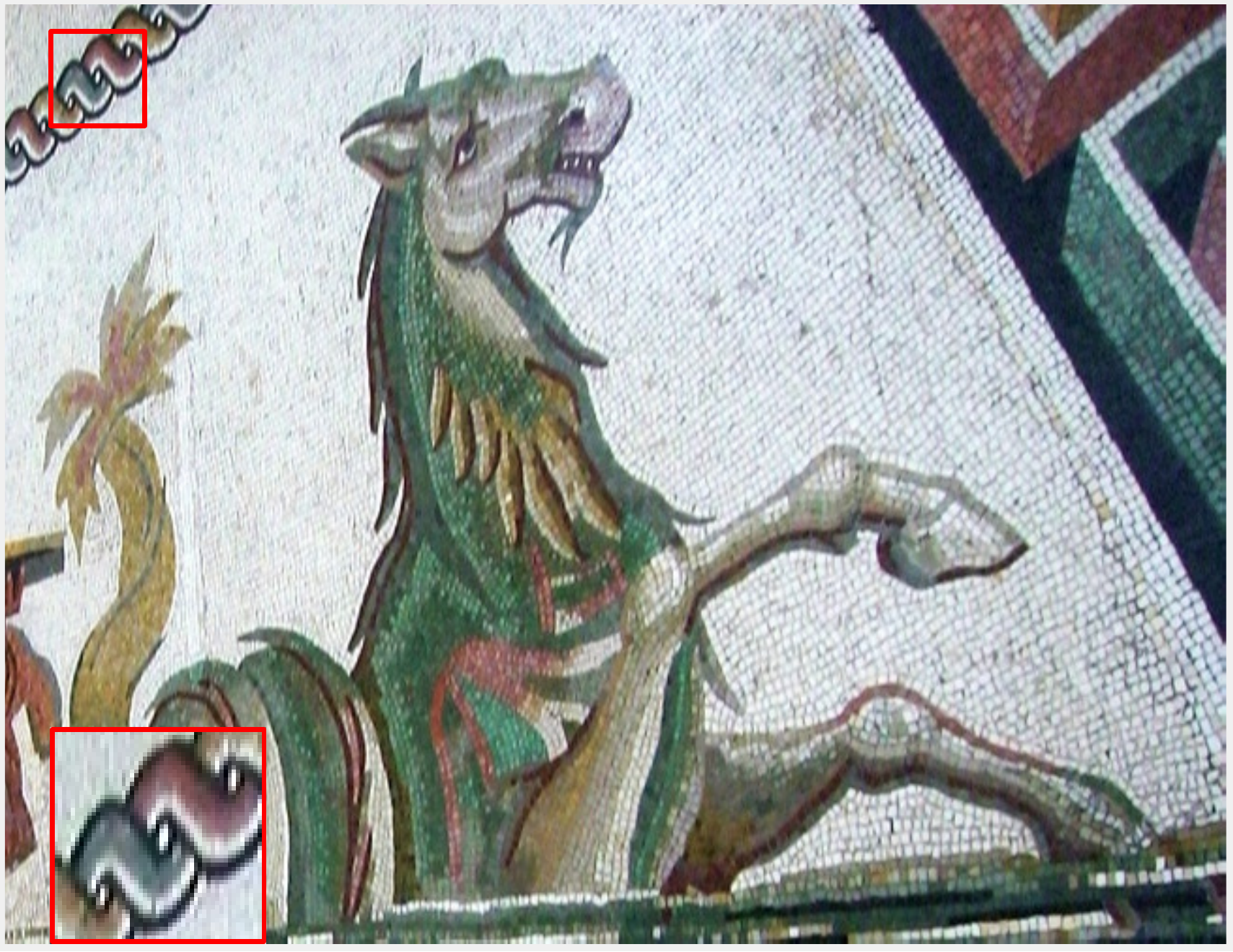}}
    \hfill
    \subfloat[$2.5$ s.]{\includegraphics[width=0.124\linewidth,keepaspectratio]{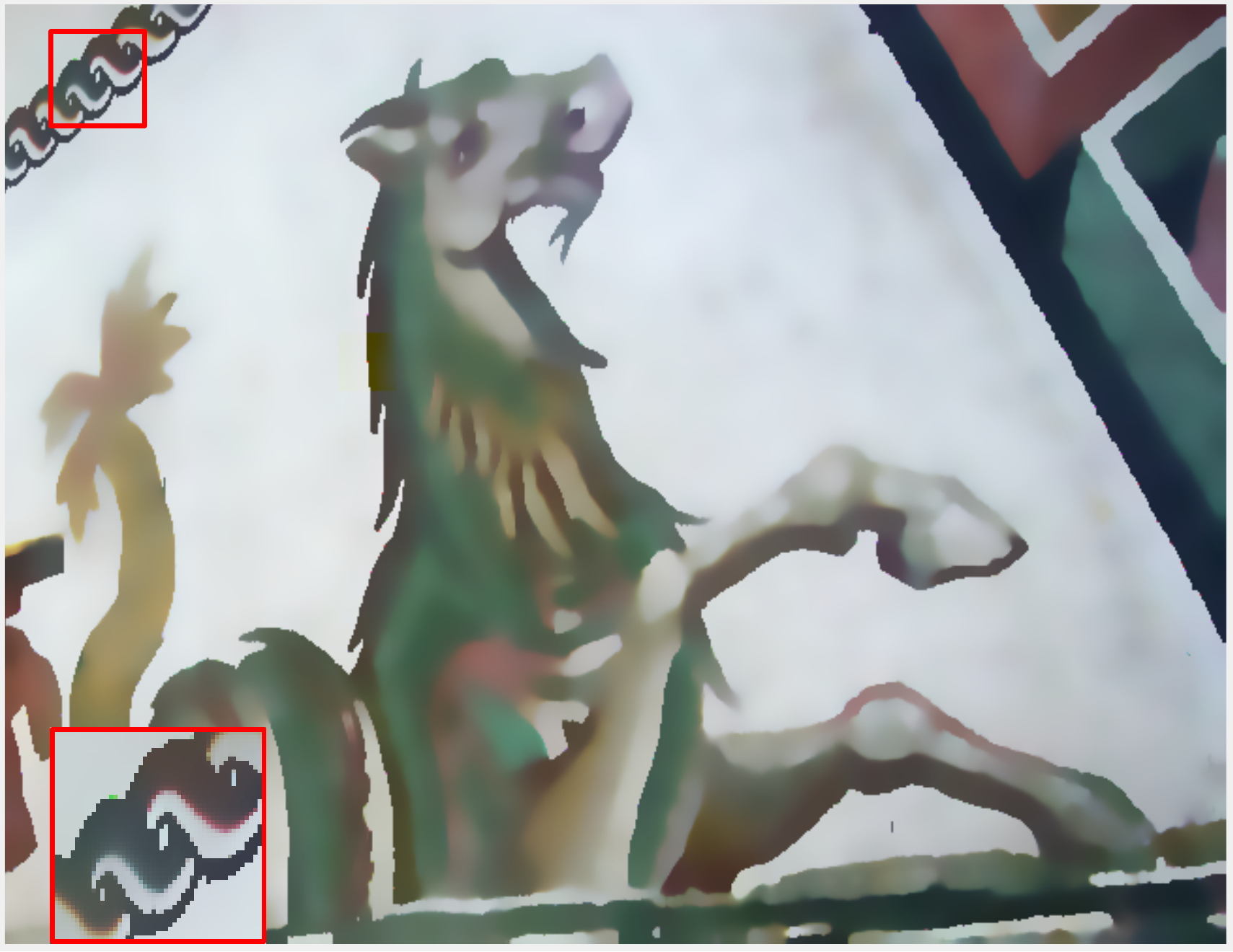}}
    \hfill
    \subfloat[$12.8$ s.]{\includegraphics[width=0.124\linewidth,keepaspectratio]{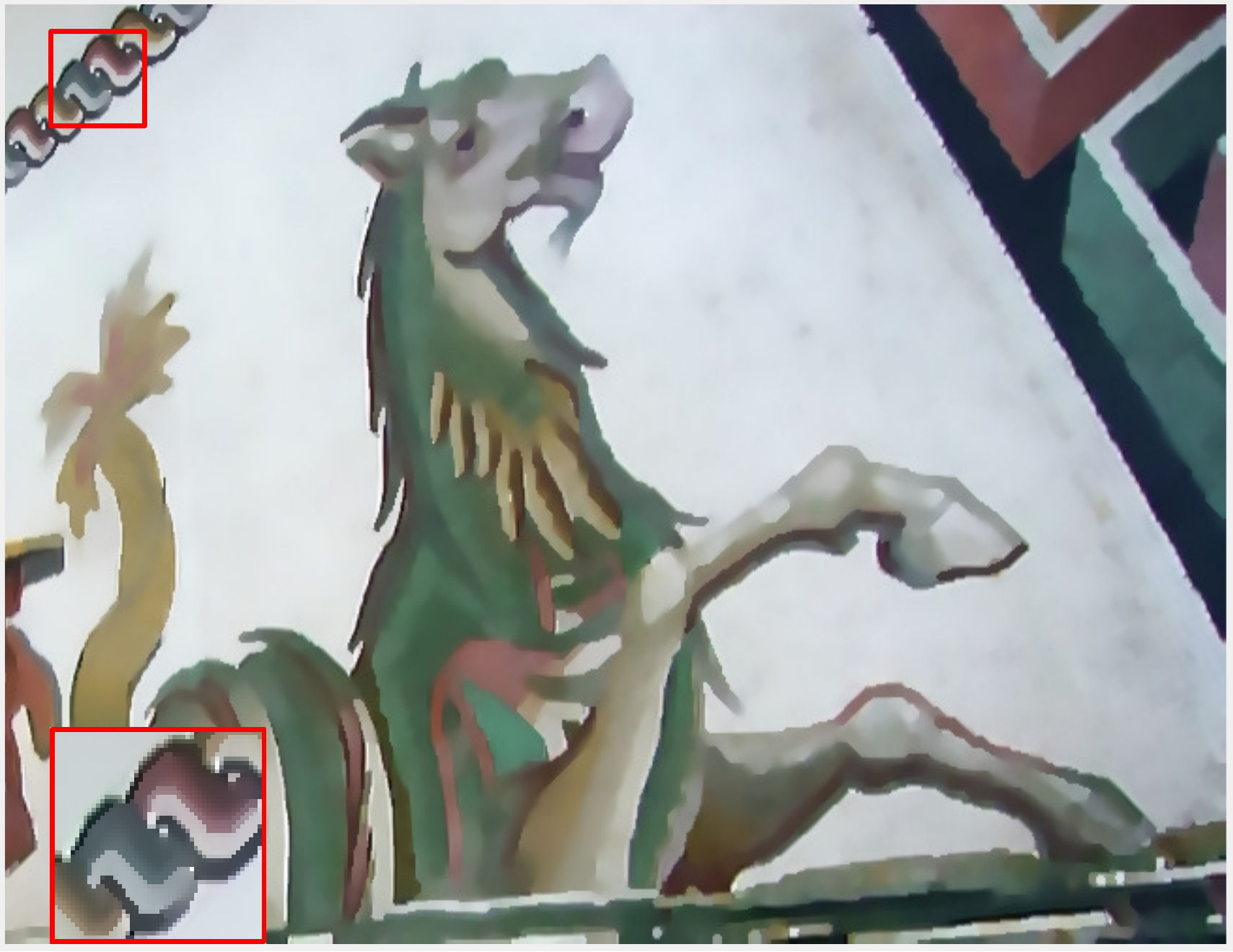}}
    \hfill
	\subfloat[$4.8$ s.]{\includegraphics[width=0.124\linewidth,keepaspectratio]{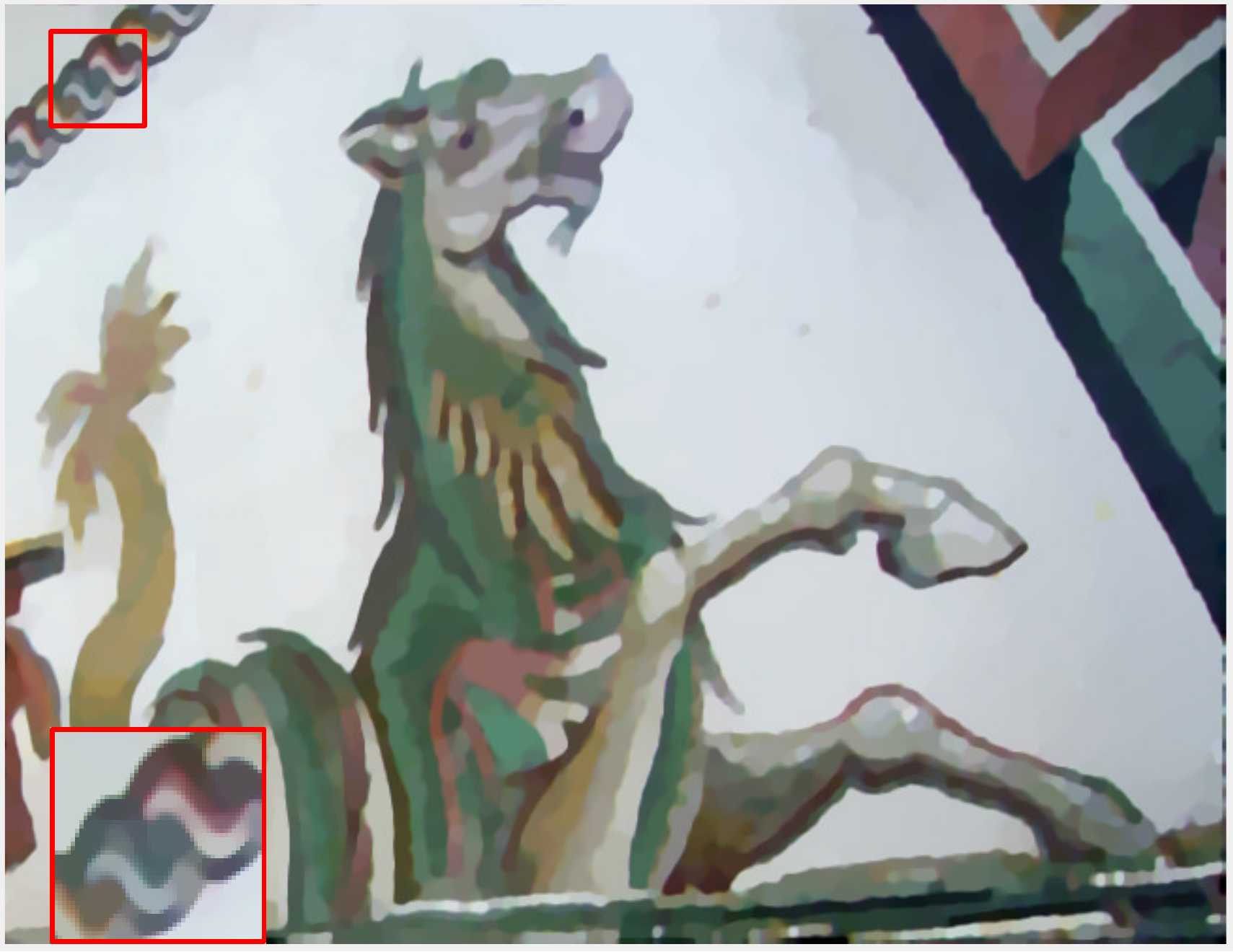}}
	\hfill
	\subfloat[$60$ s.]{\includegraphics[width=0.124\linewidth,keepaspectratio]{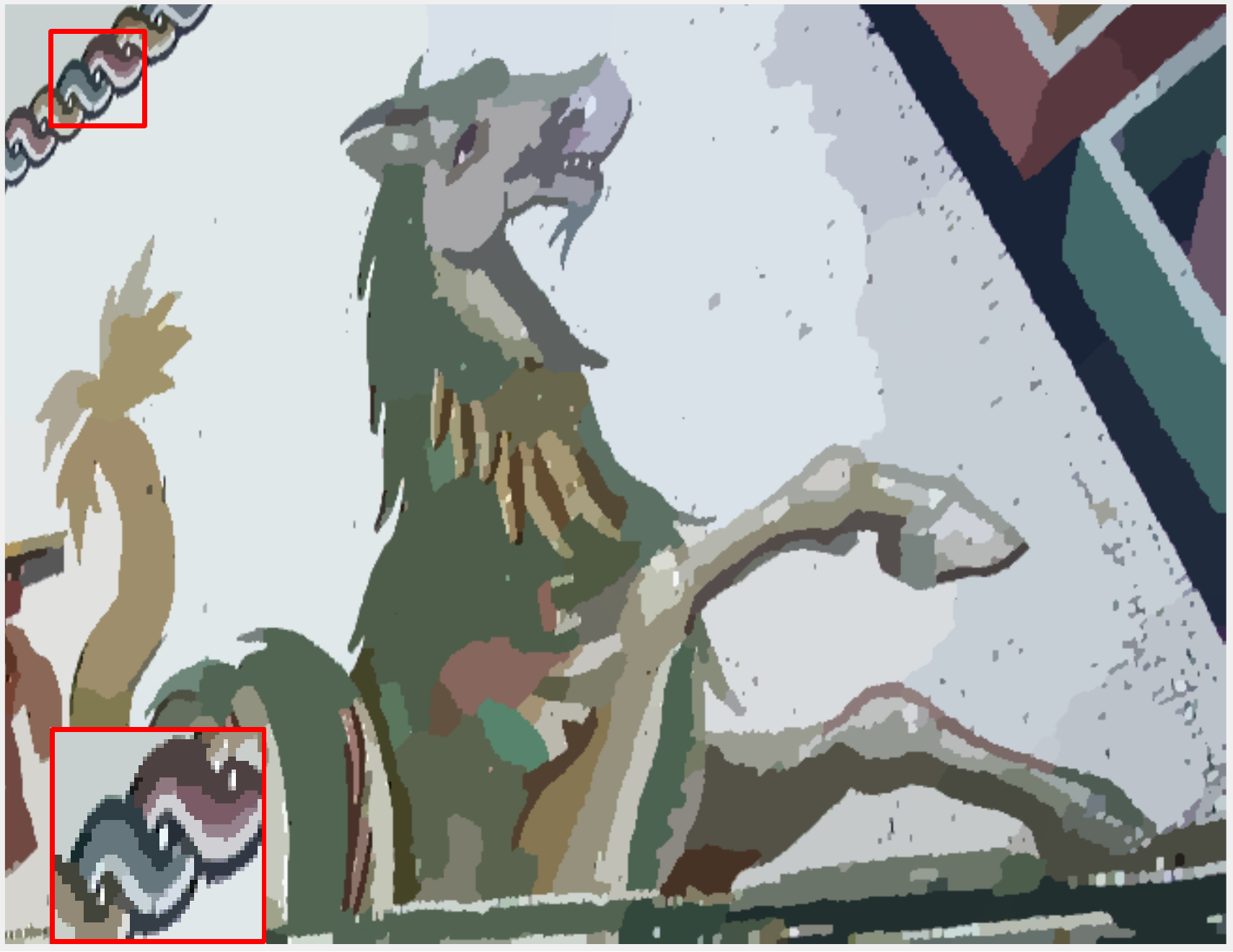}}
	\hfill
    \subfloat[$0.93$ s.]{\includegraphics[width=0.124\linewidth,keepaspectratio]{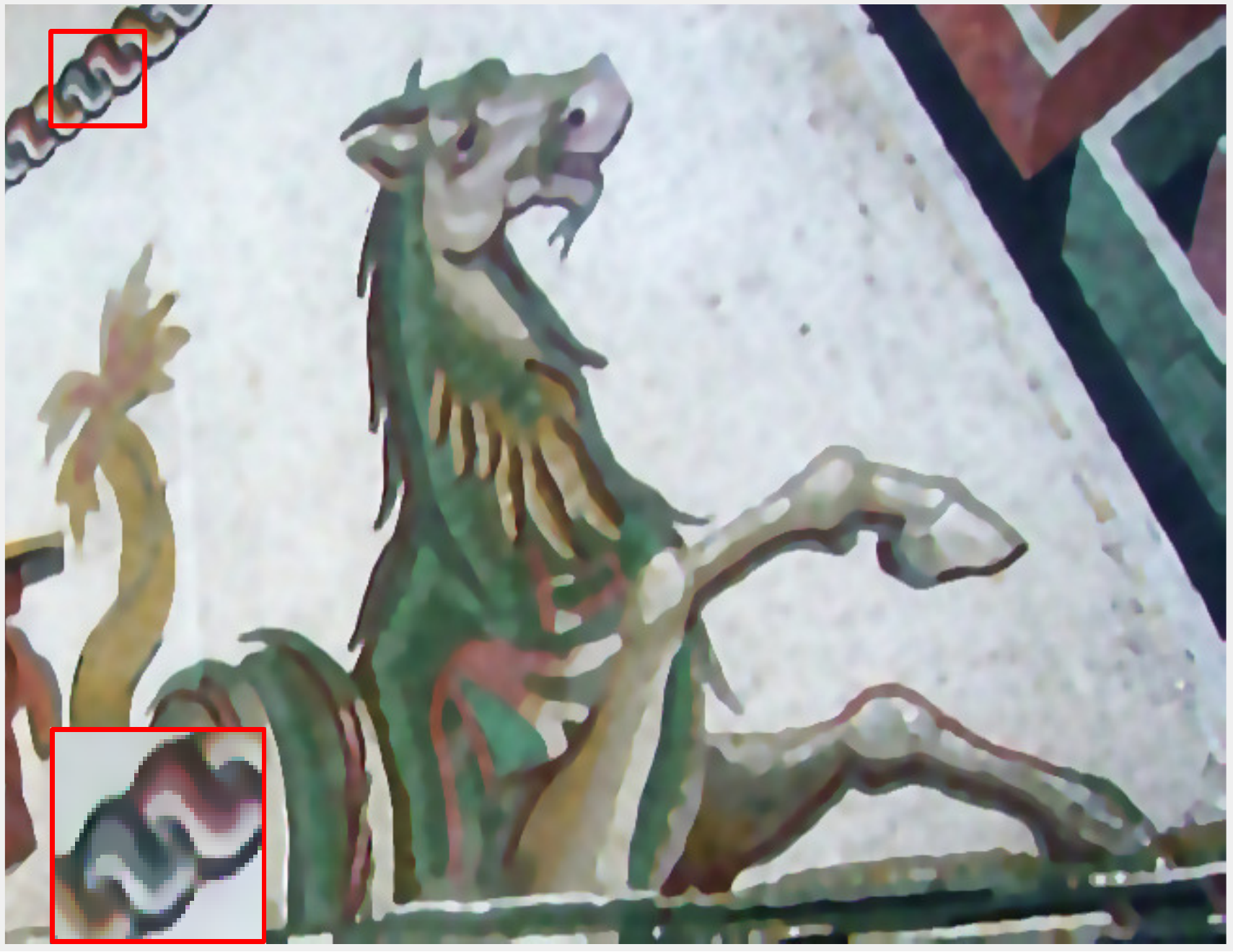}}
    \hfill
    \subfloat[$7.6$ s.]{\includegraphics[width=0.124\linewidth,keepaspectratio]{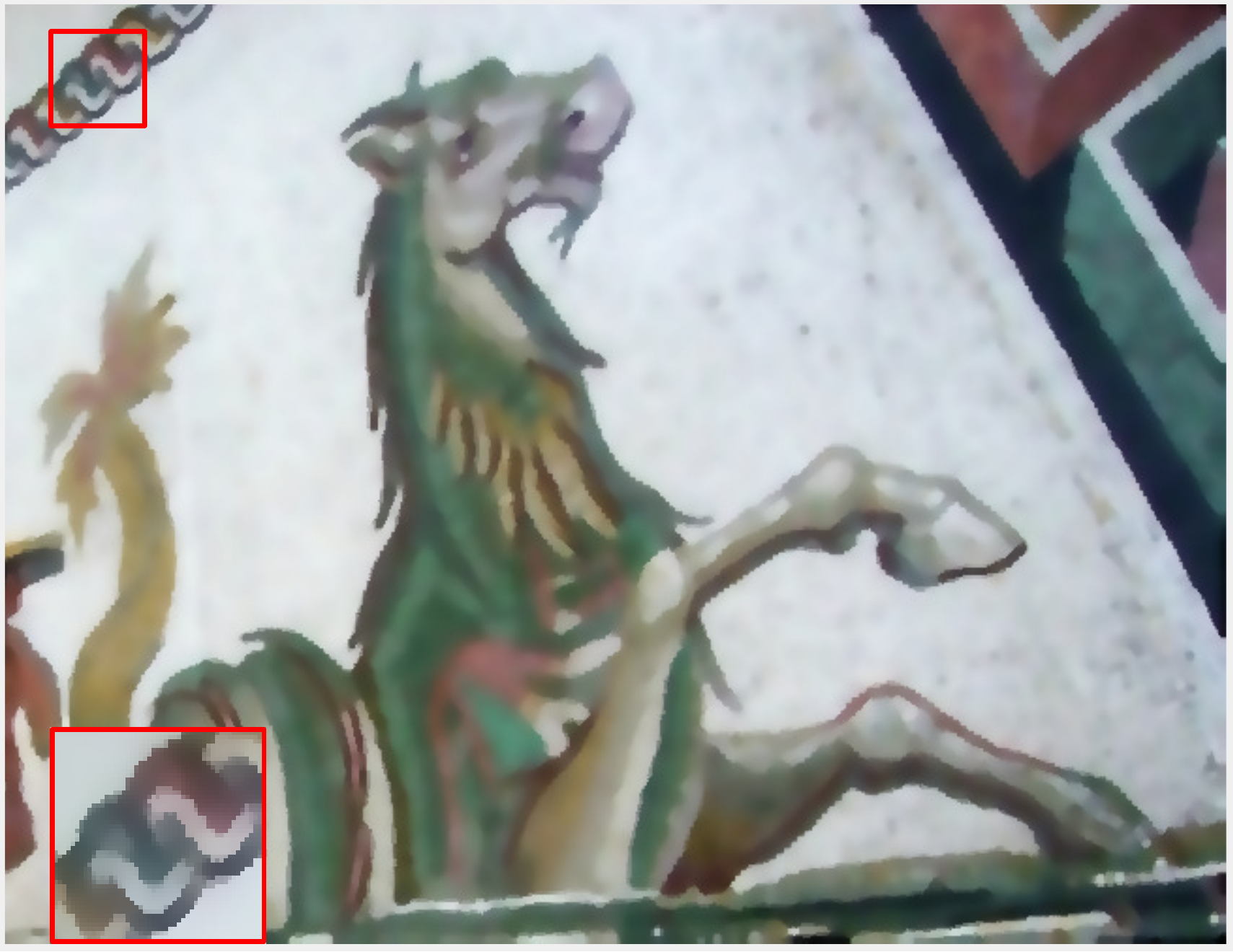}}
    \hfill
    \subfloat[$54$ s.]{\includegraphics[width=0.124\linewidth,keepaspectratio]{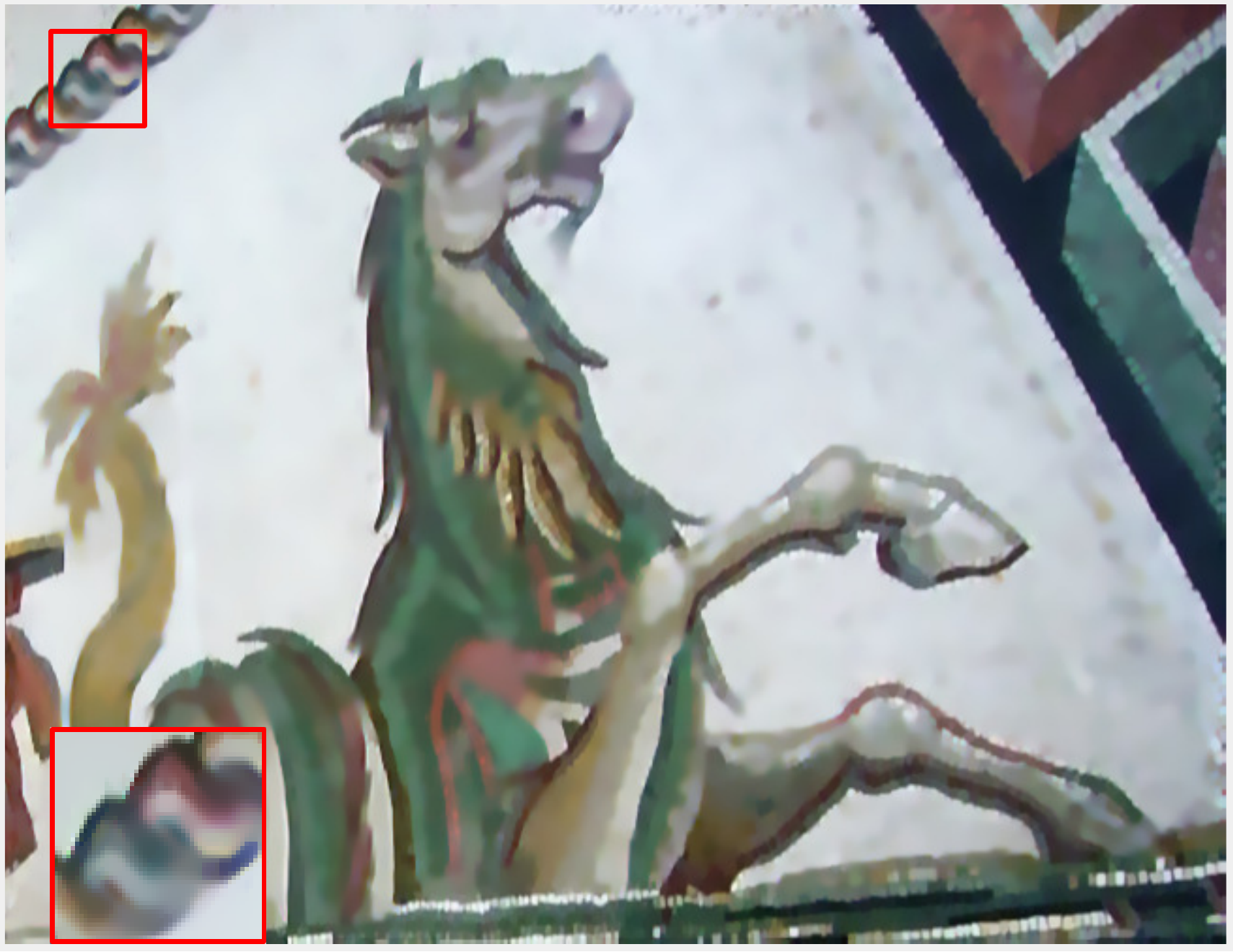}}
    \caption{Texture filtering results obtained using different methods.
    Each row contains the results for a different image.
    The first column in every row shows the input image.
    The remaining columns show the results in the following order (left to right): proposed method, \cite{Ham2018}, \cite{Belyaev2018}, \cite{Ono2017}, \cite{Zhang2014}, \cite{Cho2014}, \cite{Szolgay2012}.
    The timings are reported in the captions.
    Please zoom in for a clearer view.
    Images courtesy of \cite{Cho2014}.
    }
    \label{fig:Texture2}
\end{figure*}

We have already seen in Figure \ref{fig:Demo} that adaptive bilateral filtering offers more flexibility for texture filtering.
The problem of separating textures from the underlying structure is a well-studied one; see e.g. \cite{Buades2010,Szolgay2012,Xu2012,Cho2014,Ono2014,Zhang2014,Ham2018,Xu2018,Belyaev2018}.
While the classical bilateral filter can remove fine textures from an image, it cannot directly be used for removing coarse textures.
However, it was shown in \cite{Cho2014} that a variant of the bilateral filter, the joint bilateral filter, can be used for texture filtering.

We propose to use adaptive bilateral filtering for this task, by setting $\theta(i) = f(i)$ and adjusting $\sigma(i)$ so that textured regions are aggressively smoothed out (along with fine details and homogeneous regions), while sharp edges are not.
In this regard, we require a metric which can discriminate between texture and strong edges in an image.
One such metric is the modified relative total variation (mRTV) \cite{Cho2014}.
This is given by
\begin{equation*}
\mathrm{mRTV}(i) = \Delta(i) \frac{\max_{j \in W_i} \lVert \nabla f(j) \rVert}{\sum_{j \in W_i} \lVert \nabla f(j) \rVert + \epsilon},
\end{equation*}
where $W_i$ is a local neighborhood of $i$, $\nabla f(i)$ is the spatial gradient at $i$,
\begin{equation*}
\Delta(i) = \max_{j \in W_i} f(j) - \min_{j \in W_i} f(j),
\end{equation*}
and $\epsilon >0$ is a small number used to avoid division by zero.
The mRTV value is high near sharp edges and low in textured regions;
see $\cite{Cho2014}$ for further details about mRTV.

As before, we obtain $\sigma(i)$ via an affine transformation of $\mathrm{mRTV}(i)$, so that $\sigma(i)$ is high whenever $\mathrm{mRTV}(i)$ is low and vice versa.
This ensures that aggressive smoothing is performed in textured regions, but without excessively blurring the strong edges.
An example of a $\sigma$ map computed using this rule is presented in Figure \ref{fig:Texture1}.
These $\sigma$ values are used in Algorithm \ref{alg:Proposed}.
The computation of $\mathrm{mRTV}(i)$ and $\sigma(i)$ were performed on a grayscale version of the color image.

To completely remove coarse textures, we apply Algorithm \ref{alg:Proposed} for a second time, after scaling down $\sigma(i)$ by a factor of $0.8$ (to reduce the smoothing). This smooths out the textures completely, but introduces some blurring near sharp edges. We therefore use the method in Section \ref{sec:app}-A to sharpen the filtered image.
In summary, we apply Algorithm \ref{alg:Proposed} thrice --- twice for smoothing and once for sharpening.
Examples are shown in Figure \ref{fig:Texture2} (second column), where the parameters of Algorithm \ref{alg:Proposed} have been chosen to obtain the best visual output.
The color images are filtered on a channel-by-channel basis.

We note that the rolling guidance filter \cite{Zhang2014} uses a similar strategy, namely smoothing followed by sharpening, to remove coarse textures in an image.
In this method, a Gaussian smoothing is first applied to the input image to smooth out the textures.
This is followed by a number of iterations of the joint bilateral filter to reconstruct the blurred edges.

We compare our method with six different methods, namely \cite{Cho2014,Ham2018,Belyaev2018,Ono2017,Zhang2014}, and \cite{Szolgay2012}
\footnote{Codes: \cite{Ham2018}: \url{https://github.com/bsham/SDFilter}\\
\cite{Ono2017}: \url{https://sites.google.com/site/thunsukeono/}\\
\cite{Zhang2014}: \url{http://www.cse.cuhk.edu.hk/leojia/projects/rollguidance/}\\
\cite{Szolgay2012}: \url{https://www.researchgate.net/publication/261438330_CartoonTexture_decomposition_-_source_and_test_sample}}.
The filtering results on three different images are shown in Figure \ref{fig:Texture2}.
For all methods, we have used the default parameters as far as possible; whenever the default parameters are not available, we adjusted the parameters to get the best visual output.

For the image in the first row of Figure \ref{fig:Texture2}, the best filtering results are obtained using \cite{Ham2018} and \cite{Belyaev2018}.
Our method comes close, but is faster.
The method in \cite{Ono2017} removes textures very well, but causes loss of shading.
On the other hand, the shading is well preserved by  \cite{Zhang2014}, but textured regions are not completely smoothed out.
The output of \cite{Cho2014} looks somewhat blurry, while in \cite{Szolgay2012} some residual textures still remain near edges (compare the highlighted patches).

A similar observation applies to the results in the other two rows of Figure \ref{fig:Texture2}.
An interesting observation is that the structure of the eye in the second row is preserved well by our method and \cite{Ono2017}, but other methods introduce smearing along the edges.
This is evident from the highlighted patches.
For the image in the third row, we note that our method preserves small-scale structures quite well (see the highlighted regions), a property which is also shown by \cite{Ham2018} and \cite{Ono2017}.

Comparing the timings for each of the three images, we note that the rolling guidance filter \cite{Zhang2014} is the only method which is faster than our method.
However, as noted above, the proposed method exhibits a better texture suppressing property than \cite{Zhang2014}.
The proposed method is about $2 \times$ faster than \cite{Belyaev2018}, about $5 \times$ faster than \cite{Ham2018}, and about $3 \times$ faster than \cite{Cho2014}.
It is more than $20 \times$ faster than both \cite{Ono2017} and \cite{Szolgay2012}.

\section{Conclusion}
\label{sec:con}

We developed a fast algorithm for adaptive bilateral filtering by approximating histograms using polynomials.
To the best of our knowledge, this the first such algorithm.
In particular, we showed that our proposal significantly improves on the idea from \cite{Mozerov2015} of approximating histograms using uniform priors. Moreover, we demonstrated that our algorithm can produce acceptable results using just few convolutions; this enables us to achieve significant accelerations. 
In particular, the proposed algorithm is at least  $20\times$ faster than the brute force implementation.
It should be noted that the algorithm can work with any spatial filter, and not just Gaussians.
Moreover, if an $O(1)$ implementation is available for the spatial filter, then the overall filtering can be computed at $O(1)$ cost with respect to the spatial filter size.

We also demonstrated the effectiveness of the proposed algorithm for sharpening, deblocking, and texture filtering.
In particular, the proposed algorithm is accurate enough to perform image sharpening as per the proposal in \cite{Zhang2008}.
By adapting the method in \cite{Zhang2009}, we showed that our algorithm can also be used to satisfactorily deblock JPEG compressed images. 
In particular, we showed that our algorithm is much faster than iterative optimization methods for deblocking, though the outputs are visually similar.
Finally, we proposed a novel method for texture filtering using our algorithm, that is competitive with state-of-the-art methods in terms of runtime and performance.

Due to the flexibility offered by the adaptive bilateral filter, it can in principle be used to improve upon any application that has traditionally been performed using the bilateral filter, provided one can come up with the right rule for adapting the local $\sigma(i)$ values.
Having offered a fast algorithm, we hope to draw the interest of the community towards harnessing this versatility.
An interesting open question is whether the present ideas can be adapted for processing color images in the joint color space.

\section{Acknowledgements}

We thank the editor and the anonymous reviewers for their suggestions, and the authors of \cite{Li2014,Zhang2016_concolor,Zhao2017,Szolgay2012,Zhang2014,Ono2017,Ham2018} for sharing their codes publicly.
We also thank Jan~Allebach for sharing the code of \cite{Zhang2008} and the image in Figure \ref{fig:Sharpening3}, and Alexander~Belyaev for sharing the code of \cite{Belyaev2018}.

\section{Appendix}

\subsection{Proof of Proposition \ref{prop:Moments}}
\label{proof1}

Note that
\begin{equation*}
\sum_{t \in \Lambda_i} t^k h_i(t) = \sum_{t \in \Lambda_i} t^k \left\{ \sum_{j \in \Omega} \omega(j) \delta(f(i-j)-t) \right\}.
\end{equation*}
On exchanging the sums over $\Lambda_i$ and $\Omega$, we obtain
\begin{align*}
\sum_{t \in \Lambda_i} t^k h_i(t)  &= \sum_{j \in \Omega} \omega(j)   \left\{ \sum_{t \in \Lambda_i} t^k \delta(f(i-j)-t) \right\} \nonumber \\
&= \sum_{j \in \Omega} \omega(j) f(i-j)^k\nonumber \\
&= (\omega \ast f^k) (i).
\end{align*}
This is simply the output of the Gaussian filtering of the $k$-th power of $f(i)$. Since Gaussian convolutions can be performed at fixed cost independent of $\rho$ \cite{Deriche1993,Young1995}, the claim follows.

\subsection{Proof of Proposition \ref{prop:mhist_moments}}
\label{proof2}

Clearly, $\mu_0=m_0$. Moreover, using \eqref{Lambda_tilde} and \eqref{mhist}, we have
\begin{align*}
\mu_k &= \sum_{t \in \Gamma_i} t^k h_i(\alpha_i+t(\beta_i-\alpha_i)) \nonumber \\
&= \sum_{u \in \Lambda_i} \left( \frac{u-\alpha_i}{\beta_i-\alpha_i} \right)^k h_i(u) \nonumber \\
&= \left(\beta_i-\alpha_i\right)^{-k}
\sum_{r=0}^k \binom{k}{r} (-\alpha_i)^{k-r} \sum_{u \in \Lambda_i} u^r h_i(u) \nonumber \\
&= \left(\beta_i-\alpha_i\right)^{-k} \sum_{r=0}^k \binom{k}{r} (-\alpha_i)^{k-r} m_r. 
\end{align*}

\subsection{Proof of Proposition \ref{prop:BFmhist}}
\label{proof3}

Let $y = (\beta_i-\alpha_i)^{-1} (t-\alpha_i)$. In terms of $y$, we can write the numerator of \eqref{BF1hist}  as
\begin{equation*}
\begin{aligned}
\sum_{y \in \Gamma_i}  \big( \alpha_i + y (\beta_i-\alpha_i) \big) H_i(y) \phi_i \big( \alpha_i + y (\beta_i-\alpha_i) - f(i) \big).
\end{aligned}
\end{equation*}
From \eqref{eq:r_kernel} and \eqref{sub}, it can be verified that
\begin{equation*}
\phi_i \big( \alpha_i + y (\beta_i-\alpha_i) - f(i) \big) = \psi_i \big( y - \theta_0(i) \big).
\end{equation*}
In other words, we can write the numerator \eqref{BF1hist} as
\begin{align*}
\alpha_i  \sum_{y \in \Gamma_i} & H_i(y) \psi_i \big(y - \theta_0(i) \big) \  + \\
	& (\beta_i-\alpha_i)  \sum_{y \in \Gamma_i} y H_i(y) \psi_i \big(y-\theta_0(i) \big).
\end{align*}
Similarly, we can write the denominator \eqref{BF1hist} as
\begin{equation*}
\sum_{y \in \Gamma_i} H_i(y) \psi_i \big(y-\theta_0(i) \big).
\end{equation*}
On dividing the above expressions for the numerator and the denominator, we get \eqref{BFmhist}.

\bibliographystyle{IEEEtran}
\bibliography{citations}

\end{document}